\documentclass{article}

\usepackage{graphicx}
\usepackage{wrapfig}

\usepackage[tight,normalsize,bf,IT]{subfigure}

\usepackage{tabulary}
\usepackage{multirow}
\usepackage[margin=1in]{geometry}
\usepackage{titlesec}
\usepackage[labelfont=bf,textfont=it,font=footnotesize]{caption}
\usepackage{parskip}
\usepackage{mathpazo}
\usepackage[comma,sort&compress,numbers]{natbib}
\usepackage{booktabs}
\usepackage{listings}
\usepackage{float}

\newcommand\blfootnote[1]{%
  \begingroup
  \renewcommand\thefootnote{}\footnote{#1}%
  \addtocounter{footnote}{-1}%
  \endgroup
}

\usepackage{amssymb}
\usepackage{amsthm}
\usepackage{amsmath}
\usepackage{nicefrac}        
\usepackage{siunitx}

\graphicspath{{Figures/}}
\DeclareGraphicsExtensions{.pdf,.png,.jpg}

\titlespacing\section{0pt}{6pt plus 1pt minus 0pt}{3pt plus 1pt minus 0pt}
\titlespacing\subsection{0pt}{4pt plus 1pt minus 0pt}{3pt plus 1pt minus 0pt}
\titlespacing\subsubsection{0pt}{4pt plus 1pt minus 1pt}{3pt plus 1pt minus 1pt}
\titleformat{\section}{\large\bfseries\sffamily}{\thesection}{1em}{}
\titleformat{\subsection}{\normalsize\bfseries\sffamily}{\thesubsection}{1em}{}
\titleformat{\subsubsection}{\normalsize\bfseries\sffamily}{\thesubsubsection}{1em}{}

\usepackage[ruled]{algorithm2e}

\usepackage{flushend}

\usepackage{tabulary}
\usepackage{multirow}

\usepackage{array}
\usepackage{enumitem}
\usepackage{todonotes}
\usepackage{url}
\usepackage[T1]{fontenc}

\usepackage{hyperref}
\hypersetup{
    colorlinks,
    linkcolor={magenta},
    citecolor={blue},
    urlcolor={blue!80!black},
    breaklinks=true,
	plainpages=true
}
\newcommand{\cref}[2]{\hyperref[#2]{#1~\ref*{#2}}}
\newcommand{\colref}[2]{\hyperref[#2]{#1~\ref*{#2}}}

\newcommand{\figref}[1]{\colref{Figure}{#1}}

\newcommand{\secref}[1]{\colref{Section}{#1}}
\newcommand{\appref}[1]{\colref{Appendix}{#1}}
\newcommand{\tabref}[1]{\colref{Table}{#1}}
\newcommand{\coloredref}[2]{\hyperref[#2]{#1~\ref*{#2}}}
\newcommand{\coloredsubref}[3]{\hyperref[#2]{#1~\ref*{#2}{#3}}}
\newcommand{\comment}[1]{}

\begin{document}

\begin{center}
{\usefont{OT1}{phv}{b}{n}\selectfont\Large{Algorithmically-Consistent Deep Learning Frameworks for

Structural Topology Optimization}}

{\usefont{OT1}{phv}{}{}\selectfont\normalsize
{Jaydeep Rade$^1$, Aditya Balu$^1$, Ethan Herron$^1$, Jay Pathak$^2$, Rishikesh Ranade$^2$, \\
Soumik Sarkar$^1$, Adarsh Krishnamurthy$^1$*
\blfootnote{Paper accepted in Engineering Applications of AI, 2021 \url{https://doi.org/10.1016/j.engappai.2021.104483}.\\ Correspondences to \url{adarsh@iastate.edu}}}}

{\usefont{OT1}{phv}{}{}\selectfont\normalsize
{$^1$ Iowa State University\\
$^2$ Ansys Corporation\\
}}
\end{center}


\section*{Abstract}
Topology optimization has emerged as a popular approach to refine a component's design and increase its performance. However, current state-of-the-art topology optimization frameworks are compute-intensive, mainly due to multiple finite element analysis iterations required to evaluate the component's performance during the optimization process. Recently, machine learning (ML)-based topology optimization methods have been explored by researchers to alleviate this issue. However, previous ML approaches have mainly been demonstrated on simple two-dimensional applications with low-resolution geometry. Further, current methods are based on a single ML model for \emph{end-to-end} prediction, which requires a large dataset for training. These challenges make it non-trivial to extend current approaches to higher resolutions. In this paper, we develop deep learning-based frameworks consistent with traditional topology optimization algorithms for 3D topology optimization with a reasonably fine (high) resolution. We achieve this by training multiple networks, each learning a different step of the overall topology optimization methodology, making the framework more consistent with the topology optimization algorithm. We demonstrate the application of our framework on both 2D and 3D geometries. The results show that our approach predicts the final optimized design better (5.76$\times$ reduction in total compliance MSE in 2D; 2.03$\times$ reduction in total compliance MSE in 3D) than current ML-based topology optimization methods.

\subsection*{Keywords}
Topology Optimization  $|$ Deep Learning  $|$ Sequence Models  $|$ Physics-Consistent Learning

\vspace{0.2in}
\section{Introduction}
Over the past few decades, there has been an increased emphasis on designing components with optimal performance, especially using topology optimization~\citep{orme2017designing,liu2016survey}. Topology optimization (a subset of design optimization methods), initially developed by \citet{bendsoe1988generating}, refers to a set of numerical design optimization methods developed to find appropriate material distribution in a prescribed design domain to obtain geometric shapes with optimal performances. Here, the performance could be any physical phenomenon such as structural strength (or mechanical design), heat transfer, fluid flow, acoustic properties, electromagnetic properties, optical properties, etc.~\citep{sigmund2013topology}. The domain refers to a 2D or 3D volumetric mesh representation of the CAD geometry, typically used for finite element analysis. Among the different topology optimization methods, some of the most prominent approaches are solid isotropic material with penalization (SIMP)~\citep{bendsoe1989optimal}, level-sets~\citep{wang2003level}, and evolutionary optimization~\citep{das2011optimal,xie1993simple}. These approaches are used for several topological design problems where structural, acoustic, or optical performance needs to be optimal~\citep{eschenauer2001topology,sigmund2013topology} while removing the material to satisfy a total material (or volume) constraint.

\begin{figure*}[ht!]
\centering
\includegraphics[width=0.99\linewidth,trim={0.0in 0.0in 0.0in 0.0in},clip]{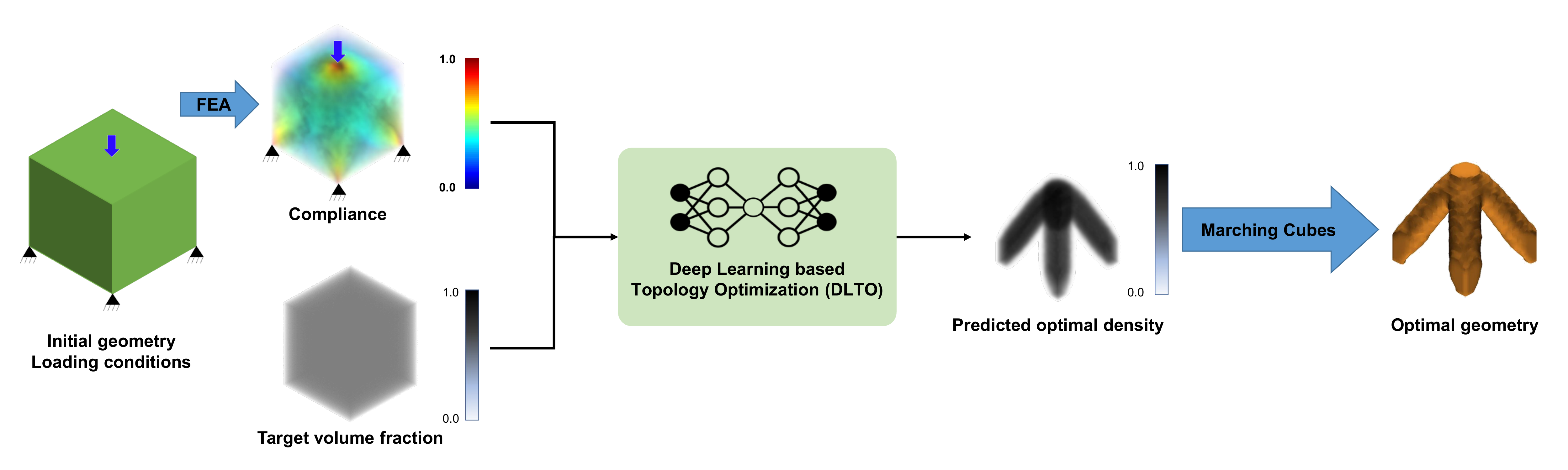}
\caption{\textbf{Overview:} The proposed deep learning-based topology optimization framework. The input to this framework is the compliance of the initial geometry along with the target volume fraction. Unlike SIMP, the DLTO framework predicts the optimal density of the geometry without any requirement of iterative finite element evaluations. The predicted optimal density of the geometry is then converted into triangular surface mesh representation using the marching cubes algorithm.}
\label{fig:overview}
\end{figure*}

One of the main challenges in performing topology optimization is the high computational cost associated with it. The performance measure that is being optimized needs to be computed after each iteration of the optimization process. These performance measures are usually obtained from physics simulations (often using numerical solution approaches, such as finite element analysis) that are typically compute-intensive. Due to this computational challenge, performing topology optimization for a fine (high resolution) topological mesh could take a few hours to even days. This computational challenge has inspired several researchers to develop deep learning-based topology optimization to reduce or eliminate the need for numerical simulations.

Although deep learning has many diverse applications and has demonstrated exceptional results in several real-world scenarios, our focus in this paper is the recent application of deep learning to learn the underlying physics of the system. There has been an increased interest in learning physical phenomena with neural networks to reduce the computational requirements and achieve better performance with very little or no data~\citep{pakravan2020solving,zhang2019quantifying,shah2019encoding,teichert2019machine_1,lu2019deepxde,jagtap2019adaptive,pan2019physics,raissi2017physics,bhatnagar2019prediction}. A popular approach relies on modifying the loss function to ensure that a set of physical constraints (boundary conditions) are satisfied. This approach has been especially successful in using deep learning to solve partial differential equations such as Burger's equation, Navier-Stokes equation, and Cahn-Hilliard's equation~\citep{shah2019encoding, singh2018physics, lu2019deepxde, jagtap2019adaptive, zhang2019quantifying, pan2019physics}. These approaches help the framework learn about the physical phenomena and make the learning consistent with the underlying physics. At the same time, better performance has been achieved by aligning the neural network architecture with the leaned phenomena~\cite{xu2020can}. With this motivation, we propose an algorithmically consistent deep learning framework for structural topology optimization.

A deep learning framework for the structural topology optimization need to (i) learn the underlying physics for computing the compliance, (ii) learn the topological changes that occur during the optimization process, and (iii) produce results that respect the different geometric constraints and boundary conditions imposed on the domain. To simplify the problem, we first discuss three essential elements that form the backbone of any data-driven approach: (i) the data representation, (ii) training algorithms, and (iii) the network architecture. As mentioned before, aligning the deep-learning framework with existing algorithms can provide better results and improved performance. For the framework to be algorithmically consistent, each of the three elements must be consistent with the classical structural topology optimization algorithm. In this particular instance, we focus on topology optimization using the solid isotropic material with penalization (called SIMP~\citep{bendsoe1989optimal}) algorithm for our framework. Thus our proposed framework is algorithmically consistent with SIMP topology optimization.

First, we align the data representation for the specific problem. Structural topology optimization is an iterative process where the design is modified through several iterations until the objective function (total compliance) converges to an optimal value. Further, each element's compliance is used in the sensitivity analysis for updating the element densities at each iteration. Thus, the element compliance is a valid and consistent representation of the geometry compared to other representations (such as voxel densities, strains, etc.) used in current deep learning approaches. Therefore, in the proposed framework, we use the element compliance as the CAD model representation of the geometry, loading, and boundary conditions (as shown in \coloredref{Figure}{fig:overview}). Note that, unlike the use of strain tensor and displacement tensor as proposed by \citet{zhang2019deep}, this representation is compact, leading to better scaling at higher resolutions.

Next, the training and inference pipelines need to be consistent with the classical structural topology optimization pipeline. In our experiments, we observe a non-trivial transformation of the densities from the first iteration to the final converged one. Due to this non-trivial transformation, learning the mapping between the initial topology and the final optimized topology is not a trivial \emph{one-step} learning task. Therefore, we use the intermediate densities obtained during data generation to enhance the performance of our proposed framework along with the initial compliance and target volume fraction as input and the final optimal density as the target.

Finally, the framework should simultaneously satisfy two constraints for structural topology optimization: the topological constraint of matching the target volume (often prescribed as a volume fraction or percentage of volume removed) and the physical constraint of minimizing the compliance. While computing the volume fraction is trivial, calculating the compliance involves performing a finite element solve. To avoid this computation, we propose developing a surrogate model for learning the mapping of a given intermediate density to its corresponding compliance.

In summary, we have developed two algorithmically consistent frameworks for structural topology optimization, namely, the Density Sequence (DS) prediction and the Coupled Density and Compliance Sequence (CDCS) prediction. The first approach uses a sequential prediction model to transform the densities without compliance. In the second approach, we add intermediate compliance to train a compliance-predicting surrogate model to improve results. We compare the proposed approaches with the baseline method, Direct Optimal Density (DOD) prediction. DOD prediction is an end-to-end learning approach where the final optimal density is directly predicted using just the initial compliance and the target volume fraction. The DS framework involves two convolutional neural networks (CNNs) for obtaining the final prediction, while the CDCS uses three CNNs iteratively during inference to predict the final optimal density.

In this paper, we develop a scalable, algorithmically consistent, deep-learning framework for 2D and 3D structural topology optimization. The main contributions are:
\begin{itemize}[]
    \item Two novel algorithmically consistent deep learning based structural topology optimization frameworks.
    \item An algorithmically consistent representation for topology optimization using the initial compliance of the design and the target volume fraction.
    \item Using intermediate densities and compliance data from the different optimization iterations obtained while generating the dataset to enhance the performance of our framework.
    \item Performance comparison of our proposed networks on both 2D and 3D geometries. We also validate and compare the performance of our approaches with the baseline SIMP-based topology optimization results.
\end{itemize}

The rest of the paper is arranged as follows. First, we discuss the formulation and related works to this paper in \secref{sec:formulation}. Next, we explain the deep learning methods proposed in our paper in \secref{sec:PCDL}.  We cover the details of the data generation process in \secref{sec:data_gen}, which is used as training data for our proposed approaches. In \secref{sec:results}, we show the statistical results from our experiments and demonstrate the performance of our proposed methods on both 2D and 3D structural topology optimization. Finally, we conclude this work with some future directions of research in \secref{sec:conclusions}.

\section{Formulation and Related Work}\label{sec:formulation}

\subsection{Formulation}

Formally, topology optimization can be formulated as:
\begin{equation}
 \begin{aligned}
     \text{minimize:}\; C(U)&\\
     \text{subject to:}\; \mathbf{K}\mathbf{U} &= \mathbf{F}\\
     g_i(\mathbf{U}) &\leq 0.
 \end{aligned}
\end{equation}
\noindent Here, $C(U)$ refers to the objective function of topology optimization. In the case of structural topology optimization, this is the compliance of the system,
\begin{equation}
C = \int_{\Omega \in \mathcal{S}} bu \; d\Omega + \int_{\tau\in d\mathcal{S}} tu\; d\tau\;
\end{equation}
where $b$ represents the body forces, $u$ displacements, $t$ surface traction, and $\Omega$ and $\tau$ are volume and surface representations of solid. The constraint $g_i(U)$ includes a volume fraction constraint, $g_i = (v/v_0) - v_f$. Since this optimization is performed at every element of the mesh, the combinatorial optimization is computationally intractable. An alternative solution is to represent the topology optimization equations as a function of density $\rho$ for every element.
\begin{equation}\label{eqn:TOFormulation}
\begin{aligned}
     \text{Minimize:}\; C(\rho,U)&\\
     \text{subject to:}\; \mathbf{K}(\rho)\mathbf{U} &= \mathbf{F}\\
     g_i(\rho,\mathbf{U}) &\leq 0\\
     0 < \rho &\leq 1
\end{aligned}
\end{equation}

\begin{algorithm}[t!]
    \caption{SIMP topology optimization~\citep{bendsoe1989optimal}}\label{Alg:TO}
    \SetKwInOut{Input}{Input}
    \SetKwInOut{Output}{Output}

    \Input{$\mathcal{S}$, L, BC, $V_0$}
    \Output{$D_{fin}$(set of all densities for each element, $\rho$)}
    \text{Load design; apply loads and boundary conditions}\\
    \text{Initialize: $D_0 \rightarrow V_0/\int_\Omega d\Omega$}\\
    \text{Initialize : $ch = \inf$}\\
    \While{$ch < threshold$}
    {
        Assemble global stiffness matrix $\mathbf{K}$ for element stiffness matrix $k_e(\rho_e)$\\
        Solve for $\mathbf{U}$, using $\mathbf{K}$, loads (L) and boundary conditions (BC)\\
        Compute objective function,
        $\mathbf{C} = \mathbf{U}^T\mathbf{K}\mathbf{U} = \sum_{e=1}^{N} \rho^{p} u^T k_e u $ \\ 
        Perform sensitivity analysis, $\frac{\partial{c}}{\partial{e}} = -p \rho^{(p-1)} u^T k_e u$\\
        Update the densities ($D_i$) using a optimality criterion\\
        $ch = ||D_i - D_{i-1}||$
    }
\end{algorithm}

This design problem is relaxed using the SIMP algorithm, where the stiffness for each element is described as, $E= E_{min} + \rho^p (E_{max} - E_{min})$. Here, $p$ is the parameter used to penalize the element density to be close to $1.0$. A typical SIMP-based topology optimization pipeline is shown in \coloredref{Algorithm}{Alg:TO}. While this is a naive implementation, more sophisticated methods for structural topology optimization such as level-set methods~\citep{wang2003level} and evolutionary optimization methods~\citep{das2011optimal,xie1993simple} are also popularly employed. Despite several advancements in structural topology optimization, a common challenge in all these approaches is that it requires several iterations of the finite element queries to converge on the final density distribution. Different optimization methods result in different yet comparable, optimal solutions alluding to the fact that multiple optimal solutions exist for the same topology optimization problem. Deep learning-based methods are a natural fit for accelerating this task, which has been explored previously, as described below.

\begin{figure*}[t!]
    \centering
    \includegraphics[width=0.99\linewidth,trim={0.0in 0.0in 0.0in 0.0in},clip]{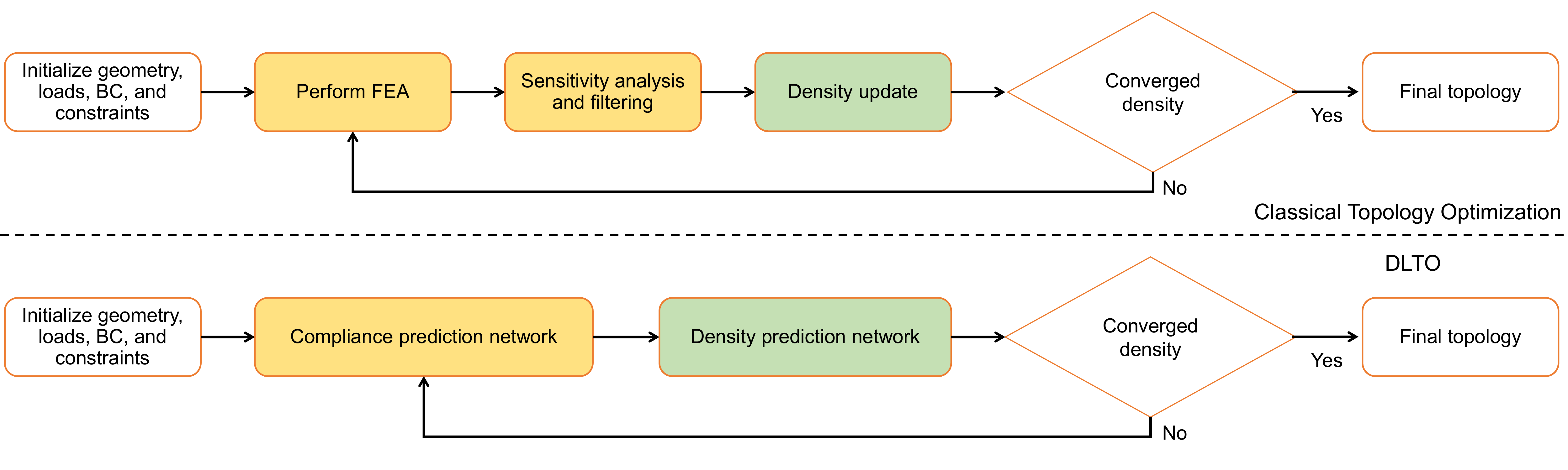}
    \caption{\textbf{Topology optimization pipeline:} The traditional topology optimization performs several iterations of finite element analysis, followed by sensitivity analysis and filtering. Using the filtered densities and compliance, we perform a density update. These iterations are performed several times till the density has converged. The DLTO approach replaces the repetitive performance of finite element analysis using a compliance prediction network and the density update with density prediction network.}
    \label{fig:pipelinedlto}
\end{figure*}

\subsection{Deep Learning for Topology Optimization}
Several deep learning-based topology optimization frameworks have been proposed~\citep{sosnovik2019neural,banga20183d,yu2019deep,zhang2019deep,nie2020topologygan,chandrasekhar2020tounn, qiyinLin,kollmann2dmetamaterials, rawat2019novel,Yu_2018,HAMDIA201921,shuheidoi, lagaros,oh_2019,baotongLi, diabWAbueidda,saskaiIPMmotor,zhang2020deep, oh2018GANTO,zhou2020,guo2018, leeSeunghye,de2019topology,bujny2018,takahashi2019convolutional, chaoQian2020,jang2020generative,rodriguez2021improved,poma2020optimization}. While we enlist several deep learning based topology optimization, several machine learning based methods without deep learning which have been explored in the recent years~\citep{mohammadzadeh2015new,sabzalian2019robust,kong2021fixed}. Further, there are several metaheuristics based topology optimization methods to reduce the computational time~\citep{tejani2018size,alberdi2015connection,gholizadeh2014topology,mortazavi2018comparison}.

Among the deep learning works, \citet{banga20183d} and \citet{sosnovik2019neural} proposed to perform the fine refinement of the design using deep convolutional autoencoders since the fine refinement stage usually requires several finite element iterations during the optimization process. \citet{sosnovik2019neural} used the densities obtained after five iterations of the SIMP-based structural topology optimization as input to a deep learning network that directly predicts the final density. \citet{banga20183d} extend this idea to 3D design geometries, along with an additional input of the boundary conditions, but for a very coarse geometric resolution ($12\times12\times24$). \citet{yu2019deep} developed a framework that takes the input design, boundary conditions, and the prescribed volume fraction and predicts the final target shape. They also create a generative framework where they generate several optimal designs. However, their research was restricted to only one type of boundary condition. A more generic framework to accommodate all possible boundary conditions using this method would require an impractically large dataset. 

\citet{zhang2019deep} developed an improved representation of the geometry, loading conditions, and boundary conditions using the strain tensor and displacement tensor as input. They demonstrate this framework using 2D geometries and represent each component of the strain tensor and displacement tensor as a different channel of the 2D image input. Using convolutional neural networks, they predict the final density. While their results are an improvement over earlier methods, this representation is not scalable to 3D. The strain tensor has three more components in addition to the increase in overall data size due to representing the geometry using 3D voxels, leading to several computational challenges. Recently, \citet{chandrasekhar2020tounn} proposed a topology optimization algorithm using neural networks where the neural network is used for identifying the density for each element at each iteration of the optimization process. This approach produces faster convergence and results that are comparable to those from SIMP. However, this approach's main drawback is that some finite element evaluations are still needed (although fewer than SIMP-based structural topology optimization). To the authors' best knowledge, very few researchers consider using compliance and the intermediate densities and compliances to improve the learning of structural topology optimization. Further, most of the implementations and results in the area have only been demonstrated in 2D or very low-resolution 3D geometries. Therefore, a scalable 3D framework for structural topology optimization using algorithmically consistent deep learning approaches is needed.

\section{Algorithmically-Consistent Deep Learning}\label{sec:PCDL}

A deep learning framework is algorithmically consistent if the data representation, training algorithm, and network architecture are all consistent with the underlying computational algorithm that the framework is designed to learn. Our framework is algorithmically consistent with the SIMP topology optimization. We first explain the baseline deep learning approach, which we use to compare our results. We also compare the performance of our proposed frameworks and the baseline against the classical SIMP-based structural topology optimization. After explaining the baseline, we explain the two proposed frameworks, the density sequence (DS) prediction, and the coupled density and compliance sequence (CDCS) prediction. \figref{fig:pipelinedlto} shows how our proposed frameworks are algorithmically consistent with the SIMP topology optimization.

\subsection{Baseline Direct Optimal Density Prediction}\label{sec:dod}
\begin{figure*}[t!]
    \centering
    \includegraphics[width=0.95\linewidth,trim={1.25in 4.2in 0.75in 1.5in},clip]{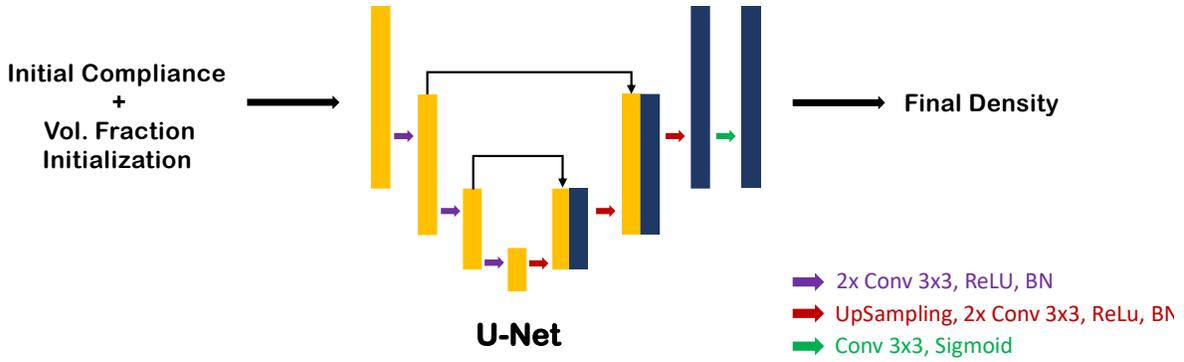}
    \caption{\textbf{Direct optimal density (DOD) prediction:} This baseline model is used for comparing our proposed frameworks. The input in this approach is the initial compliance for the geometry along with the target volume fraction initialized. Then we use a U-Net architecture for predicting the optimal density.}
    \label{fig:directprediction}
\end{figure*}

Recently, U-Nets~\citep{ronneberger2015u,cciccek20163d} have been known to be effective for applications such as semantic segmentation and image reconstruction. Due to its success in several applications, we chose a U-Net for this task. The input to U-Net is a tuple of two tensors. The first is the initial compliance (represented in the voxel or pixel space); the second is a constant tensor of the same shape as the compliance tensor. Each element of the constant tensor is initialized to the target volume fraction, which is a number between $[0, 1]$. First, a block of convolution, batch normalization, is applied. Then, the output is saved for later use for the skip-connection. This intermediate output is then downsampled to a lower resolution for a subsequent block of convolution, batch normalization layers, which is performed twice. The upsampling starts where the saved outputs of similar dimensions are concatenated with upsampling output for creating the skip-connections followed by a  convolution layer. This process is repeated until the final image shape is reached. At this point, the network utilizes a final convolution layer before producing the final density. The network architecture is shown in~\figref{fig:directprediction}.

We preprocess the compliance to transform it to the $[0, 1]$ range. We first take the $log_{10}$ of the compliance and then normalize it by subtracting the minimum value and then dividing by the difference of maximum and minimum values to scale the log values to $[0, 1]$ range, so all the inputs are in the same range. To train the neural network model such that it is robust to the loads applied on the input geometry, we augment the inputs by rotating the input tensor by 90$^\circ$ clockwise and counter-clockwise, 180$^\circ$ around all three axes and by mirroring the tensor along the X-Y plane, X-Z plane and, Y-Z plane. To understand data augmentation visually, we illustrate these augmentation operations on 2D-image in \coloredref{Figure}{fig:data_augmentation}. We threshold the final target density to get a binary density value. The density value of 1 corresponds to the element where the material is present, while the density value of 0 corresponds to the region where the material is absent or removed. We do not use intermediate compliance or intermediate densities to train this network. We use the Adam~\citep{kingma2014adam} optimizer for training, with an adaptive learning rate. To guide the optimizer, we use the binary cross-entropy function to calculate the loss between the predicted and the target density.

\begin{figure}[t!]
    \centering
    \includegraphics[width=0.65\linewidth,trim={2.5in 2.3in 2.5in 2.3in},clip]{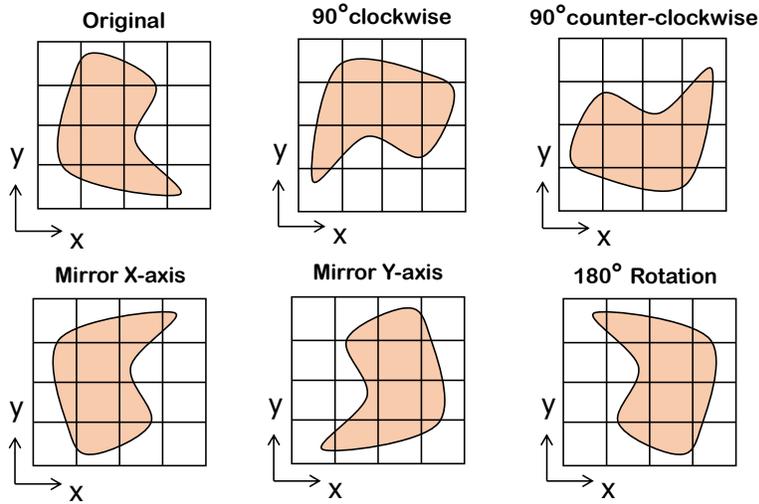}
    \caption{\textbf{Data Augmentation:} We show augmentation operations on 2D image including 90$^\circ$ clockwise and counter-clockwise rotation, 180$^\circ$ rotation and mirroring it vertically and horizontally. }
    \label{fig:data_augmentation}
\end{figure}


\subsection{Density Sequence Prediction}
For the data representation to be algorithmically consistent, we learn the structural topology optimization from compliance of the initial geometry. However, the compliance keeps evolving during the iterations since the densities also change during optimization. Therefore, the mapping between the original compliance and the final density is not trivial and may not directly correlate with the final density. To improve the performance, we develop the framework in two phases, as shown in \coloredref{Figure}{fig:pipelineDS}. The first phase is called an initial density prediction network (IDPN), which predicts the topology's initial density distribution based on the initial compliance per element obtained for the original geometry. With initial density, we use the iterative density transformation information available from the topology optimization process to transform the initially proposed density to the final optimized density. We perform this transformation using another network (density transformation network, DTN). The DTN does not use any information about the compliances. Therefore, using IDPN and DTN, we can predict the final densities for a given initial design and its corresponding original compliances.

The two phases of the Density Sequence Prediction method require two different network architectures, with each performing algorithmically consistent transformations of the given input information to obtain the final optimized shape. The first architecture corresponds to the first phase, where the task is to predict an initial density. The second architecture corresponds to the second phase, where the density obtained from phase 1 is transformed to a final density.

\begin{figure*}[b!]
    \centering
    \includegraphics[width=0.95\linewidth,trim={0.5in 2.7in 0.5in 2.6in},clip]{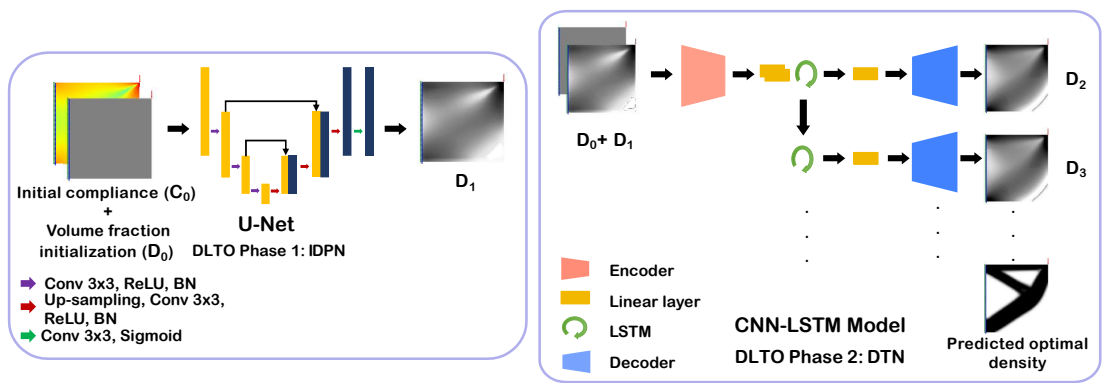}
    \caption{\textbf{Density sequence (DS) prediction:} In this framework, we perform the task in two phases as shown in \coloredref{Algorithm}{Alg:ds}; phase 1 (left block) and phase 2 (right block). We take the initial compliance and volume fraction initialization in the first phase to predict an initial density map. Using the initial density and the volume fraction initialization, we predict a series of densities similar to the prediction from a SIMP topology optimizer to finally predict the optimal density. The details of the training process are covered in the text.}
    \label{fig:pipelineDS}
\end{figure*}

\noindent\textbf{Phase 1: Initial Density Prediction:}
As a first phase of the method, the IDPN uses the initial elemental compliances and initialized volume fraction as input and predicts an initial density. We use U-Net~\citep{ronneberger2015u,cciccek20163d} network architecture for this phase. The architecture is similar to the architecture described in \secref{sec:dod} and is shown in \figref{fig:pipelineDS} on the left. 

For 2D phase 1 (IDPN), the initial compliance and the volume fraction constraint are represented as a two-channel \textit{``image''}, and the target is a one-channel \textit{``image''} of the element densities obtained after the first iteration of structural topology optimization. For 3D structural topology optimization, the input is a four-dimensional tensor with two 3D inputs concatenated along the fourth axis, and the target is a 3D element density. Data processing of the compliance (as described in \secref{sec:dod}) is necessary for IDPN.

\begin{algorithm}[t!]
    \caption{DS: Density sequence inference}\label{Alg:ds}
    \SetKwInOut{Input}{Input}
    \SetKwInOut{Output}{Output}

    \Input{$\mathcal{S}$, L, BC, $V_0$}
    \Output{$D_{k}$, final optimized geometry}
    \text{Load design; apply loads and boundary conditions}\\
    \text{Initialize: $D_0 \rightarrow V_0/\int_\Omega d\Omega$}\\
    Compute $C_0$ using L, BC and $D_0$\\
    $D_1 = IDPN(C_0, D_0)$ \tcc{IDPN Inference} 
    \For{$i=1:k$}
    {
        $D_{i+1} = DTN(D_{i-1},D_{i})$ \tcc{DTN Inference}
    }
    return $D_k$
\end{algorithm}

\noindent\textbf{Phase 2: Density transformation:} 
The training of phase 2 is more involved than phase 1. We train a convolutional neural network (CNN) with long short-term memory (LSTM), enabling learning from temporal data. In phase 2, there is a sequence of density transformations. Given these transformations are non-linear, a short-term history is not sufficient for robust prediction of the transformation. Capturing both long-term and short-term temporal dependencies is one of the salient features of LSTMs. Therefore, we use LSTMs and CNNs (traditionally used for spatial data such as images) to transform the densities. The architecture of the CNN-LSTM used for DTN is shown in \coloredref{Figure}{fig:pipelineDS} on the right.

The CNN-LSTM architecture starts with a set of convolution, max pooling, and batch normalization layers (called the encoder), which transforms the image to a latent space flattened embedding used by the LSTM. A sequence of LSTM layers is used to obtain a transformed latent layer. A set of deconvolution and upsampling layers (called Decoder) is used to obtain an image (representing the element densities after one iteration of structural topology optimization). The LSTM is unrolled for predicting a sequence in order to provide back-propagation through time. So, the intermediate densities of the structural topology optimization process are loaded as a sequence and processed to obtain the transformed density during training.

For phase 2 (DTN), the intermediate densities (each represented as a one-channel image) are used for performing the training. However, all the iterations of topology optimization are not significant in the learning process. Therefore, we curate the intermediate densities to only have unique densities (defined by a metric of $L_2$ norm). This uniquely curated set of densities are used for performing the training of DTN. Since DTN only deals with densities, no processing is required. To make the neural network more robust, we implement on-the-fly data augmentation, as discussed in \secref{sec:dod}.

\subsubsection{Training Algorithms}
For training IDPN, we use two different loss functions: (i) the mean-squared error between the predicted and target densities and (ii) the mean-squared error between the mean of the predicted and target densities. The second loss function ensures that the volume fraction of the target and predicted densities are the same. While training DTN, an additional loss function is added. Since the final geometry cannot have densities between $(0.0,1.0)$, the densities should belong to the set $\{0,1\}$ because of solid isotropic material. To impose this condition, we use the binary cross-entropy loss function and the two loss functions used for IDPN. In addition to the loss functions, we use stochastic gradient descent based optimizers such as Adam~\citep{kingma2014adam} for performing the optimization.

Once the training is performed, the \emph{learned} parameters for both the networks are joined such that an \emph{end-to-end} inference scheme can be performed. This inference scheme only requires the initial compliance and the volume fraction constraint (input to IDPN). The output of IDPN is used as input to DTN to get the final density without any additional information required. This \emph{end-to-end} scheme makes it applicable to any generic design.

\begin{figure*}[t!]
    \centering
    \includegraphics[width=0.99\linewidth,trim={0.5in 2.3in 0.5in 2.3in},clip]{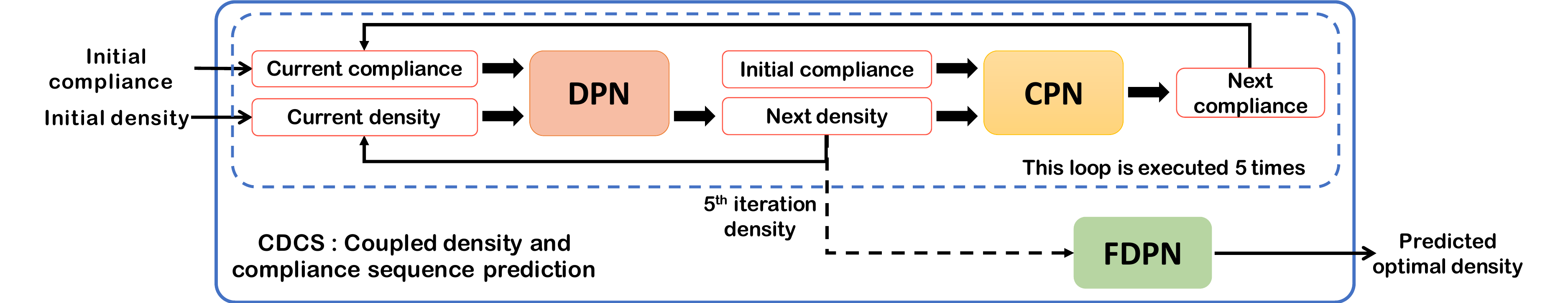}
    \caption{\textbf{Coupled density and compliance sequence (CDCS) prediction:} In this framework, the initial compliance (see text for more details) and volume fraction initialization is transformed by an iterative coupled prediction from a density prediction network (DPN) and compliance prediction network (CPN) as shown in \coloredref{Algorithm}{Alg:cdcs}. Five iterations of this process is performed to finally get the density and predicting the optimal density using a final density prediction network (FDPN). The details of the training process is covered in the text.}
    \label{fig:coupled_inference}
\end{figure*}

\subsection{Coupled Density and Compliance Sequence Prediction}\label{sec:CDCS}
Inspired by the iterative SIMP method, we use deep neural networks to develop a coupled density and compliance sequence prediction framework. In our dataset, we observed that the first five density iterations from the SIMP-based topology optimization method underwent more significant transformations compared to later iterations (also referred to as coarse and fine refinement by \citet{sosnovik2019neural}). We design three network architectures that use the intermediate compliances and intermediate densities to predict the final optimal density. The first two networks, namely, compliance prediction network (CPN) and density prediction network (DPN), feed their output as an input to each other as coupled interaction, and the third network, the final density prediction network (FDPN), uses the final output of density prediction network to produce the final optimal density (similar to the approach taken by \citet{sosnovik2019neural}).

\begin{algorithm}[b!]
    \caption{CDCS: Coupled density compliance sequence inference}\label{Alg:cdcs}
    \SetKwInOut{Input}{Input}
    \SetKwInOut{Output}{Output}

    \Input{$\mathcal{S}$, L, BC, $V_0$}
    \Output{$D_{fin}$, final optimized geometry}
    \text{Load design; apply loads and boundary conditions}\\
    \text{Initialize: $D_0 \rightarrow V_0/\int_\Omega d\Omega$}\\
    Compute $C_0$ using L, BC and $D_0$\\
    \For{$i=0:k$}
    {
        $D_{i+1} = DPN(D_{i},C_{i})$ \tcc{DPN Inference}
        $C_{i+1} = CPN(D_{i+1},C_{0})$ \tcc{CPN Inference}
    }
    return $D_{fin} = FDPN(D_k)$ \tcc{FDPN Inference}
\end{algorithm}

The compliance prediction network predicts the elemental compliance for a given iteration's density. It uses initial elemental compliance and the current iteration density obtained from the DPN. For CPN, we use Encoder-Decoder architecture. In the encoder, we use blocks of two convolutional layers followed by batch normalization. Similarly, we use an upsampling layer, two convolutional layers, and batch-normalization blocks for the decoder. The encoder encodes the input to the lower resolution latent space, and the decoder then decodes the encoded input to the next elemental compliance.

We use the current iteration elemental compliance and the current iteration density to predict the next iteration density for the density prediction network. We use U-SE-ResNet~\citep{nie2020topologygan} architecture for the DPN. Adding SE-ResNet~\citep{nie2020topologygan} blocks in the bottleneck region of U-Net architecture, in addition to the skip connections of U-Net from the encoder to the decoder, builds the U-SE-ResNet. The SE-ResNet block consists of two convolutional layers followed by SE(Squeeze-and-Excitation) block~\citep{huSE} with residual skip-connection from the input of the block. The encoder and decoder of U-SE-ResNet are the same as used in CPN architecture.  Refer to \coloredref{}{sec:appendix_architecture} for more details on the architectures of U-SE-ResNet.

The final model in this method is FDPN. As mentioned earlier, the elemental density has undergone a significant transformation during the first five iterations. So, we avoid the iterative process to obtain the final density by taking advantage of the neural network. We only use the fifth iteration density to predict the final optimal density directly. For FDPN we implement U-Net~\citep{ronneberger2015u,cciccek20163d} architecture. The encoder and decoder part of the U-Net used here is the same as discussed in CPN architecture.

Compliance is preprocessed before feeding it to the neural networks. We normalize the compliance values to be in the $[0, 1]$ range. The method for normalizing the compliance is explained in detail in \secref{sec:dod}.  In addition to this, we perform data augmentation discussed in \secref{sec:dod} for all three networks. More details on architectures mentioned in this section can be found in \ref{sec:appendix_architecture}.

\begin{figure*}[t!]
    \centering
    \includegraphics[width=0.9\linewidth,trim={0.75in 3.28in 0.75in 3in},clip]{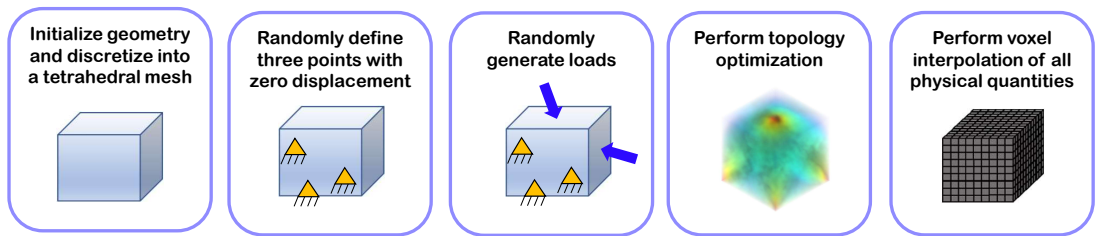}
    \caption{\textbf{3D data generation pipeline:} Each sample in the dataset if generated using this data generation pipeline. First we initialize the geometry (a cube with side length of 1 meter). This geometry is discretized into tetrahedrons to get the mesh. On this mesh, we define three non-collinear nodes to fix the mesh from any rigid body motion. Then we apply randomly generated boundary conditions and loading conditions with different magnitude and direction.}
    \label{fig:datagen}
\end{figure*}

\subsubsection{Training Algorithms}
All three networks are trained independently. During the training phase, we use Adam~\citep{kingma2014adam} optimizer for all three networks. For more efficient training, we use an adaptive learning rate.  The mean absolute error loss function is used for CPN. Moreover, for DPN and FDPN, the binary cross-entropy loss function is used to predict the densities. 

During inference, the first two networks are used iteratively (see \coloredref{Figure}{fig:coupled_inference}). We start with the initial compliance and initial density, which is initialized with a volume fraction value as a tensor with the same shape as the initial compliance tensor. Using the density prediction network, we predict the subsequent iteration's density and feed it as input to the compliance prediction network, producing the compliance corresponding to the new predicted density. This loop is executed five times, so we get the fifth iteration's density prediction at the end of the loop. We use this predicted fifth iteration density as input to the final density prediction network and directly predict the final optimal density.

\section{Data Generation}\label{sec:data_gen}

\subsection{2D Data Generation} 

The data required for training the networks is obtained by performing several simulations of topology optimization on different designs and volume fraction constraints. We represent each design using a 2D mesh made up of quadrilateral elements. The nodes of the mesh form a regular grid such that each element represents a square element. With this representation, we can directly convert the elements of the mesh to pixels of an image. Therefore, we represent the geometry as an image such that the pixel intensity values represent the element compliances and the element densities of the 2D mesh.

For training data, we need raw compliance values, the volume fraction constraint, the intermediate element densities obtained during the intermediate iterations of the structural topology optimization process, and the final element densities. We generated 30,141 simulations of the structural topology optimization with different randomly generated load values, loading directions, load locations, and a randomly generated set of nodes in the mesh, fixed with zero displacements. We performed each simulation for 150 iterations of SIMP-based structural topology optimization. All the relevant information from each structural topology optimization simulation is stored for use during the training process.

\subsection{3D Data Generation}\label{sec:3Ddatagen}
The 3D data used for DLTO is generated using ANSYS Mechanical APDL v19.2. We use a cube of length 1 meter in the form of 3D mesh as an initial design domain (see \coloredref{Figure}{fig:datagen}). The mesh created has 31093 nodes and 154,677 elements, and each element consists of 8 nodes. To ensure we sample a diverse set of topologies from the complete distribution of topologies originating from the cube, we use several available sets of boundary and load conditions in ANSYS software such as Nodal Force, Surface Force, Remote Force, Pressure, Moment, Displacement. First, we randomly sample three non-collinear nodes on one side of the cube, and we define zero displacements for these points; so they are fixed. This is necessary to avoid any rigid body motion of the geometry. The next step is to randomly select the load location, which is not close to the fixed support nodes. The nature of the load (nodal, surface, remote, pressure, or moment), the value, and direction is sampled randomly. The motivation behind the random sampling is to ensure the generated dataset has a variety of shapes and is independent of the type of load and its magnitude and direction. We employ a rejection sampling strategy to ensure that each sampled topology is unique. We obtained a total of 1500 configurations of load and boundary conditions, and then by sampling the volume fraction, we generated 13500 samples. In our dataset, the topology optimization took an average of 13 iterations; the minimum number of iterations is 6, and the maximum is 72; this number depends on several factors such as the mesh resolution, boundary conditions, and the target volume fraction.

In ANSYS, we store the topology optimization output, the original strain energy, and the intermediate results stored using the starting mesh representation. We now need to convert the mesh representation to a voxel representation for training 3D CNN models. This conversion process first discretizes the axis-aligned bounding box into a regular structured grid of voxels based on the grid's grid size/resolution. We compute the barycentric coordinates for each tetrahedron in the mesh for each voxel center. Using the barycentric coordinates, we can estimate if the grid point is inside the tetrahedron or not. If the grid point is inside that tetrahedron, we now interpolate the field values (such as density, strain energy, etc.) from the tetrahedron nodes to the voxel centers. Through this process, we obtain the voxel-based representation of the topology optimization data. Each sample's voxelization takes about 5-15 minutes, depending on the resolution and the number of tetrahedral elements. We parallelize this process using GNU parallel to complete this process in a few hours (depending on compute nodes' availability). To calculate the element compliance, we multiply strain energy obtained from ANSYS with the cube of the density to obtain the compliance ($C = \rho^p\mathbf{u}k_e\mathbf{u} = \rho^p * SE$, where $p$ is the penalty of the SIMP approach, set to 3 in our data generation process, $SE$ refers to the elemental strain energy).

Once we obtain the voxel-based representation, we perform other preprocessing steps, such as normalizing the compliance by the maximum compliance value and converting the compliances to log scale for better learning. We even perform on-the-fly data augmentation by rotating the model in any of the six possible orientations. Thus we finally get the data for training the neural network. 

\section{Results}\label{sec:results}

We split both datasets (i.e., 2D and 3D geometries) into two parts for training the neural networks: training and testing dataset. Out of all data generated, we use 75\% of the topology optimization data for training and the remaining 25\% for testing. We use the testing dataset to evaluate the performance of all three methods. We will discuss the results for 2D and 3D topologies in the following subsections.

\subsection{Results on 2D Topology Optimization}\label{sec:2dresults}

\begin{figure*}[t!]
\centering
\includegraphics[width=0.33\linewidth,trim={0in 0.0in 0.5in 0.0in},clip]{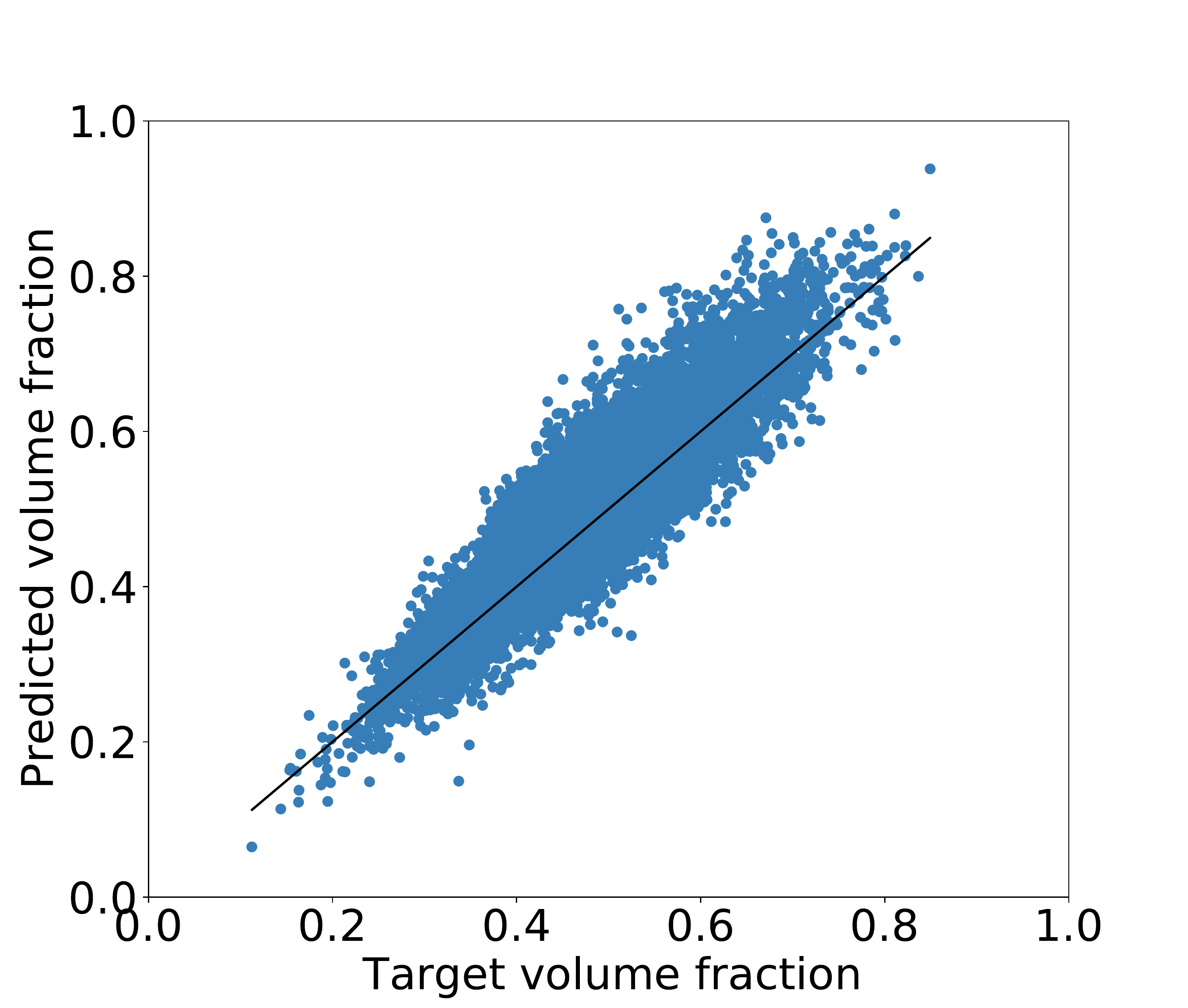}
\includegraphics[width=0.33\linewidth,trim={0in 0.0in 0.5in 0.0in},clip]{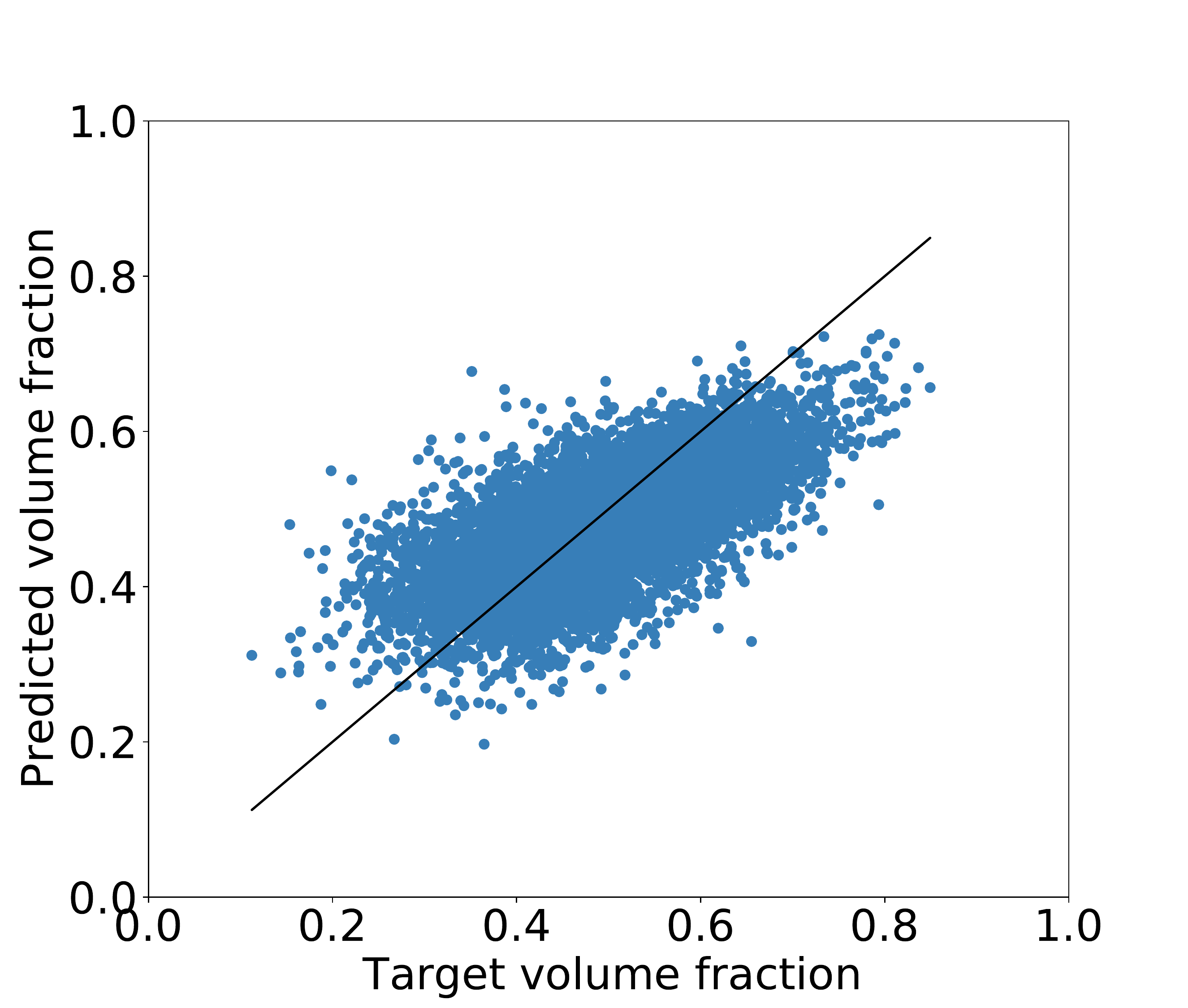}
\includegraphics[width=0.33\linewidth,trim={0in 0.0in 0.5in 0.0in},clip]{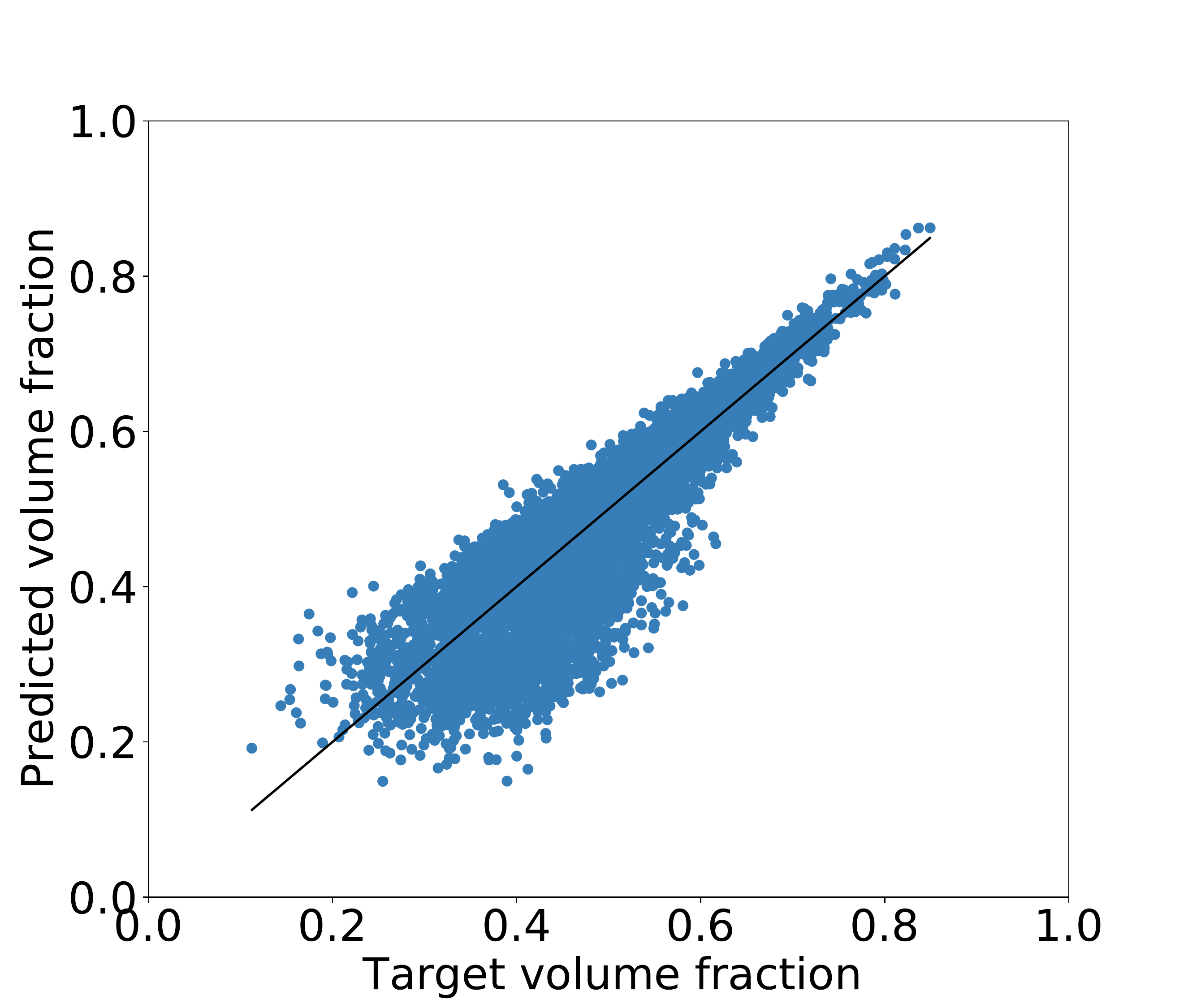}
\caption{Correlation plots between predicted volume fraction and target volume fraction on 2D test data for: (a) DOD, (b) DS, (c) CDCS.}
\label{fig:2D_vf_correlation} 
\end{figure*}

\begin{figure*}[t!]
    \centering
    \includegraphics[width=0.45\linewidth,trim={0in 0.0in 0.0in 0.0in},clip]{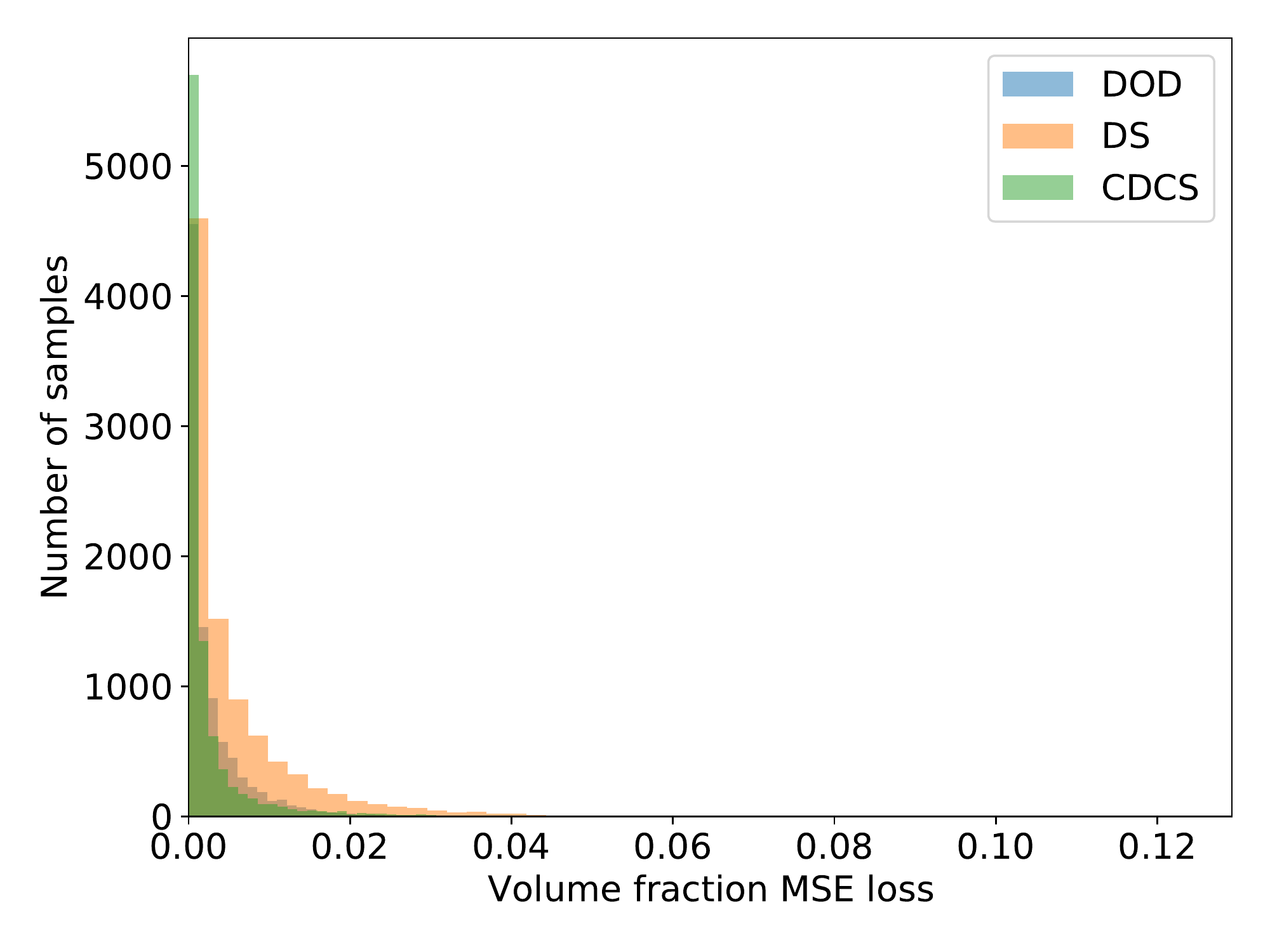}
    \includegraphics[width=0.45\linewidth,trim={0in 0.0in 0.0in 0.0in},clip]{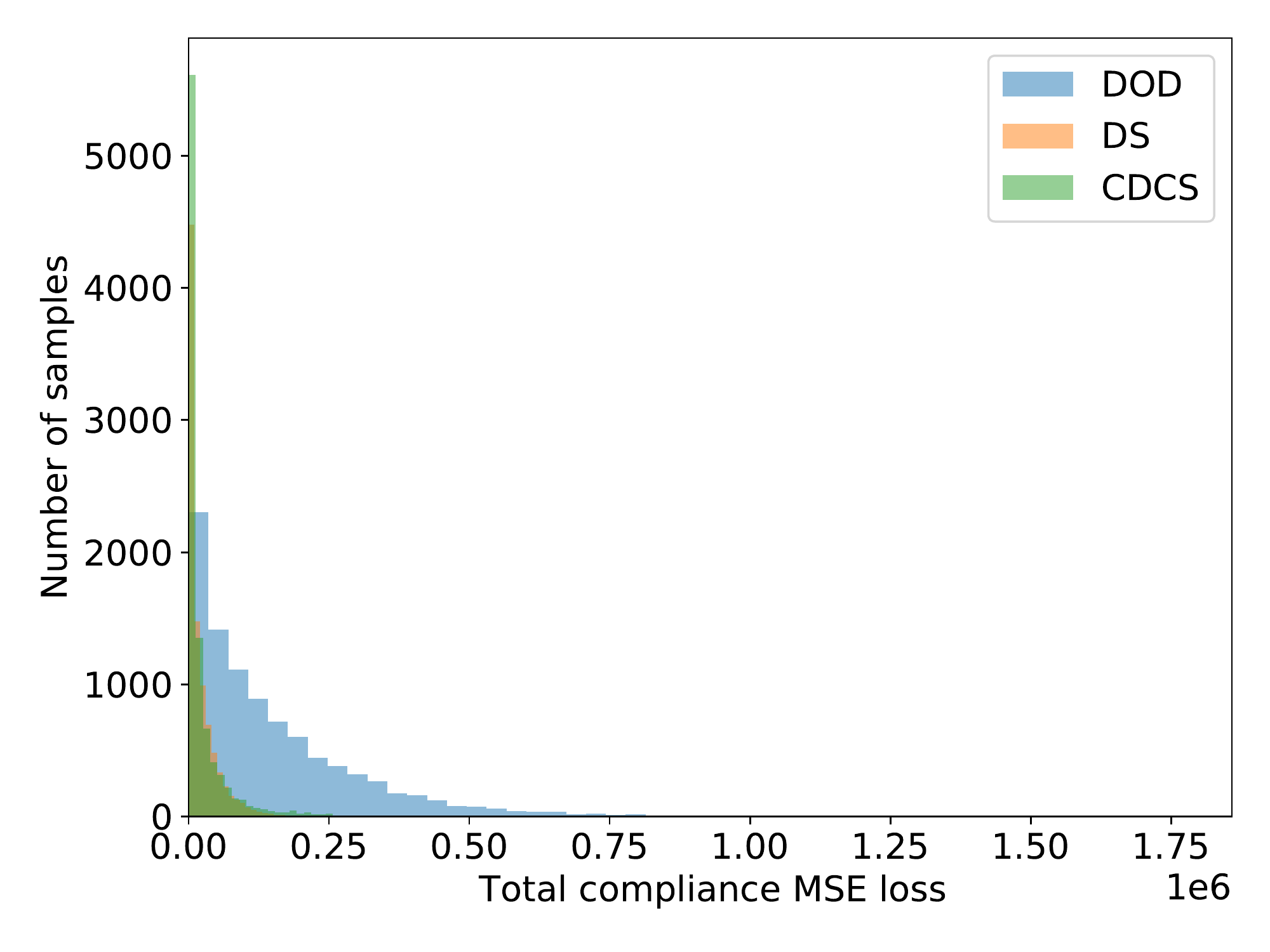}
    \caption{Histogram of (a) total volume fraction loss and (b) total compliance loss on the 2D test data.}
    \label{fig:2d_histograms}
\end{figure*}

To compare the performance of our proposed methods with the baseline DOD method, we start with the volume fraction (VF) constraint. We compute the predicted volume fraction of the final predicted topology by averaging the density values over the whole design domain. We compute the mean-squared error (MSE) between the predicted VF and the actual VF on the test data. This metric is shown as MSE of volume fraction in \tabref{tab:compare_2d}. We also plot the histogram for MSE values for VF from three methods in \figref{fig:2d_histograms}(a). Further, we plot the correlation plot between the predicted and actual VF for all the three methods in \figref{fig:2D_vf_correlation} and compute the Pearson's correlation coefficient between the predicted VF and actual VF for 2D test data in \tabref{tab:correlationcoeff_2d}.

\begin{table}[h!]
    \setlength\extrarowheight{3pt}
    \newcommand{\tabincell}[2]{\begin{tabular}{@{}#1@{}}#2\end{tabular}}
    \centering
    \small
    \caption{Comparison of test loss metrics of our three methods on 2D test data.}
    \label{tab:compare_2d}
    \begin{tabular}{| r | r | r | r | r | r |}
    \hline
    \multirow{2}{*}{\textbf{Method}}  &  \textbf{VF} & \textbf{TC} &  \multicolumn{3}{|c|}{\textbf{Density}} \\ \cline{2-6}
      &  \textbf{MSE} & \textbf{MSE} &  \textbf{BCE} & \textbf{MAE} & \textbf{MSE} \\
    \hline 
    DOD & 0.0035 & 15.1e+04  & \textbf{0.2354} & 0.1368 & \textbf{0.0737} \\
    DS & 0.0068 & \textbf{2.35e+04}   & 0.4421 & 0.1826 & 0.1206 \\
    CDCS & \textbf{0.0027} & 2.62e+04  & 0.3146 & \textbf{0.1195} & 0.0812\\
    \hline
    \end{tabular}
\end{table}


\begin{table}[t!]
    \setlength\extrarowheight{5pt}
    \small
    \newcommand\T{\rule{0pt}{2.7ex}}
    \newcommand\B{\rule[-1.3ex]{0pt}{0pt}}
    \newcommand{\tabincell}[2]{\begin{tabular}{@{}#1@{}}#2\end{tabular}}
    \centering
    \caption{Comparison of correlation coefficient(R) for volume fraction and total compliance on on 2D test data.}
    \label{tab:correlationcoeff_2d}
    \begin{tabular}{| r | l | l |}
    \hline
            \textbf{Method}  &  \textbf{R for volume fraction} & \textbf{R for total compliance}\\
    \hline 
    DOD & 0.8986 & 0.8403 \\
    \hline
    DS & 0.6883 & 0.9551 \\
    \hline
    CDCS & {0.8945} & {0.8926}\\
    \hline
    \end{tabular}
\end{table}

\begin{figure*}[t!]
    \centering
    \includegraphics[width=0.33\linewidth,trim={0in 0.0in 0.5in 0.0in},clip]{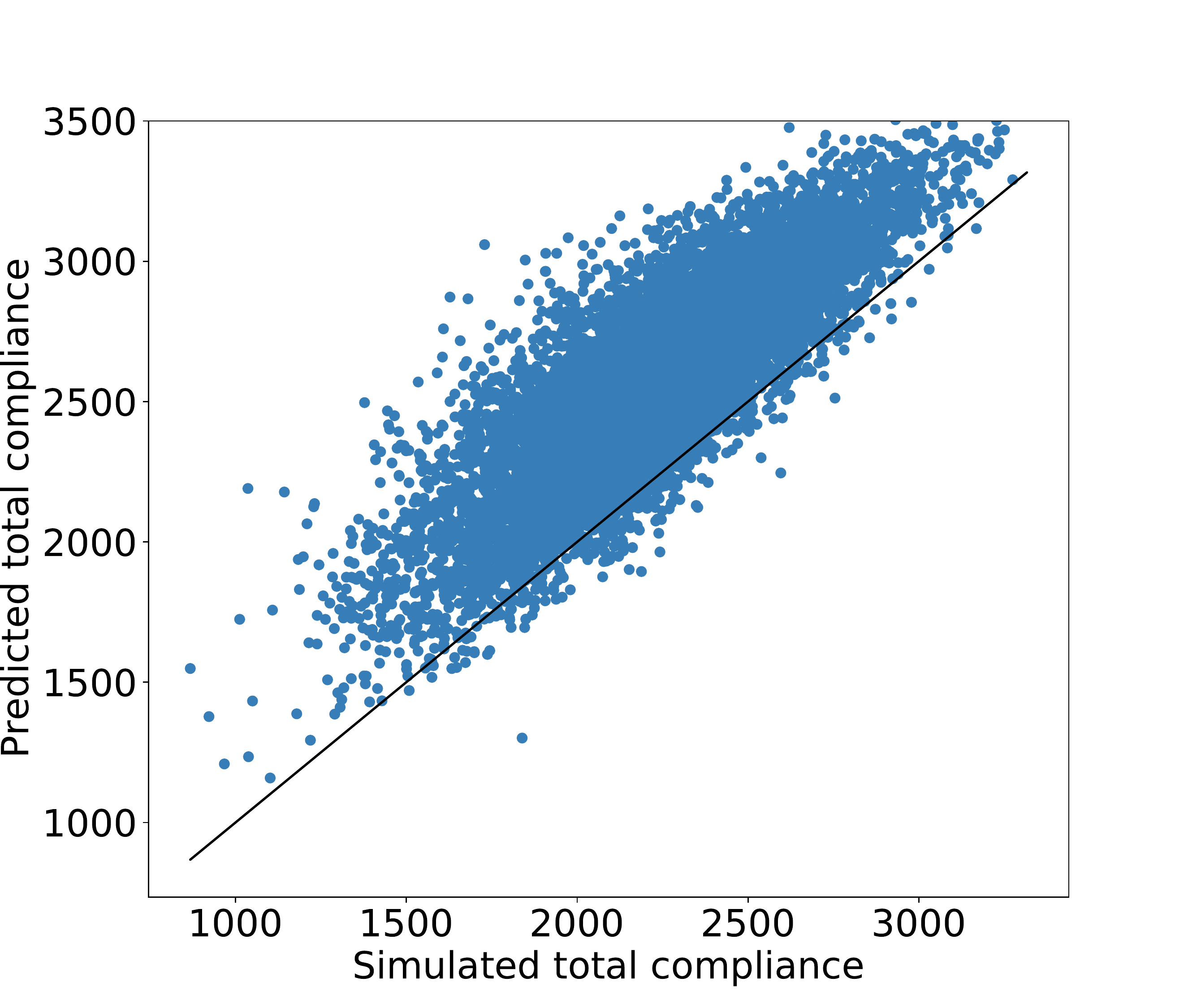}
    \includegraphics[width=0.33\linewidth,trim={0in 0.0in 0.5in 0.0in},clip]{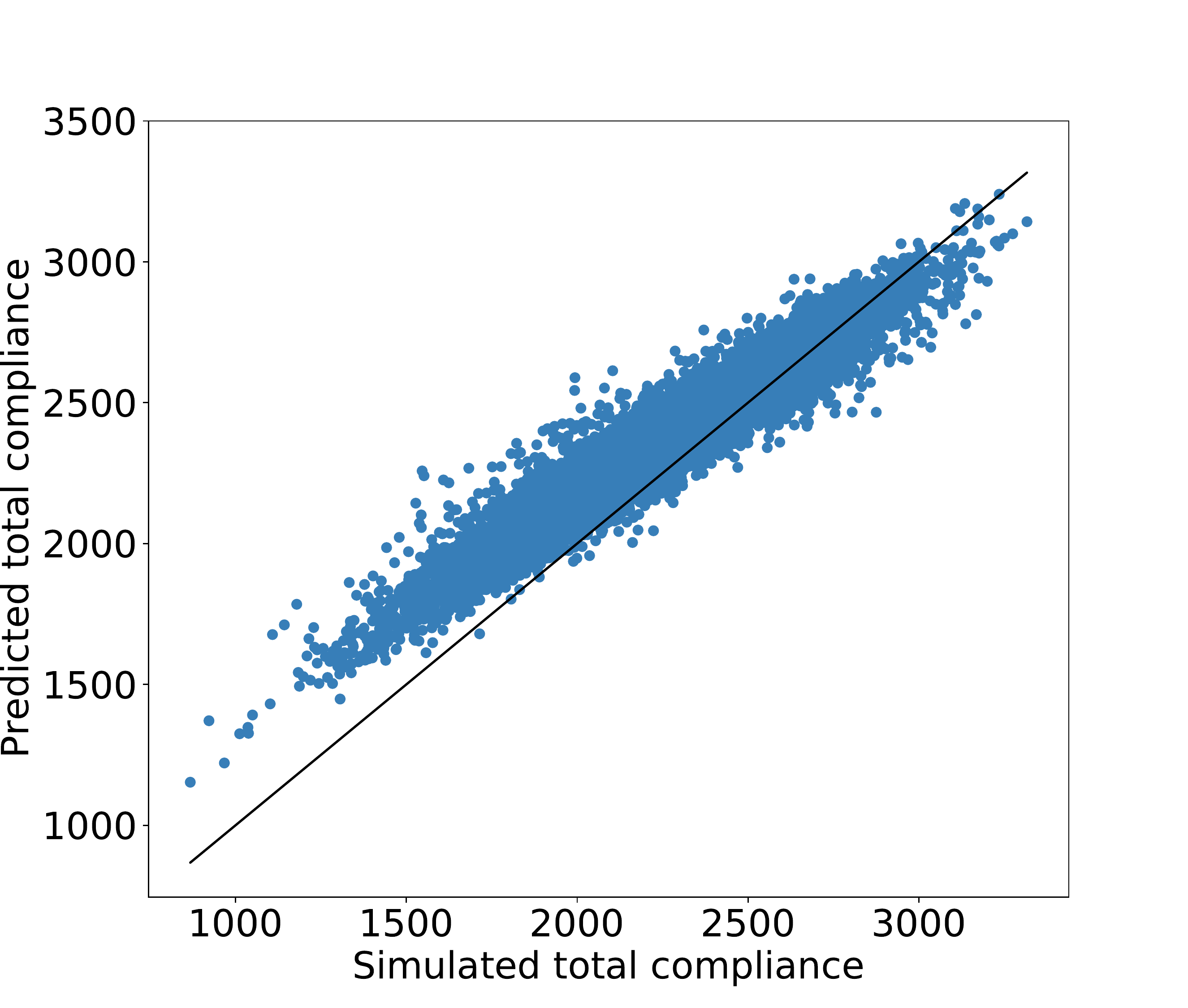}
    \includegraphics[width=0.33\linewidth,trim={0in 0.0in 0.5in 0.0in},clip]{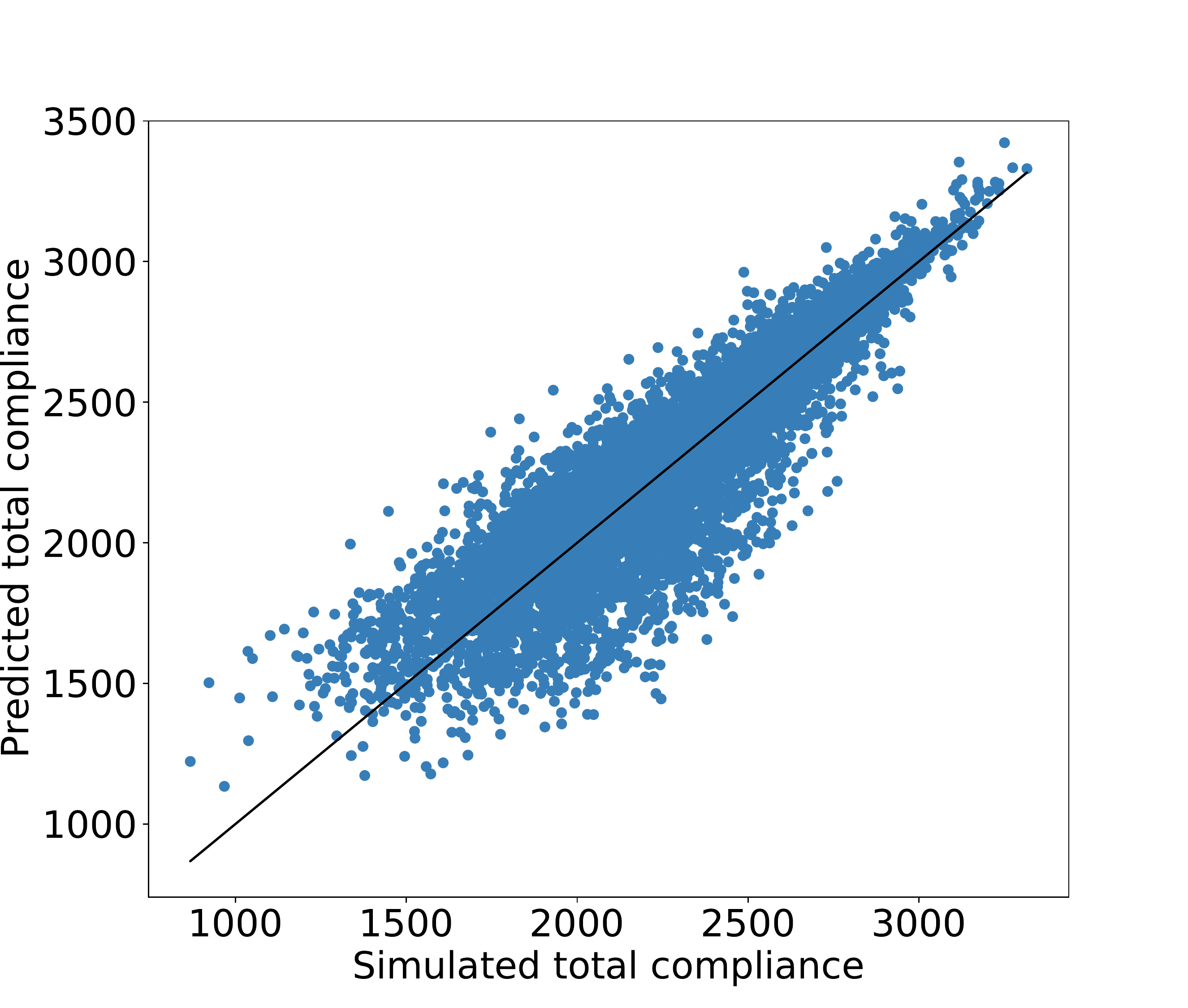}
    \caption{Correlation plots between predicted total compliance and simulated total compliance on 2D test data for all three methods: (a) DOD, (b) DS, (c) CDCS.}
    \label{fig:2D_comp_correlation} 
\end{figure*}

Next, we evaluate the performance of the methods using the physical constraint of topology optimization: total compliance (TC). TC is the SIMP algorithm's objective function value, which it tries to minimize while simultaneously satisfying the volume fraction constraint. We compute and compare the MSE between the predicted and simulated TC values. To determine the TC of the predicted final topology, we use the compliance prediction network (CPN), part of the CDCS framework, to predict the elemental compliance and take a sum of it over the whole design domain. We sum the elemental compliance of the target optimal topology to get the simulated total compliance value; this is the optimal minimum value achieved at the end of the SIMP method. We compare the MSE for TC in \tabref{tab:compare_2d} and also plot the histogram of MSE values from all three methods in \figref{fig:2d_histograms}(b). We also compute the Pearson's correlation coefficient between predicted total compliance and simulated total compliance values in \tabref{tab:correlationcoeff_2d} and the correlation plots between these two values for each of the three methods is in \figref{fig:2D_comp_correlation}.

\begin{table*}[t!]
    \newcommand\T{\rule{0pt}{2.7ex}}
    \newcommand\B{\rule[-1.3ex]{0pt}{0pt}}
    \centering
    \small
    \caption{Statistics on the volume fraction and total compliance loss on 2D test data.}
    \label{Tab:2dstats}
    \begin{tabular}{|c|c|c|c|c|c|c|}
    \hline
     \textbf{Statistics} \B & \multicolumn{3}{c|}{\T\B \textbf{MSE of volume fraction}} & \multicolumn{3}{c|}{ \textbf{MSE of total compliance}}\\
     \cline{1-7}
    \textbf{Method} &\T\B Min.& Median& Max& Min.& Median& Max\\
    \hline
    DOD\T &  5.96e-08 & 1.30e-03 & 6.08e-02 & 5.33e-01 & 1.01e+05 & 1.76e+06 \\
    DS\T & 6.47e-07 & 2.58e-03 & 1.23e-01 & 7.47e-06 & 1.13e+04 & 5.06e+05\\
    CDCS\T & 1.85e-08 & 7.74e-04 & 6.12e-02 & 2.87e-04 & 8.07e+03 & 6.41e+05\\
    \hline
    \end{tabular}
    \vspace{-0.1in}
\end{table*}

We also compare the loss metrics like binary cross-entropy (BCE), mean absolute error (MAE), and mean squared error (MSE) between the density values of predicted topology and the ground truth optimal topology. In addition, we also perform statistical analysis on MSE loss between predicted and actual values of both topological constraints (VF) and physics constraints (TC). We summarize the minimum, median and maximum value of MSE for 2D test data in \tabref{Tab:2dstats}.

\begin{figure*}[t!]
    \centering
    \includegraphics[width=0.78\linewidth,trim={0.75in 0.7in 1.8in 0.2in},clip]{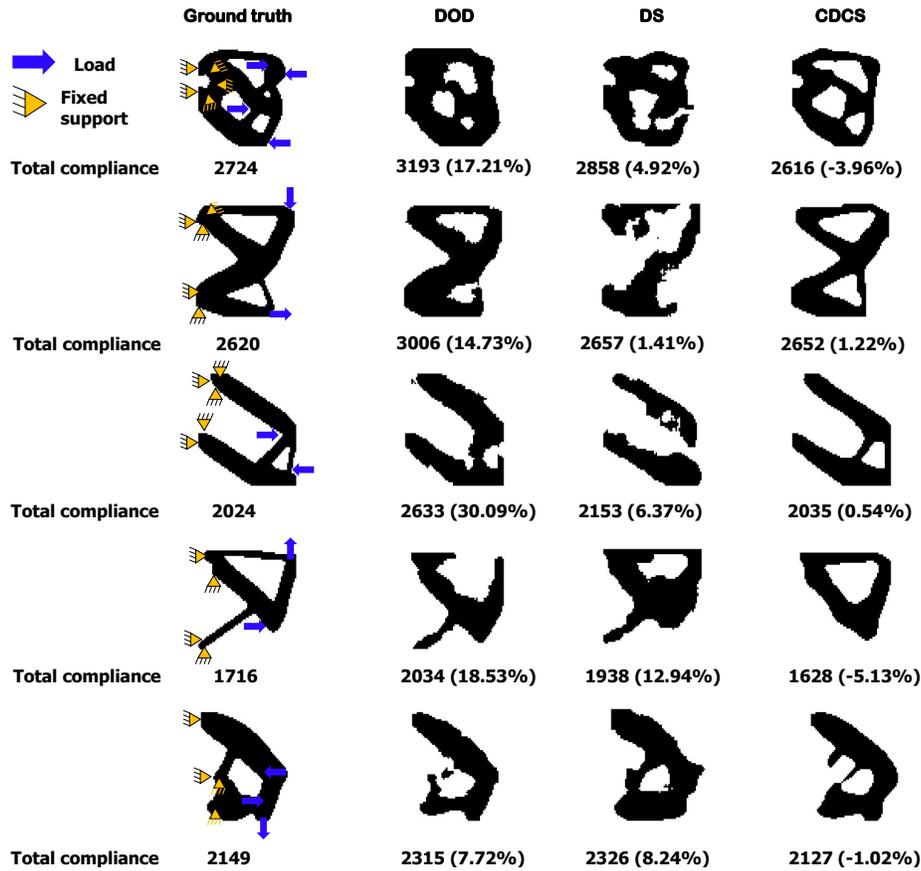}
    \caption{Visualization of test data in 2D: (i) Ground truth final topology shape with fixed supports and load locations (ii) Method 1: Baseline direct optimal density prediction, (iii) Method 2: Density sequence prediction, (iv) Method 3: Coupled density and compliance sequence prediction. The results show the target design and the predicted design.}
    \label{fig:2dvisuals}
\end{figure*}

Apart from the numerical analysis, to further qualify the performance of our method, we compare the visualizations of the predicted final topology, obtained by performing end-to-end prediction using all three methods, with ground truth final optimal topology in \figref{fig:2dvisuals}. Additionally, we compute each sample's total compliance value and compute the percentage deviation of predicted TC from simulated TC in the visualization. In the ground truth column, we also show the boundary and load conditions applied for each sample.

\begin{figure*}[t!]
    \centering
    \includegraphics[width=0.94\linewidth,trim={0.1in 1in 0.0in 0.75in},clip]{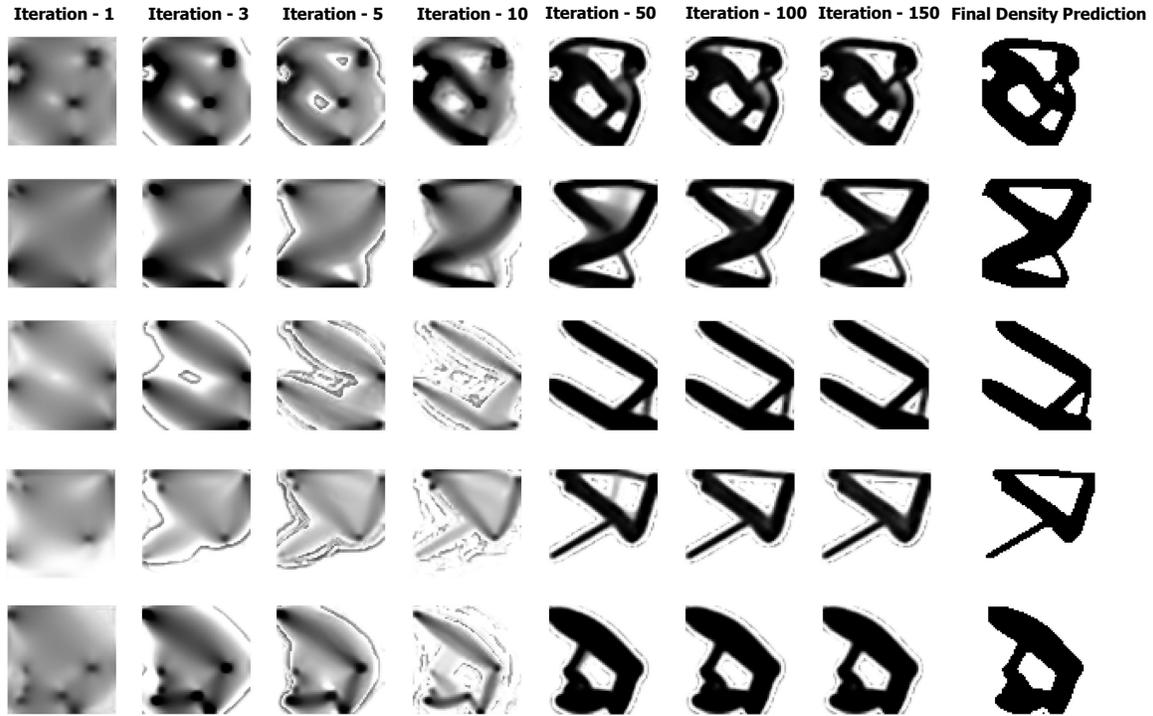}
    \caption{Visualizations of DPN predicting intermediate iterations density on 2D test data.}
    \label{fig:DPN2dintermediate}
\end{figure*}

\begin{figure*}[t!]
    \centering
    \includegraphics[width=0.94\linewidth,trim={0.1in 1in 0.0in 0.75in},clip]{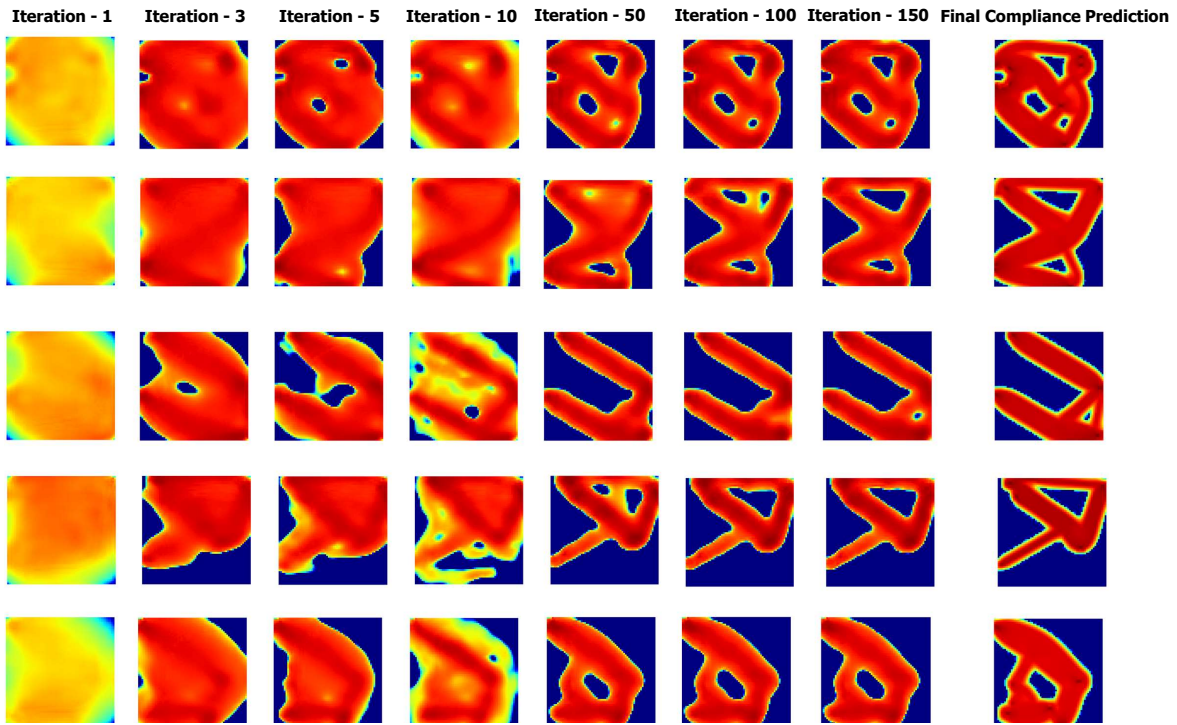}
    \caption{Visualizations of CPN predicting intermediate iterations compliance on 2D test data.}
    \label{fig:CPN2dintermediate}
\end{figure*}

We further evaluate the CDCS method by visualizing the evolution of intermediate iteration densities and elemental compliance predicted by the DPN and CPN, respectively. As discussed in \secref{sec:CDCS}, we feed the actual initial and current iteration elemental compliance and the density values to DPN and CPN to predict the next iteration quantities. We visualize this iteration-wise evolution of the topology and its compliance in \figref{fig:DPN2dintermediate} and \figref{fig:CPN2dintermediate}, respectively.

\subsection{Results on 3D Topologies}\label{sec:3dresults}

We perform a similar set of evaluations on the 3D data as discussed in \secref{sec:2dresults} to assess the performance of the CDCS method and comparing it to the DOD method. 

\begin{figure*}[t!]
    \centering
    \includegraphics[width=0.45\linewidth,trim={0in 0.0in 0.0in 0.0in},clip]{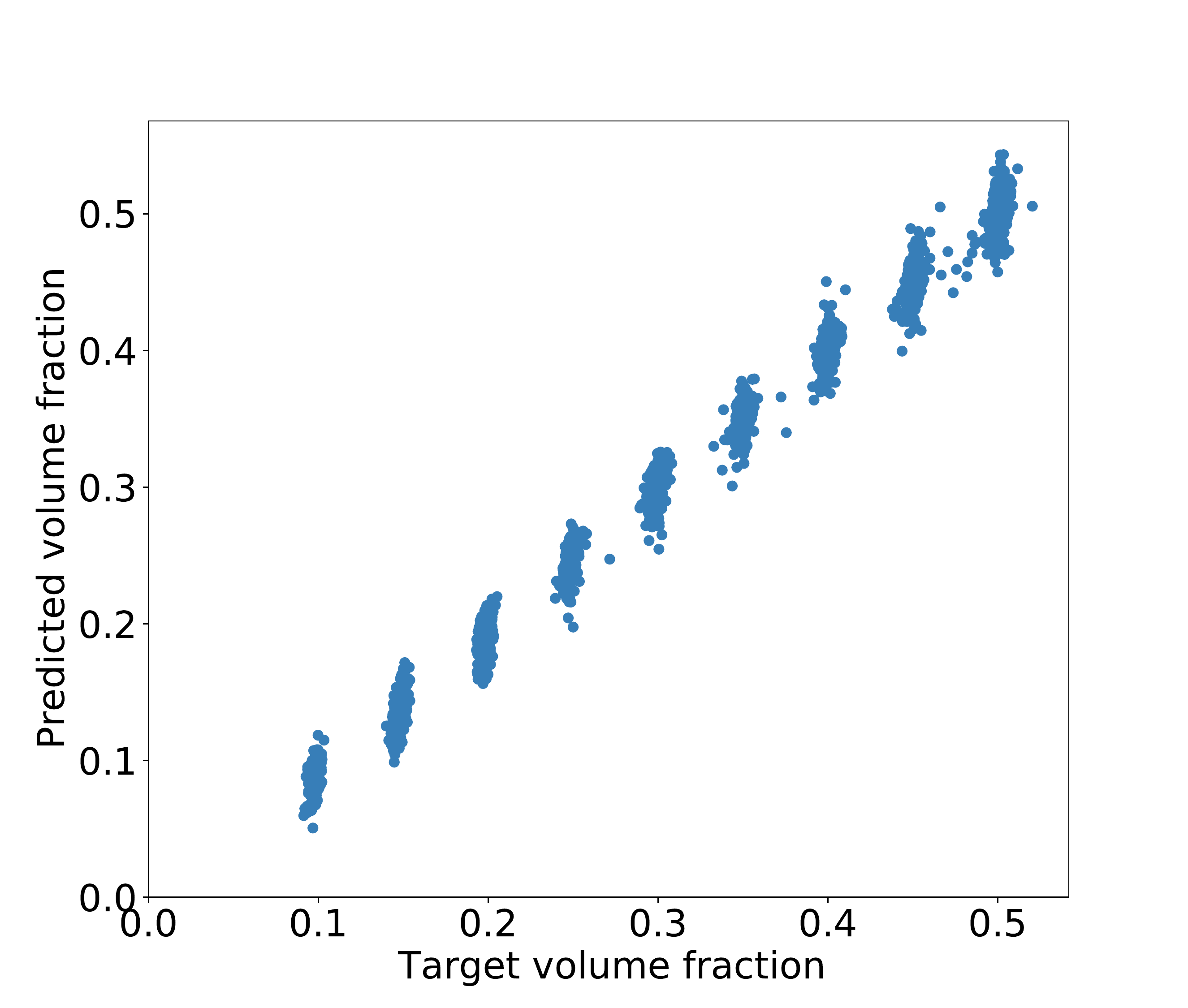}
    \includegraphics[width=0.45\linewidth,trim={0in 0.0in 0.0in 0.0in},clip]{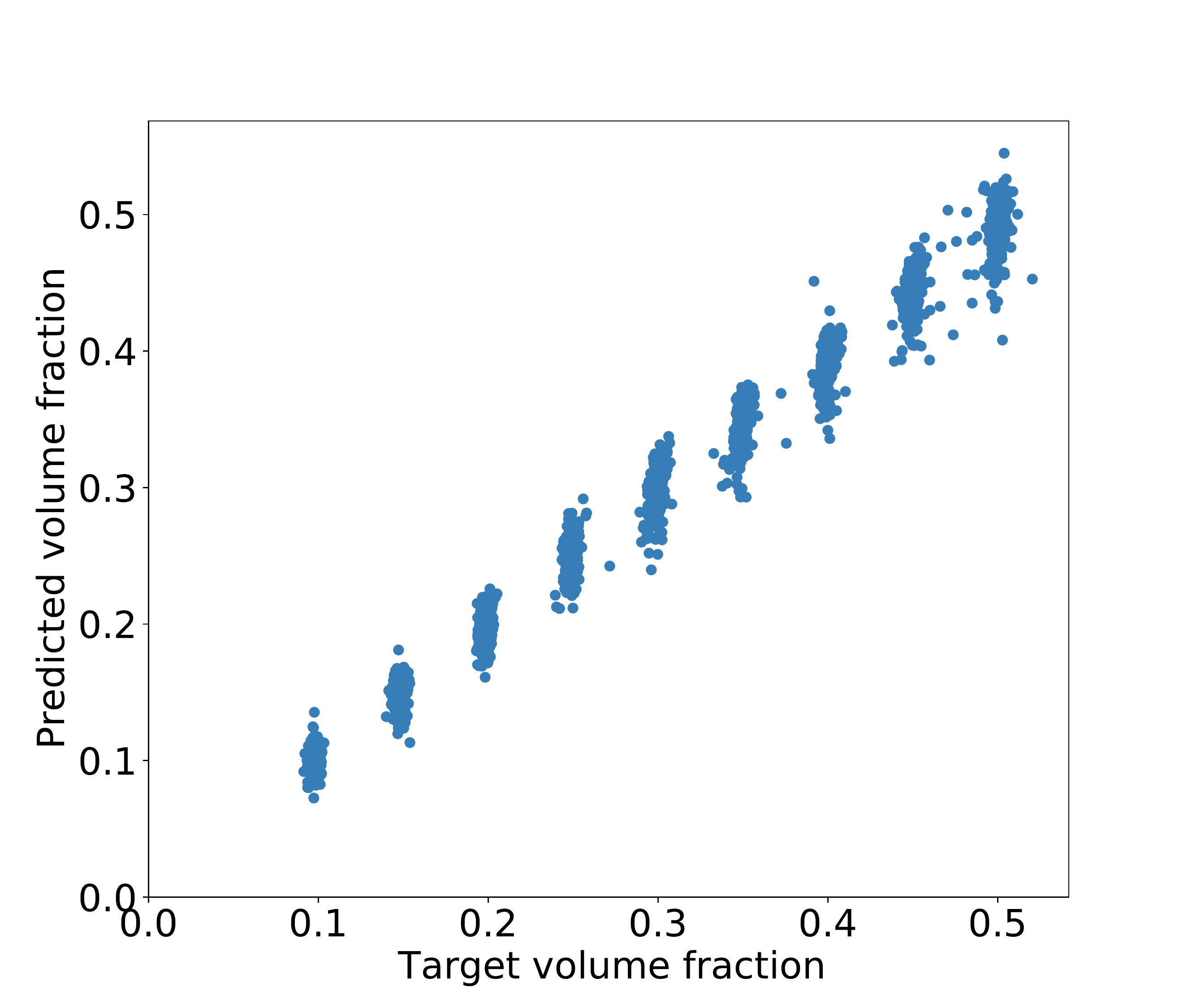}
    \caption{Correlation plots between predicted volume fraction and target volume fraction on 3D test data for DOD and CDCS.}
    \label{fig:3D_vf_correlation} 
    \vspace{0.2in}
\end{figure*}

\begin{figure*}[t!]
    \centering
    \includegraphics[width=0.45\linewidth,trim={0in 0.0in 0.0in 0.0in},clip]{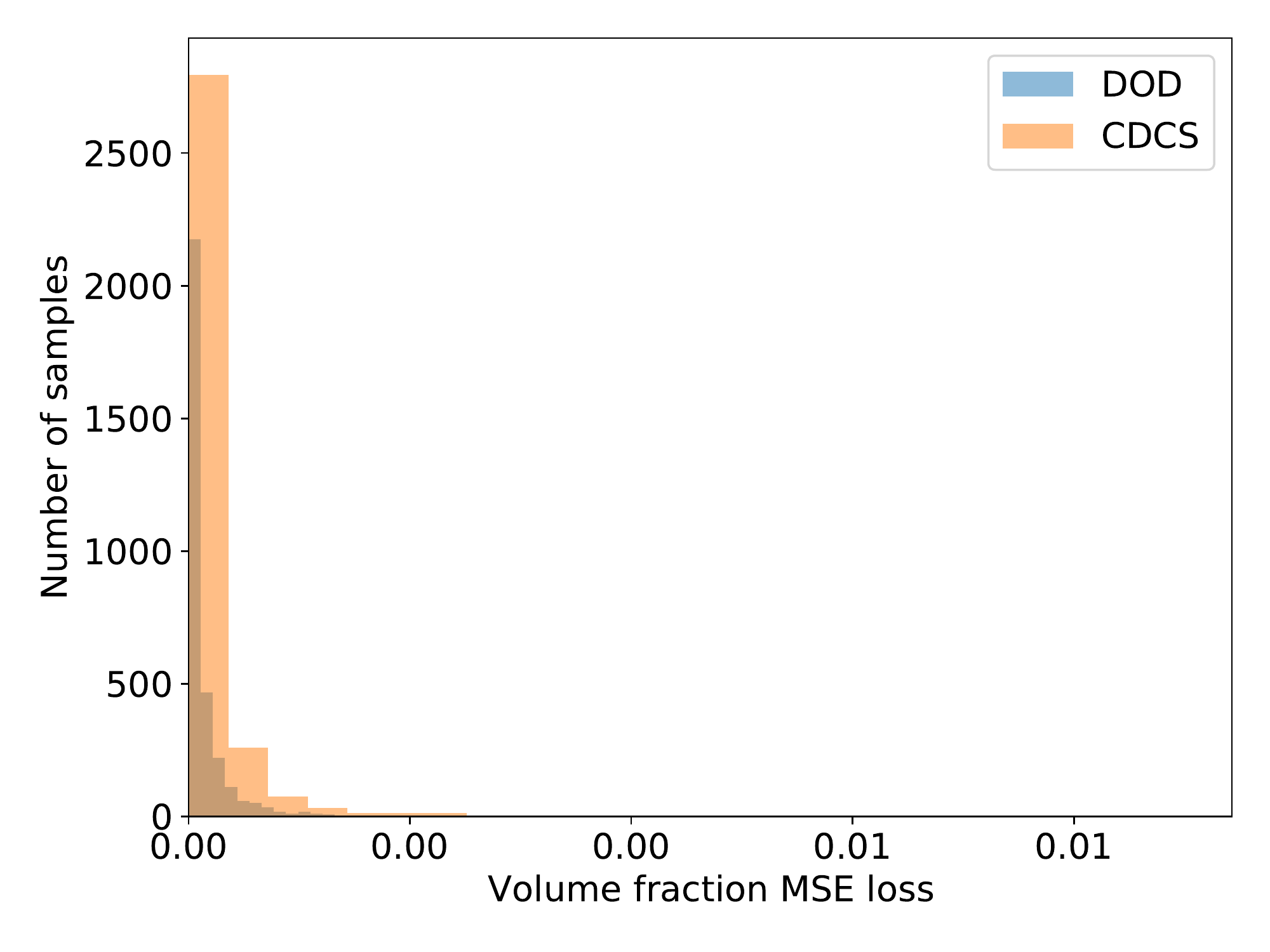}
    \includegraphics[width=0.45\linewidth,trim={0in 0.0in 0.0in 0.0in},clip]{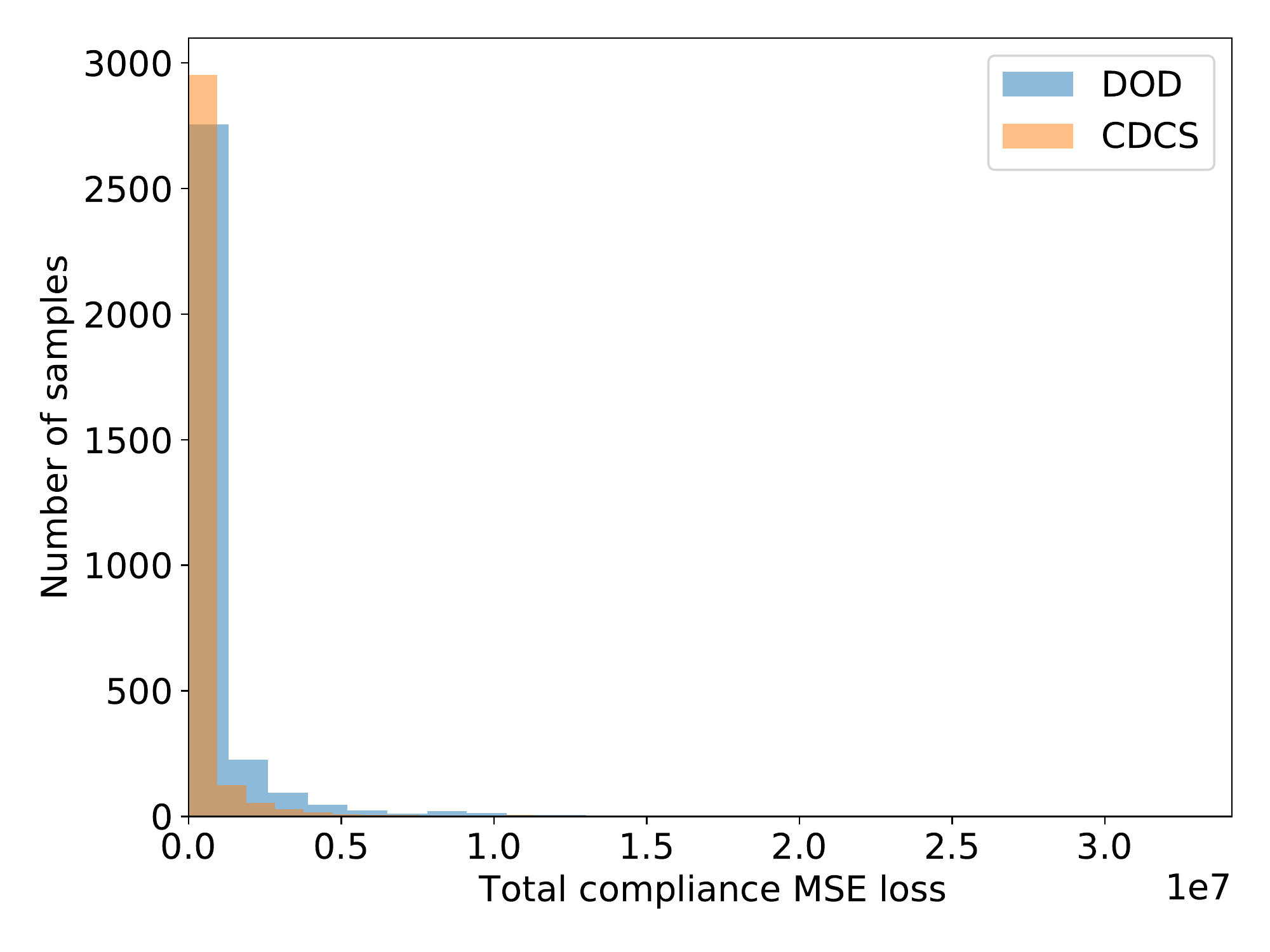}
    \caption{Histogram of (a) total volume fraction loss and (b) total compliance loss on the 3D test data.}
    \label{fig:3d_histograms}
    \vspace{0.2in}
\end{figure*}

\begin{figure*}[t!]
    \centering
    \includegraphics[width=0.42\linewidth,trim={0in 0.0in 0.5in 0.0in},clip]{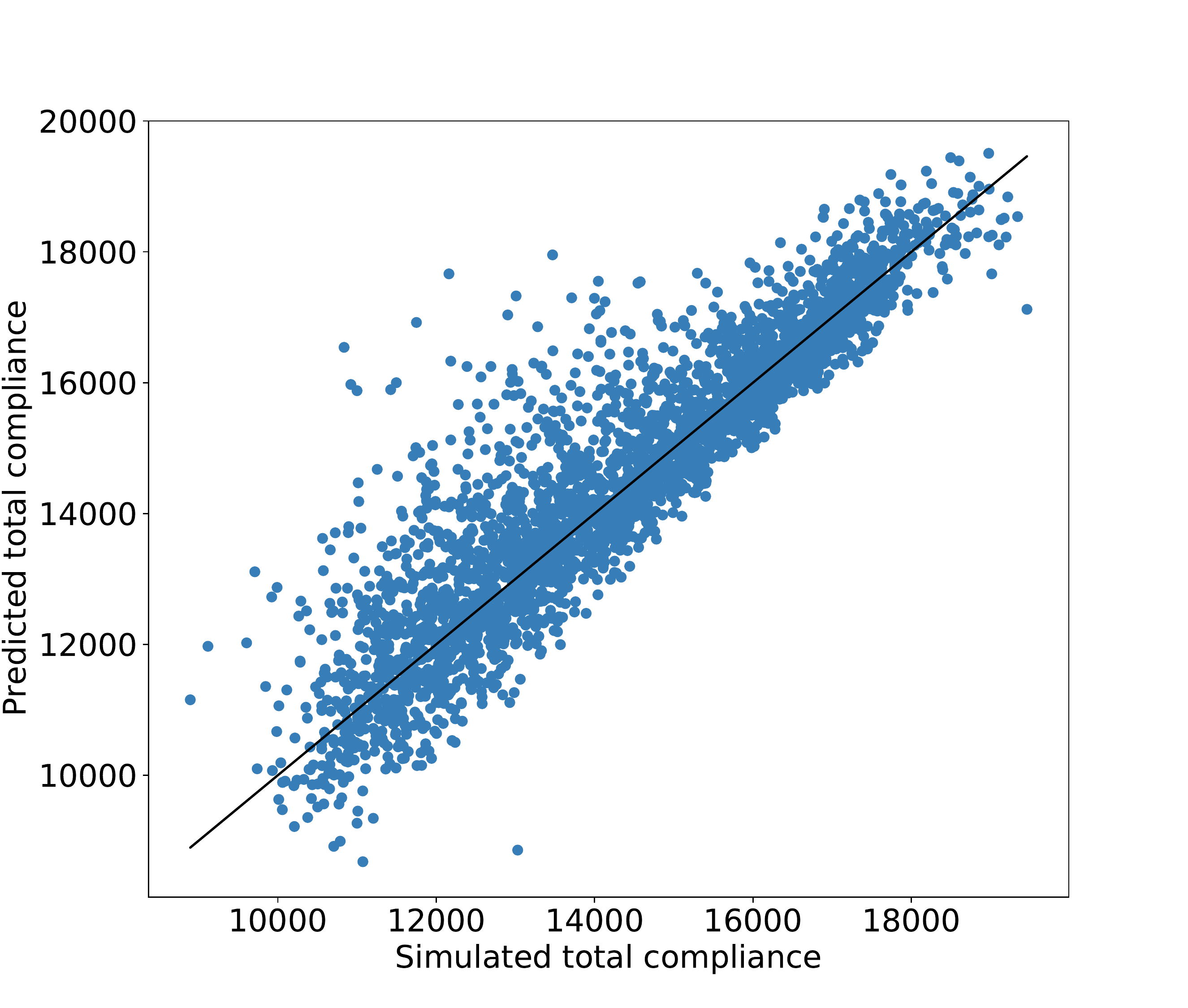}
    \includegraphics[width=0.42\linewidth,trim={0in 0.0in 0.5in 0.0in},clip]{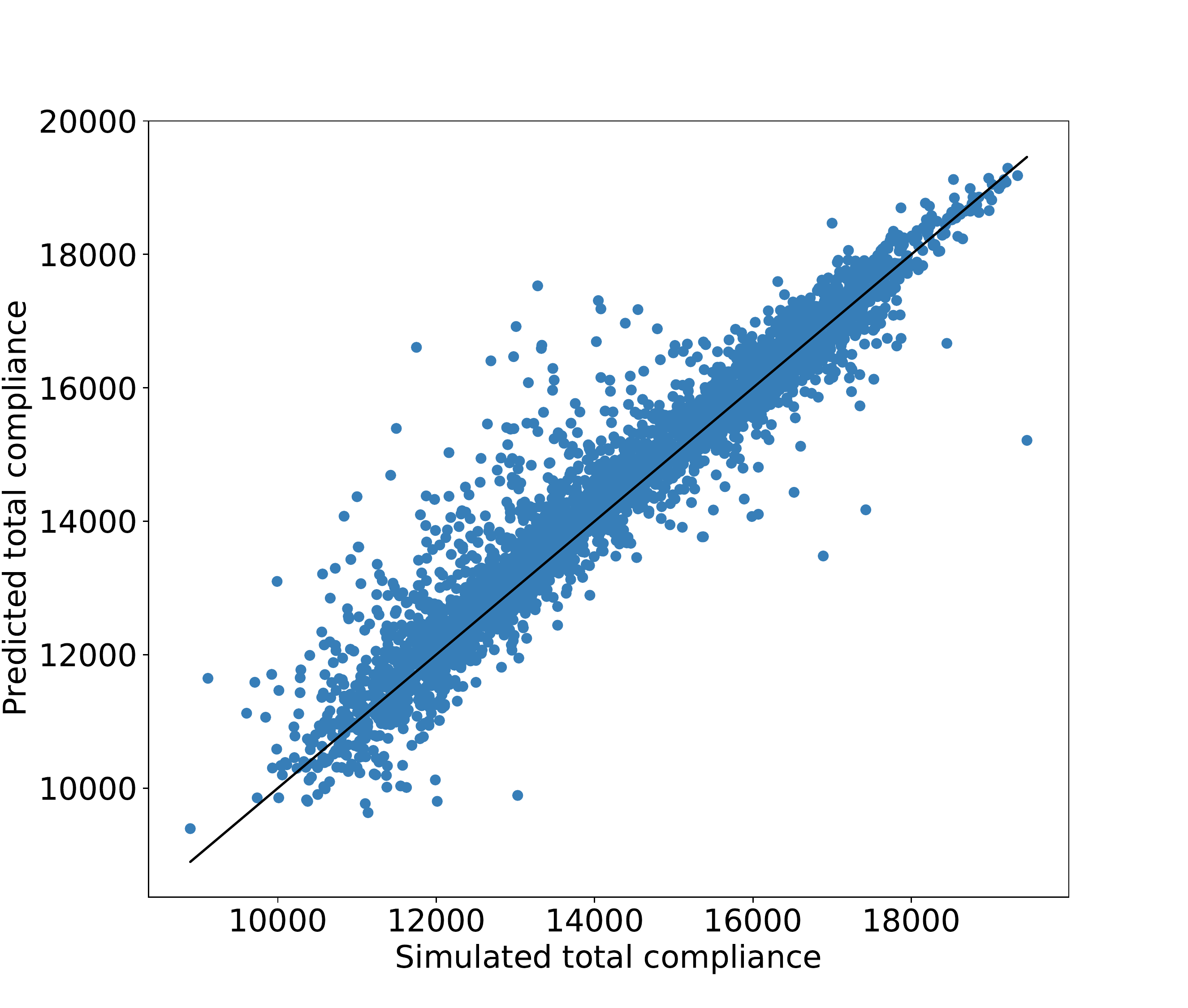}
    \caption{Correlation plots between predicted total compliance and simulated total compliance on 3D test data for DOD and CDCS.}
    \label{fig:3D_comp_correlation} 
\end{figure*}

Comparing the MSE between the volume fraction (VF) and the total compliance (TC) of predicted topology with actual final topology, in the \tabref{tab:compare_3d}. We plot the histogram of MSE values from DOD and CDCS for both, VF and TC, in \figref{fig:3d_histograms}. We also compute the Pearson's correlation coefficient for the VF and TC values between the predicted and actual final topology using both DOD and CDCS in \tabref{tab:correlationcoeff_3d} and correlation plots in \figref{fig:3D_vf_correlation} and \figref{fig:3D_comp_correlation}, respectively.

We have summarized the different loss metrics like BCE, MAE, and MSE between the predicted and actual topology for both CDCS and DOD in \tabref{tab:compare_3d}. In \tabref{Tab:3dstats}, we summarize the statistical analysis of the MSE value of both VF and TC on the 3D test data.

\begin{table}[t!]
    \setlength\extrarowheight{5pt}
    \centering
    \small
    \caption{Comparison of test loss metrics using DOD and CDCS on 3D test data.}
    \label{tab:compare_3d}
    \newcommand{\tabincell}[2]{\begin{tabular}{@{}#1@{}}#2\end{tabular}}
    \begin{tabular}{| r | r | r | r | r | r |}
    \hline
    \multirow{2}{*}{\textbf{Method}}  &  \textbf{VF} & \textbf{TC} &  \multicolumn{3}{|c|}{\textbf{Density}} \\ \cline{2-6}
      &  \textbf{MSE} & \textbf{MSE} &  \textbf{BCE} & \textbf{MAE} & \textbf{MSE} \\
    \hline 
    DOD & 0.0002 & 8.04e+05 & \textbf{0.1008} & \textbf{0.0648} & \textbf{0.0312}\\
    CDCS & \textbf{0.0001} & \textbf{3.95e+05} &  0.1965 & 0.0875 & 0.0544\\
    \hline
    \end{tabular}
\end{table}

\begin{table}[t!]
    \setlength\extrarowheight{5pt}
    \small
    \caption{Comparison of correlation coefficient(R) for volume fraction and total compliance on on 3D test data.}
    \label{tab:correlationcoeff_3d}
    \newcommand\T{\rule{0pt}{2.7ex}}
    \newcommand\B{\rule[-1.3ex]{0pt}{0pt}}
    \newcommand{\tabincell}[2]{\begin{tabular}{@{}#1@{}}#2\end{tabular}}
    \centering
    \begin{tabular}{| r | l | l |}
    \hline
            \textbf{Method}  &  \textbf{R for volume fraction} & \textbf{R for total compliance}\\
    \hline 
    DOD & 0.9966 & 0.9139 \\
    \hline
    CDCS & 0.9947 & 0.9578\\
    \hline
    \end{tabular}
\end{table}

\begin{table*}[t!]
    \newcommand\T{\rule{0pt}{2.7ex}}
    \newcommand\B{\rule[-1.3ex]{0pt}{0pt}}
    \centering
    \small
    \caption{Statistics on the volume fraction and total compliance loss on 3D test data.}
    \label{Tab:3dstats}
    \begin{tabular}{|c|c|c|c|c|c|c|}
    \hline
     \textbf{Statistics} & \multicolumn{3}{c|}{\T\B \textbf{MSE of volume fraction}} & \multicolumn{3}{c|}{ \textbf{MSE of total compliance}}\\
     \cline{1-7}
    \textbf{Method} &\T\B Min.& Median& Max& Min.& Median& Max\\
    \hline
    DOD\T &  9.31e-10 & 4.80e-05 & 2.75e-02 & 4.64e-01 & 1.83e+05 & 3.26e+07 \\
    CDCS\T & 9.31e-10 & 6.19e-05 & 8.97e-03 & 1.12e-01 & 7.63e+04 & 2.35e+06 \\
    \hline
    \end{tabular}
\end{table*}

\begin{table*}[t!]
    \newcommand\T{\rule{0pt}{2.7ex}}
    \newcommand\B{\rule[-1.3ex]{0pt}{0pt}}
    \centering
    \small
    \caption{Comparison of different neural network architectures for each task of CDCS on 3D test data.}
    \label{Tab:3d_arch_compare}
    \begin{tabular}{|c|c|c|c|c|c|c|c|c|}
    \hline
     \textbf{Method}\B\T & \multicolumn{2}{c|}{\T\B \textbf{CPN}} & \multicolumn{3}{c|}{ \textbf{DPN}} & \multicolumn{3}{c|}{\textbf{FDPN}} \\
     \cline{1-9}
    \textbf{Architecture} &\T\B MAE & MSE & BCE & MAE & MSE & BCE & MAE & MSE\\
    \hline
    AE\T &  0.0221 & 0.0009 & 0.2838 & 0.0211 & 0.0014 & 0.1140 & 0.0178 & 0.0026 \\
    U-Net\T & 0.0286 & 0.0013 & 0.2810 & 0.0144 & 0.0005 & 0.1152 & 0.0145 & 0.0019 \\
    U-SE-ResNet\T & 0.0294 & 0.0016 & 0.3157 & 0.0131 & 0.0006 & 0.1188 & 0.0155 & 0.0021\\
    \hline
    \end{tabular}
\end{table*}

\begin{table}[t!]
    \setlength\extrarowheight{5pt}
    \centering
    \small
    \caption{Comparison of average time between traditional SIMP algorithm and deep-learning based DOD and CDCS to obtain one optimized 3D topology.}
    \label{tab:compare_time_3d}
    \newcommand{\tabincell}[2]{\begin{tabular}{@{}#1@{}}#2\end{tabular}}
    \begin{tabular}{| c | c |}
    \hline
            \textbf{Method}  & \textbf{Time (sec)} \\
    \hline 
    SIMP & 390 \\
    DLTO-DOD &  0.233\\
    DLTO-CDCS & 0.102 \\
    \hline
    \end{tabular}
\end{table}

As discussed earlier, the CDCS method has three different neural networks dedicated to learning the different aspects of structural topology optimization. We have compared the performance of the different architectures for each task of TO. We have experimented with three architectures, which are: (i) Encoder-Decoder architecture, (ii) U-Net~\citep{ronneberger2015u,cciccek20163d} architecture, and (iii) U-SE-ResNet~\citep{nie2020topologygan}. For CPN, we compare MAE and MSE metrics, while for DPN and FDPN, we evaluate the performance based on the BCE, MAE, and MSE values on 3D test data. All these metric values are summarized in \tabref{Tab:3d_arch_compare}. From the \tabref{Tab:3d_arch_compare} we selected the best of three for each task, like for CPN, we implemented Encoder-Decoder, for DPN used U-SE-ResNet, and similarly, for FDPN, we used U-Net architectures.

\begin{figure*}[t!]
    \centering
    \includegraphics[width=0.85\linewidth,trim={1.75in 0in 1.5in 0in},clip]{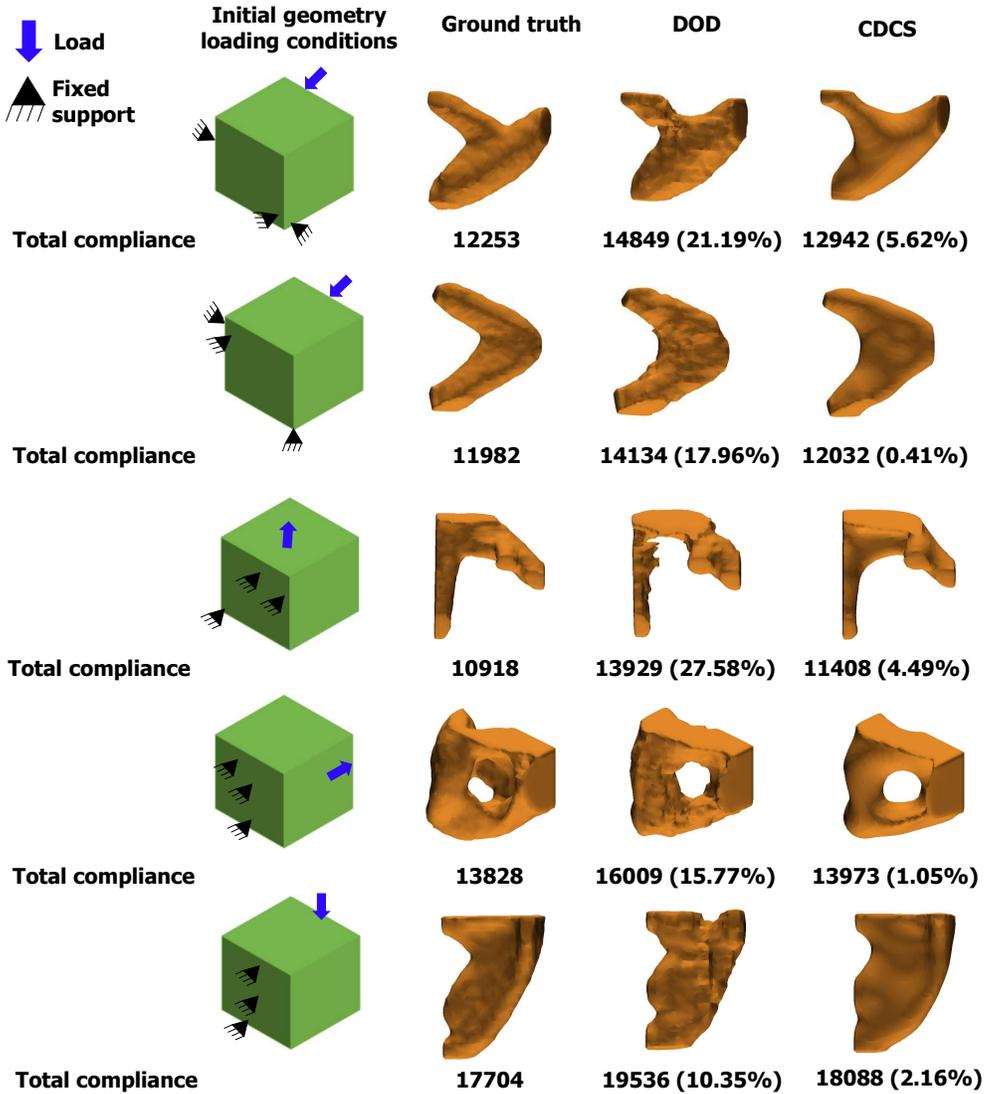}
    \caption{Visualization of In-distribution test data in 3D: (i) Initial geometry with the fixed supports and load locations, (ii) Ground Truth final topology (iii) Method 1: Baseline direct optimal density prediction, (iii) Method 3: Coupled density and compliance sequence prediction. The results show the target shape and the predicted shape.}
    \label{fig:3dvisuals_insideDistribution}
\end{figure*}

\begin{figure*}[t!]
    \centering
    \includegraphics[width=0.85\linewidth,trim={1.75in 0in 1.5in 0in},clip]{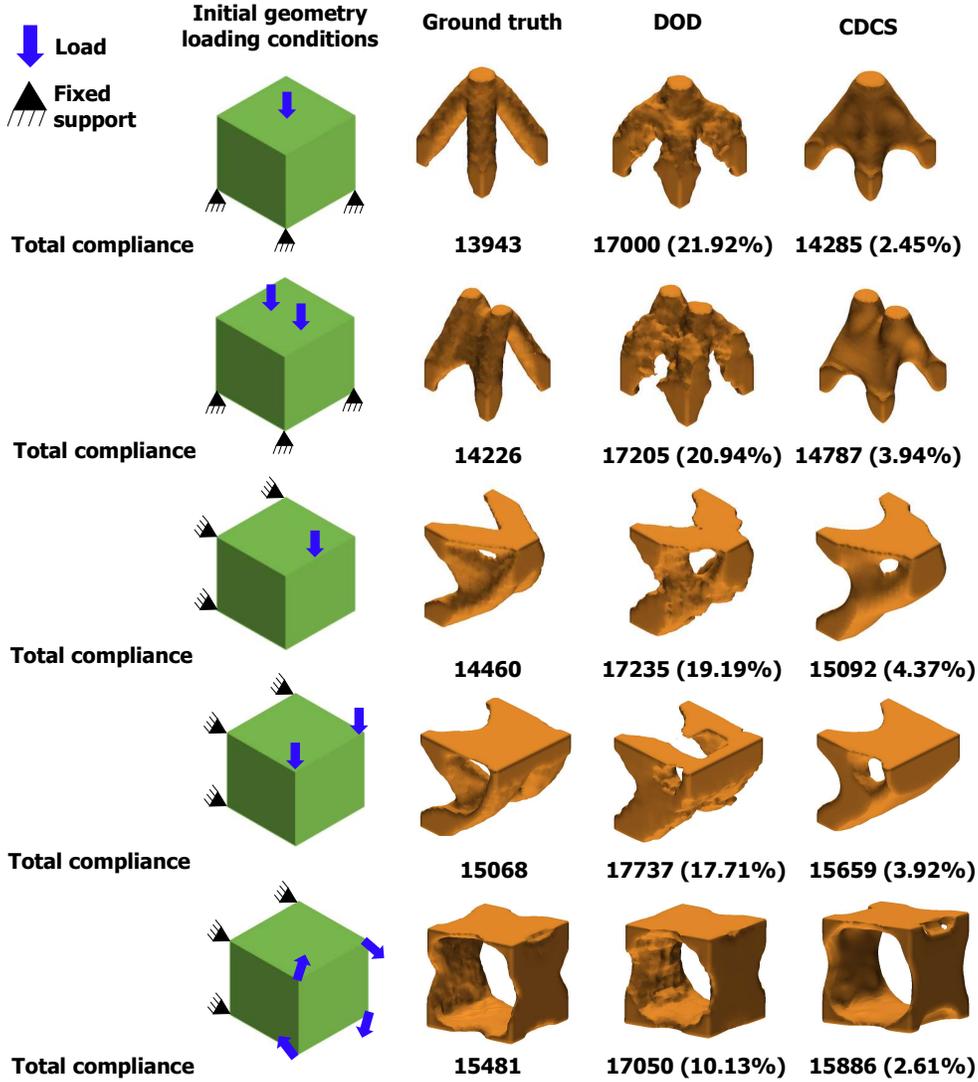}
    \caption{Visualization of Out-distribution test data in 3D: (i) Initial geometry with the fixed supports and load locations, (ii) Ground Truth final topology (iii) Method 1: Baseline direct optimal density prediction, (iii) Method 3: Coupled density and compliance sequence prediction. The results show the target shape and the predicted shape.}
    \label{fig:3dvisuals_outsideDistribution}
\end{figure*}

We use marching cube methods to visualize the predicted and actual optimal topology shapes in 3D. As mentioned earlier in \secref{sec:2dresults}, using the end-to-end prediction, we obtain the predicted final topology. We visualize some samples from the test data, which are in-distribution samples and some out-of-distribution samples. As discussed in \secref{sec:3Ddatagen} about the 3D data generation, the in-distribution dataset has three nodes with fixed support and one loading condition. On the other hand, we generated few samples with more than three fixed support locations and multiple loads such as four loads (torsional deformation) acting on the topology; we termed these samples as out-of-distribution samples. We visualize some samples from the in-distribution test data in \figref{fig:3dvisuals_insideDistribution} and the out-of-distribution samples in \figref{fig:3dvisuals_outsideDistribution}. The first column shows the initial geometry, fixed support locations, and load in both figures, giving an approximate idea of the final optimal topology. We also calculate the total compliance (TC) value and the percentage deviation of predicted TC from simulated TC for each sample and mention it under each topology.

\section{Discussion}

In terms of the volume fraction (VF) constraint, if we compare the histogram in \figref{fig:2d_histograms}(a) and the MSE values from \tabref{tab:compare_2d}, the CDCS method performs comparably to the DOD method and marginally better than the DS method. We observe comparable values of Pearson's correlation coefficient for DOD ($R=0.8986$) and CDCS ($R=0.8945$) method, while the DS ($R=0.6883$) method performs poorly in satisfying the VF constraint. From \tabref{tab:compare_2d}, we observe almost $10\times$ times lesser error value when we compare the MSE between the predicted and the simulated total compliance for both CDCS and DS  with the DOD. The histogram in \figref{fig:2d_histograms}(b) shows that the CDCS has the maximum number of test samples with an MSE value closer to zero. We can see that both methods, CDCS and DS, predict the TC very close to the predicted optimal TC minimum value. From \tabref{tab:correlationcoeff_2d}, total compliance of predicted topology by DS ($R=0.9551$) and CDCS ($R=0.8926$) is highly correlated with simulated total compliance than the baseline DOD ($R=0.8403$) approach. The CDCS and DOD method's performance is comparable in terms of the different loss metrics used and is better than the DS method, as shown in \tabref{tab:compare_2d}. In \tabref{Tab:2dstats}, we see that all the three metrics listed have comparable values for DOD and CDCS and are slightly better than the DS for volume fraction constraint. For the MSE values of total compliance, for 2D data, we see that both DS and CDCS perform much better than the DOD in all three statistics, and DS and CDCS have comparable median and maximum values. However, DS has a minimum loss value in all three methods. 
 
\figref{fig:2dvisuals} shows that the CDCS predicts the final shape significantly closer to the ground truth, and also, the predicted total compliance value is much closer to the actual value. Although there are some cases where the shape predicted by DOD and DS is slightly better than the CDCS, the predicted total compliance value is much higher than the actual value. To further evaluate the CDCS method, from the visualizations in \figref{fig:DPN2dintermediate} and \figref{fig:CPN2dintermediate}, it is evident that both DPN and CPN are efficient at predicting the next iteration density and compliance values. Also, it depicts the non-trivial transformation flow of the initial topology shape towards the final optimal shape.

The results show that the CDCS method performs better than the baseline DOD and the DS method on 2D topologies. Although DS satisfies the physics constraint (minimizing the TC) better, it does not satisfy the topological constraint (VF) to the same extent. On the other hand, CDCS accomplishes the best balance in satisfying the volume fraction constraint and achieving a total compliance value close to the actual optimal minimum value. Hence, we only extend the CDCS method to the 3D dataset and compare it with the baseline DOD method.

From \tabref{tab:compare_3d}, we see that the MSE of VF using the CDCS method is $2\times$ lower than using the DOD method. From the histogram plots in \figref{fig:3d_histograms}(a), we can infer that more samples have minimum MSE of VF using CDCS than the DOD. Comparing correlation plots in \figref{fig:3D_vf_correlation} and Pearson's correlation coefficient from \tabref{tab:correlationcoeff_3d}, we notice that both predicted and actual VF values are highly correlated. Comparing the MSE of total compliance(TC), DOD has $2\times$ more error value than the CDCS, and from the histogram in \figref{fig:3d_histograms}(b), we see that most of the samples have the lower MSE of TC value using CDCS. In \tabref{tab:correlationcoeff_3d},we observe that the TC value predicted by the CDCS ($R=0.9578$) are highly correlated to actual TC values than the DOD ($R=0.9139$). We also observe this high correlation when we plot the correlation plots in \figref{fig:3D_comp_correlation}. In terms of MAE and MSE, both CDCS and DOD are comparable, while DOD performs slightly better when comparing the BCE values from \tabref{tab:compare_3d}. But overall, like in the case of 2D, CDCS achieves the balance of satisfying both topological (VF) and physical (TC) constraints on the 3D dataset. \tabref{Tab:3dstats} shows that all the three metrics values are comparable in the case of MSE of VF. We see better performance when we consider the MSE of TC. We notice that the median value using CDCS is $2\times$ lower than the DOD method. Also, the maximum value of MSE of TC using CDCS is $15\times$ smaller than the DOD value, which affirms the greater performance of CDCS over the baseline DOD approach in satisfying the physics constraint (TC).

Apart from the numerical analysis, we notice that the CDCS method performs significantly better than the baseline DOD method when we visualize the obtained shapes of test samples from in-distribution test data in \figref{fig:3dvisuals_insideDistribution}. Even on out-of-distribution samples, which have more fixed supports and loads, the CDCS method predicts the shape of final topology much closer to actual shape than DOD. The shapes are smoother than the actual ground truth obtained by CDCS. We also observe that the TC value of topology predicted using CDCS is very close to the simulated TC value than using the DOD. 

From the numerical analysis performed and supported by the visualizations on both 2D and 3D datasets, we claim that the performance of the CDCS is better than the baseline DOD and DS. With the multiple network setup proposed to learn different steps in SIMP, our method predicts the final optimal topology and its compliance closer to the topology simulated by SIMP. Additionally, with DPN and CPN, our method can predict intermediate densities and compliances values, respectively. Also, with the deep-learning-based methods proposed, we can perform topology optimization significantly faster, and the speedup is approximately by 3900$\times$ (see \tabref{tab:compare_time_3d}).

To get more insights on the network architectures of CPN, DPN, and FDPN used in CDCS, please refer to \appref{sec:appendix_architecture}. For training performance results such as loss curves and histograms of different loss metrics, please see \appref{sec:appendix_lossplots}.

\subsection{Limitations}

While we show better performance of CDCS over the baseline method and significant computational speedup over traditional SIMP-based topology optimization approaches, our proposed approach has some limitations. The primary limitation is that our framework considers the initial geometry of a solid cube, and then different loads and boundary conditions are applied. Naturally, more general designs would not start with a cube's initial geometry but a more generic geometry. This issue can be addressed by adding more data to the current dataset with diverse examples with different initial geometries. Further, to capture key features in the initial and final geometry, we need to extend this framework to voxel resolutions beyond $32\times32\times32$. Finally, the limitation of the data requirements for training is fundamental to this approach. Here, we would like to note that our approach is amortized over the number of inferences we would be performing. Our approach is beneficial if the number of inferences is an order of magnitude higher than the number of samples generated.

\subsection{Future Work}
Future work includes 3D topology optimization performed on a generic 3D CAD model. Further, extending our framework to higher resolutions such as $128\times128\times128$ would be useful for more realistic designs with intricate features. Another avenue of future work is adding manufacturability constraints on the fly during inference and the capability of generative design. Finally, approaches to reduce the data requirements for training using information from structural mechanics as priors would be an interesting direction to explore.

\section{Conclusions}\label{sec:conclusions}
In this paper, we explore the application of algorithmically consistent deep learning methods for structural topology optimization. We developed two approaches (density sequence and coupled density compliance sequence models), consistent with the physics constraints, topological constraints, and the SIMP topological optimization algorithm. We generated datasets for topology optimization in both 2D and 3D representations and then demonstrated the superior performance of our proposed approach over a direct density-based baseline approach. Finally, we visualize a few anecdotal topology optimization samples to visually compare the three methods with the SIMP-based topology optimization process. We believe that our proposed algorithmically consistent approach for topology optimization provides superior quality results and can considerably speed up the topology optimization process over existing finite-element-based approaches.

{
\small
\section*{Acknowledgement}
This work was partly supported by the National Science Foundation under grant CMMI-1644441 and ANSYS Inc. We would also like to thank NVIDIA\textsuperscript{\textregistered} for providing GPUs used for testing the algorithms developed during this research. This work used the Extreme Science and Engineering Discovery Environment (XSEDE), supported by NSF grant ACI-1548562, and the Bridges system supported by NSF grant ACI-1445606 Pittsburgh Supercomputing Center (PSC).
}

\bibliographystyle{elsarticle-harv}
\bibliography{TopOptSubLib}

\clearpage
\appendix
\section*{Appendix}
Here, we will discuss more details of the methodology we used in this paper along with few mathematical definitions and algorithms.

\section{Training Details}\label{sec:appendix_training}
We begin by explaining the training procedure we used for training the different networks for building the three frameworks.

\begin{algorithm}[b!]
  \caption{Training Algorithm}
  \label{alg:training}
  \SetKwInOut{Input}{Input}
  \SetKwInOut{Output}{Output}

  \Input{~Network Architecture}
  \textbf{Initialize:}~\text{Weights for all layers, $W_l, (l=1,2,\dots, m$);}\\\text{patience = 0}\\
  \textbf{Load Data:}~\text{Load training data $\mathcal{\textbf{D}}$ and validation data $\mathcal{\textbf{D}}_V$}\\
  \For{($i=0;i\le num\_epochs; i++$)}
  {
      \text{Randomly shuffle the data}\\
      \text{Split $\mathcal{\textbf{D}}$ to $\mathcal{\textbf{D}}_j, (j=1,2,\dots, n$) mini-batches}\\
      \For{$j=1:n$}
      {
           \text{Predict outputs $\mathcal{O}_j$ for mini-batch $\mathcal{\textbf{D}}_j$}\\
           \text{Compute loss $\mathcal{L}(\mathcal{\textbf{D}}_j,\mathcal{O}_j,\{W\})$}\\
           \text{Update weights,$\{W\}$ using Adam optimizer}\\
      }
      \text{Predict validation outputs $\mathcal{O}_V$ for $\mathcal{\textbf{D}}_V$}\\
      \text{Compute Validation Loss $\mathcal{L}(\mathcal{\textbf{D}}_V,\mathcal{O}_V,\{W\})$}\\
      \If{\text{Avg. Validation Loss not improving}}
      {
        \text{increment patience}
      }
      \Else{\text{patience = 0}}
      \If{patience $\geq$ 30}
      {
        Exit\\
      }
    }
\end{algorithm}

\coloredref{Algorithm}{alg:training} provides a general training procedure. We first load the training dataset and the validation data. For several epochs and several mini-batches of the dataset, we compute the loss and update weights using SGD (and its variants, such as Adam) optimizer. In general, we save the weights of the model with the least validation loss. We run for 50 additional epochs (patience parameter) to check if the loss reduces further. In the \coloredref{Figure}{fig:DSlosses_2d}, we see that at the $450^{th}$ epoch, the training, and validation loss are very close to each other, and validation loss is minimum. Yet, we still run for 50 more epochs to ensure that the weights obtained are truly minimal and have good generalization capability. Finally, we stop at the end of the 485th epoch because we do not find any better weights with a lower validation loss.

\section{Metrics used for comparison}
To develop a baseline for our CDCS framework and further understand the different elements, we use several statistical metrics that we will detail in this section. 

Two major metrics used while comparing the results are: (i) the mean-squared error (MSE) and (ii) the correlation coefficient (R). The mean-squared error is computed by:$$MSE = {\frac{\sum_i^k (p_1 - p_2)^2}{N}}$$ where $p_1$ and $p_2$ are two data scalar values to be compared. In the case of vector comparison, the $MSE$ is also the $L_2$ norm of the difference vector. Similarly, the mean absolute error($MAE$) is defined as the $L_1$ norm of the difference vector. For scalar values, $$MAE = {\frac{\sum_i^k |p_1 - p_2|}{N}}$$

The correlation coefficient (more popularly known as Pearson Correlation Coefficient, $R$), which was used for the comparison of the results is given by: $$R = \frac{cov(x,y)}{\sigma_x\sigma_y},$$ where $cov(x,y)$ is the covariance between $x$ and $y$ and $\sigma$ is the standard deviation. A simpler formula used for computing the correlation coefficient is as follows:
$$R = \frac{\sum{(x- m_x)(y - m_y)}}{\sqrt{\sum(x- m_x)^2\sum{(y - m_y)^2}}}.$$
Here $m_x$ and $m_y$ represent the mean of vectors $x$ and $y$.

Apart from these metrics, another important metric we use is the binary crossentropy (BCE) loss. Binary crossentropy loss denotes the log likelihood of the predicted value for the target value. Mathematically, $$BCE = \frac{\sum_i^k |p_1\times log(1-p_2) + p_2 \times log(1-p_1)|}{N}$$

Finally, we use Accuracy to count the number of pixels/voxels accurately classified. For this, we threshold the density values predicted by $0.5$ and count the voxels classified correctly.

\section{Data convergence study}
We conducted a data convergence study to confirm the sufficiency of the dataset required for performing all the experiments mentioned in the main paper. We use a separate test dataset and train the model with different training samples to perform these experiments. The metrics reported in \tabref{tab:dataconvergence} are obtained by evaluating the trained networks on the separate test dataset. For the sake of brevity, we show this only for the 3D dataset and baseline DOD framework alone. We observe that the performance of the network increases with the increase in the number of samples used for training. However, this saturates after 5000 samples with very little improvement obtained with 10000 samples. This demonstrates the sufficiency of the dataset. 

\begin{table*}[h!]
    \setlength\extrarowheight{5pt}
    \newcommand{\tabincell}[2]{\begin{tabular}{@{}#1@{}}#2\end{tabular}}
    \centering
    \caption{Data convergence study using DOD on 3D test data.}
    \label{tab:dataconvergence}
    \begin{tabular}{| c | c | c | c | c | c | c |}
    \hline
            \textbf{Number of samples}  &  \textbf{MSE of VF} & \textbf{MSE of TC} & \textbf{Accuracy} & \textbf{BCE} & \textbf{MAE} & \textbf{MSE} \\
    \hline 
    500 & 0.0010 & 4.76e+06 & 91.90\% & 0.1832 & 0.1185 & 0.0570 \\
    1000 & 0.0004 & 3.07e+06 & 93.47\% & 0.1482 & 0.0943 & 0.0461 \\
    2500 & 0.0002 & 1.39e+06 & 94.57\% & 0.1241 & 0.0782 & 0.0385 \\
    5000 & 0.0002 & 9.82e+05 & 95.38\% & 0.1058 & 0.0672 & 0.0328 \\
    10000 & 0.0002 & 8.04e+05 & 95.61\% & 0.1008 & 0.0648 & 0.0312 \\
    \hline
    \end{tabular}
\end{table*}

\clearpage

\section{Performance plots and Histograms}\label{sec:appendix_lossplots}

\begin{figure*}[h!]
    \centering
    \includegraphics[width=0.4\linewidth,trim={0in 0.0in 0.0in 0.0in},clip]{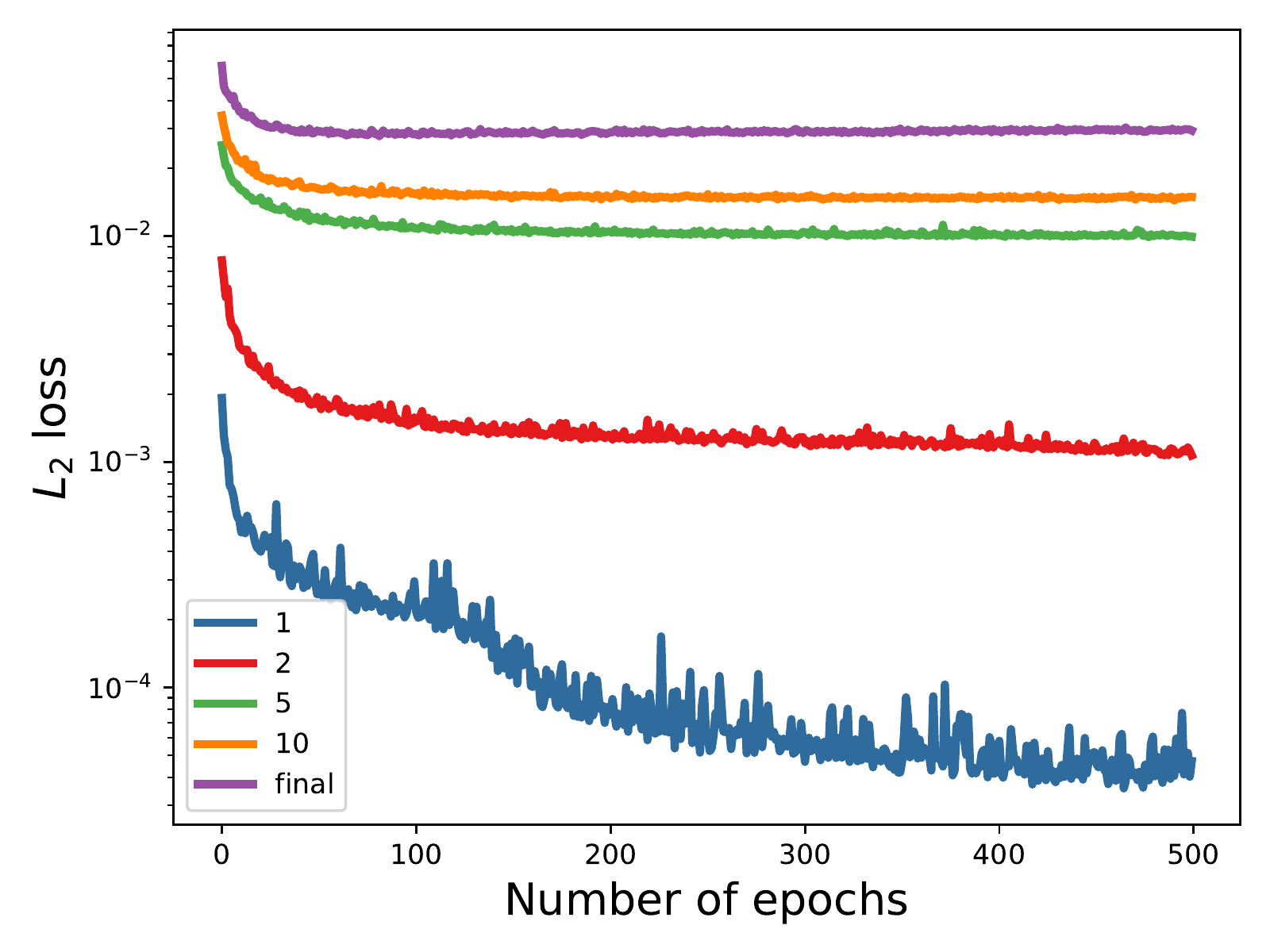}
    \includegraphics[width=0.4\linewidth,trim={0in 0.0in 0.0in 0.0in},clip]{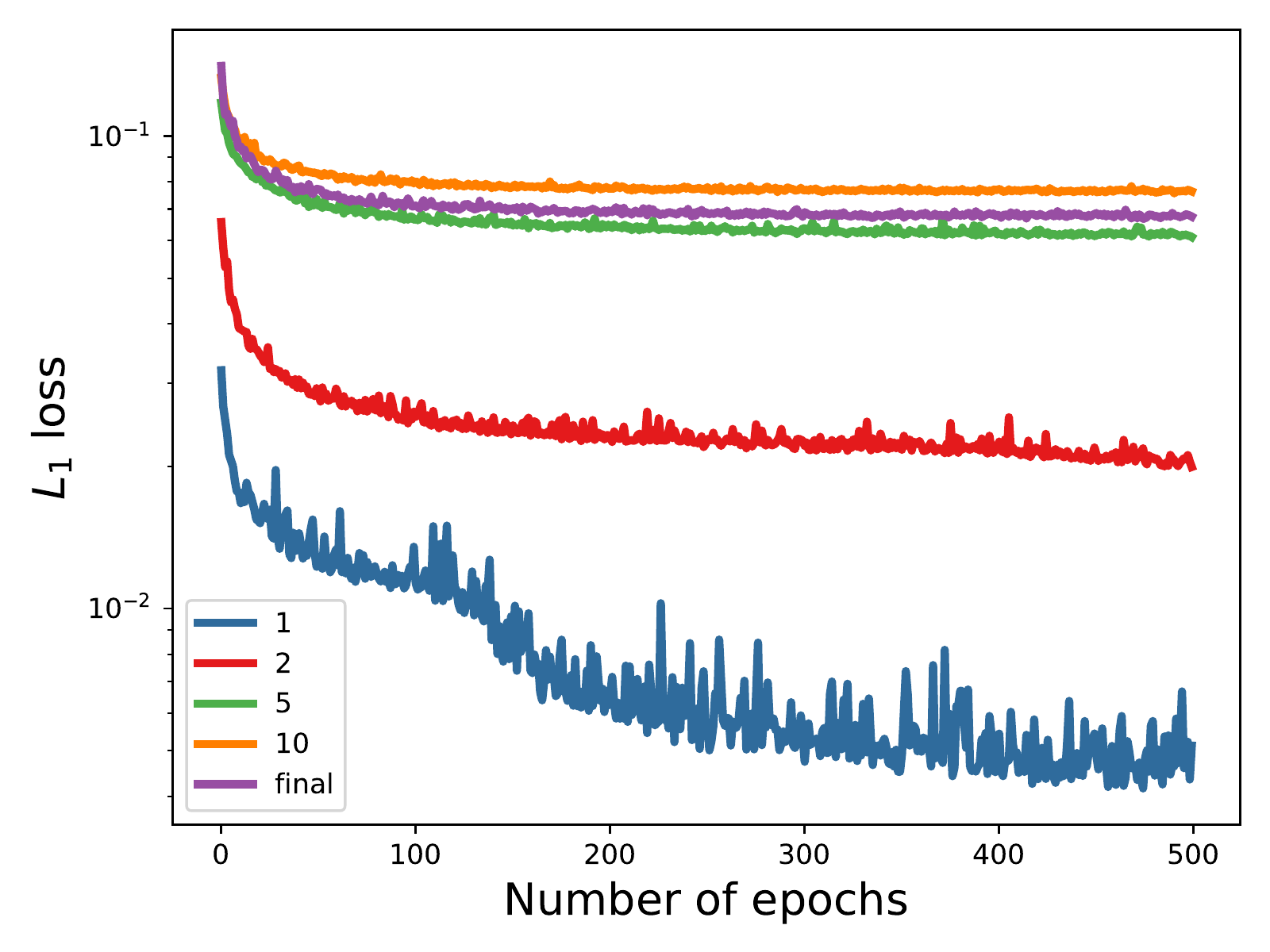}
    \caption{Performance plots of intermediate density prediction networks while predicting first, second, fifth, tenth intermediate densities and final density. (a) $L_2$ loss (b) $L_1$ loss.}
    \label{fig:perf_plots_idpn} 
\end{figure*}

\begin{figure*}[h!]
    \centering
    \includegraphics[width=0.4\linewidth,trim={0in 0.0in 0.0in 0.0in},clip]{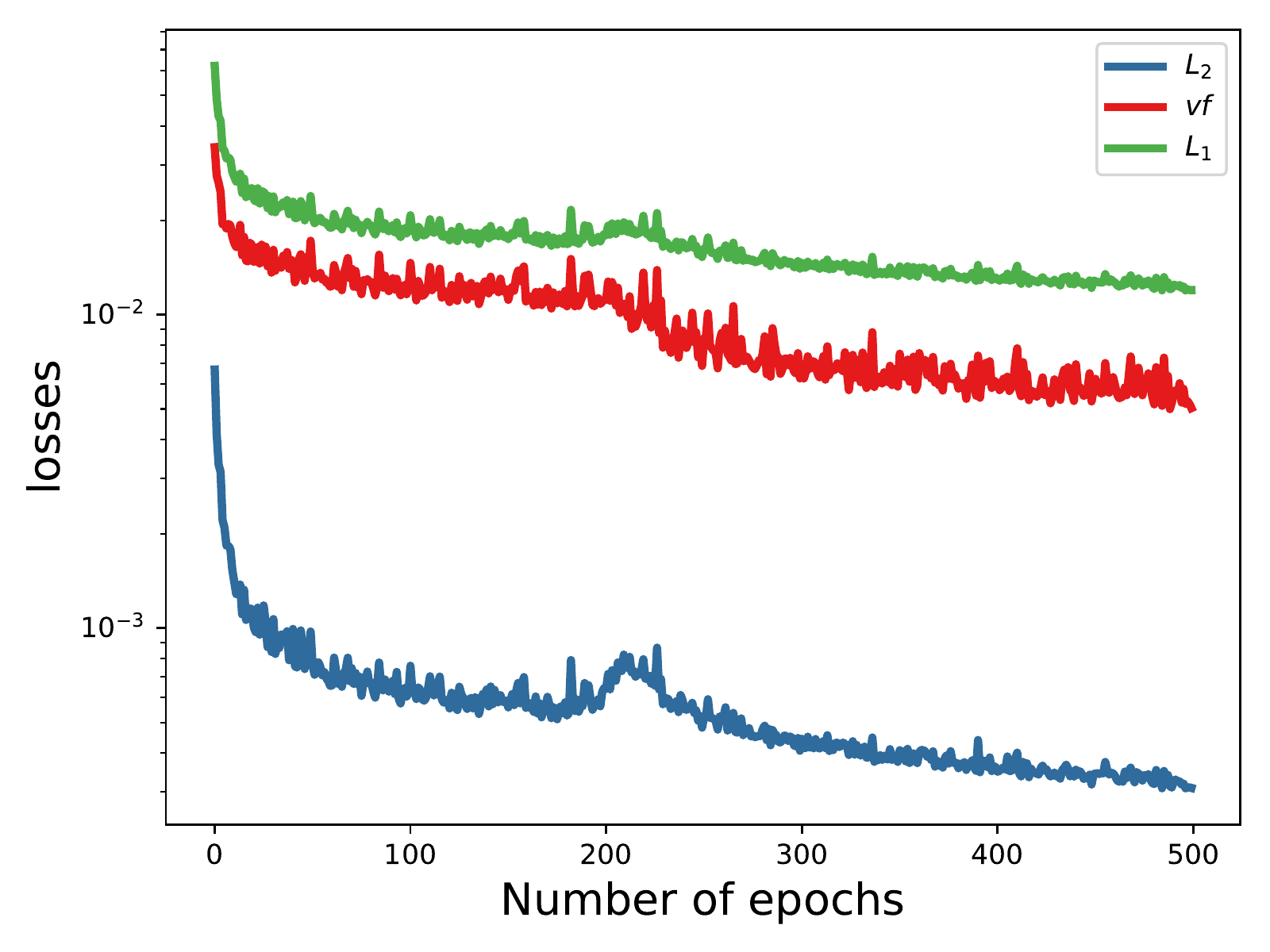}
    \includegraphics[width=0.4\linewidth,trim={0in 0.0in 0.0in 0.0in},clip]{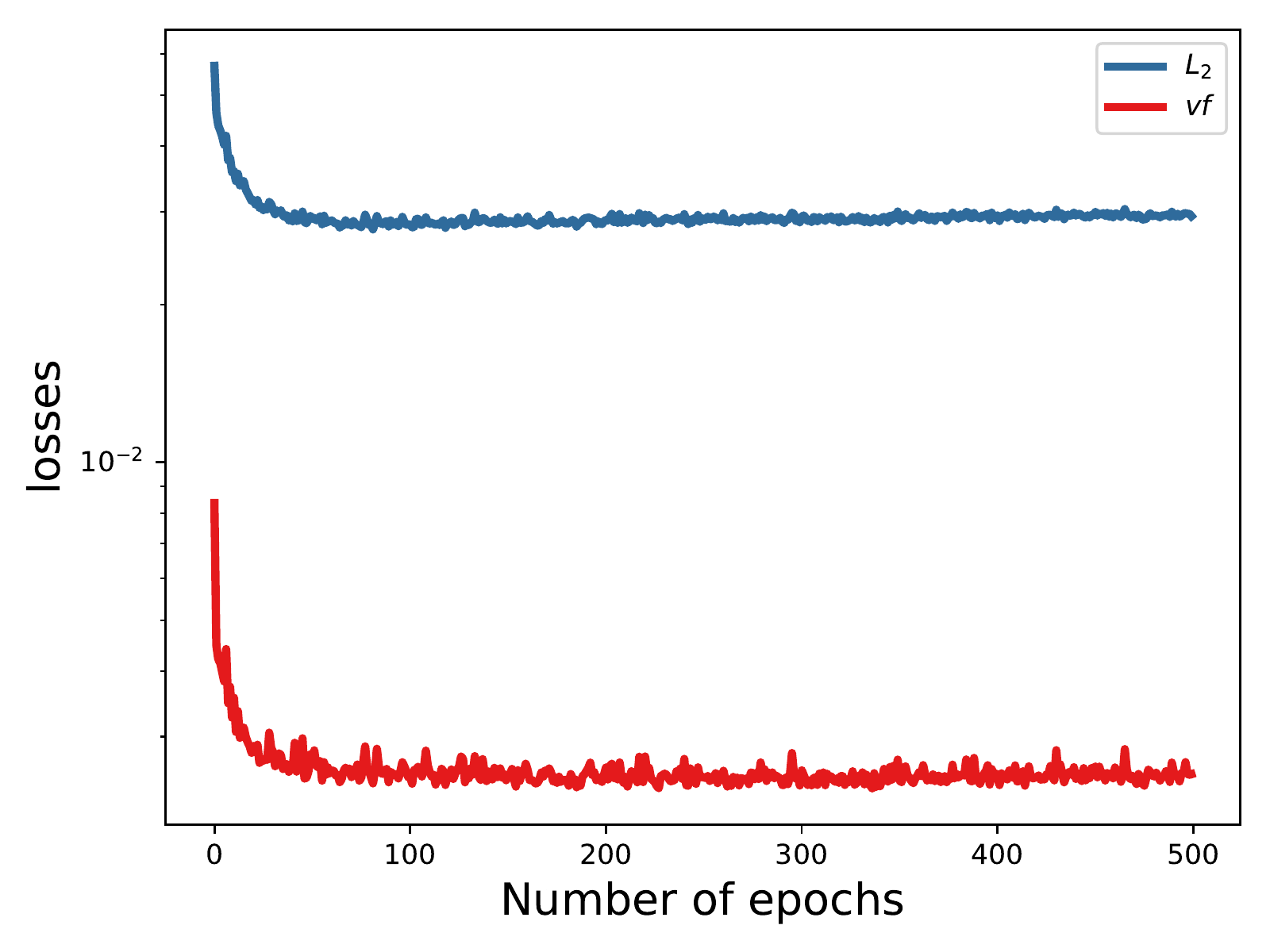}
    \caption{Performance plots of two phases of Density sequence prediction method on 2D data. (a) DS Phase1:IDPN loss (b) DS Phase2: DTN loss.}
    \label{fig:DSlosses_2d} 
\end{figure*}

\begin{figure*}[h!]
    \centering
    \includegraphics[width=0.32\linewidth,trim={0in 0.0in 0.0in 0.0in},clip]{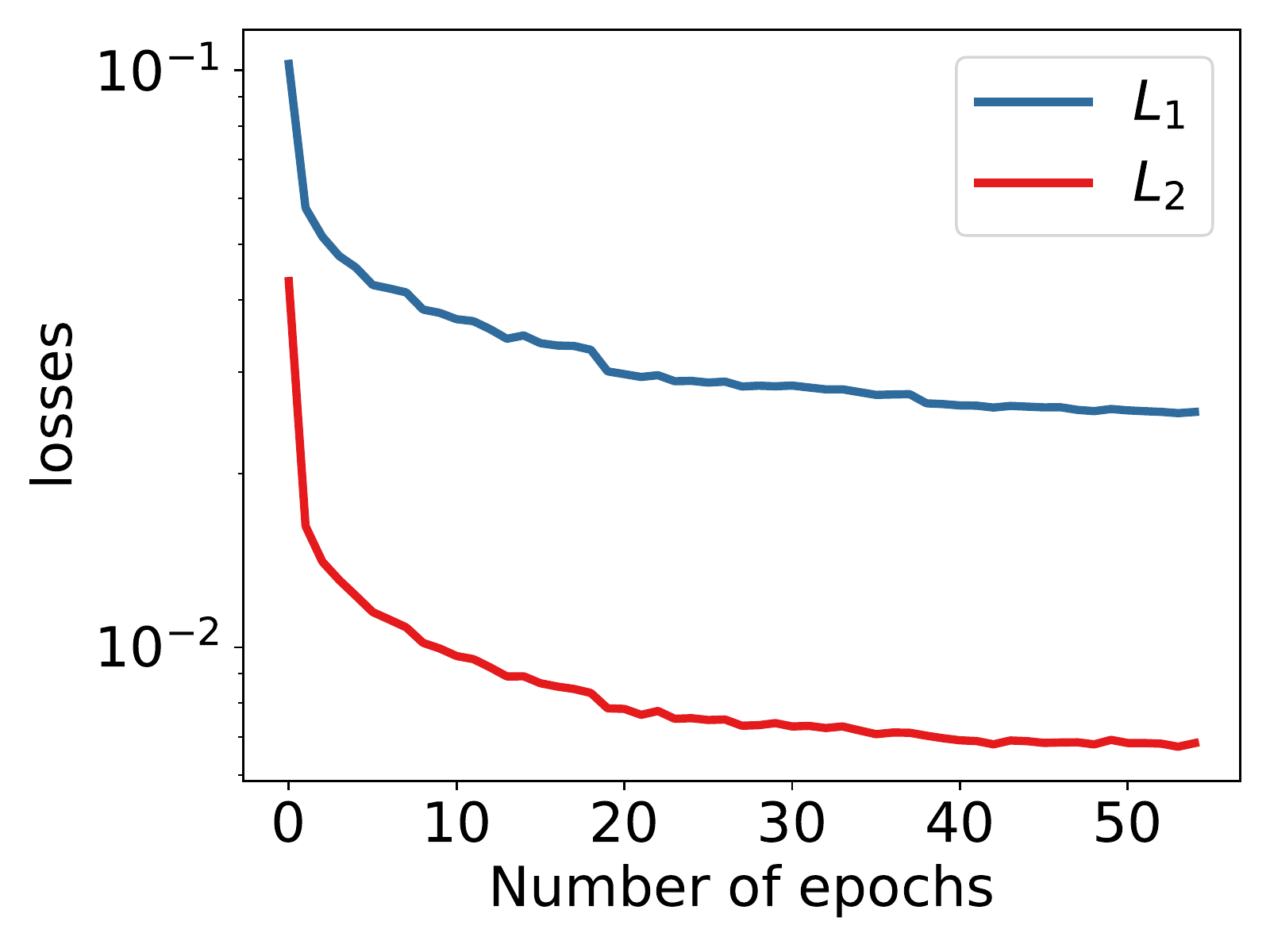}
    \includegraphics[width=0.32\linewidth,trim={0in 0.0in 0.0in 0.0in},clip]{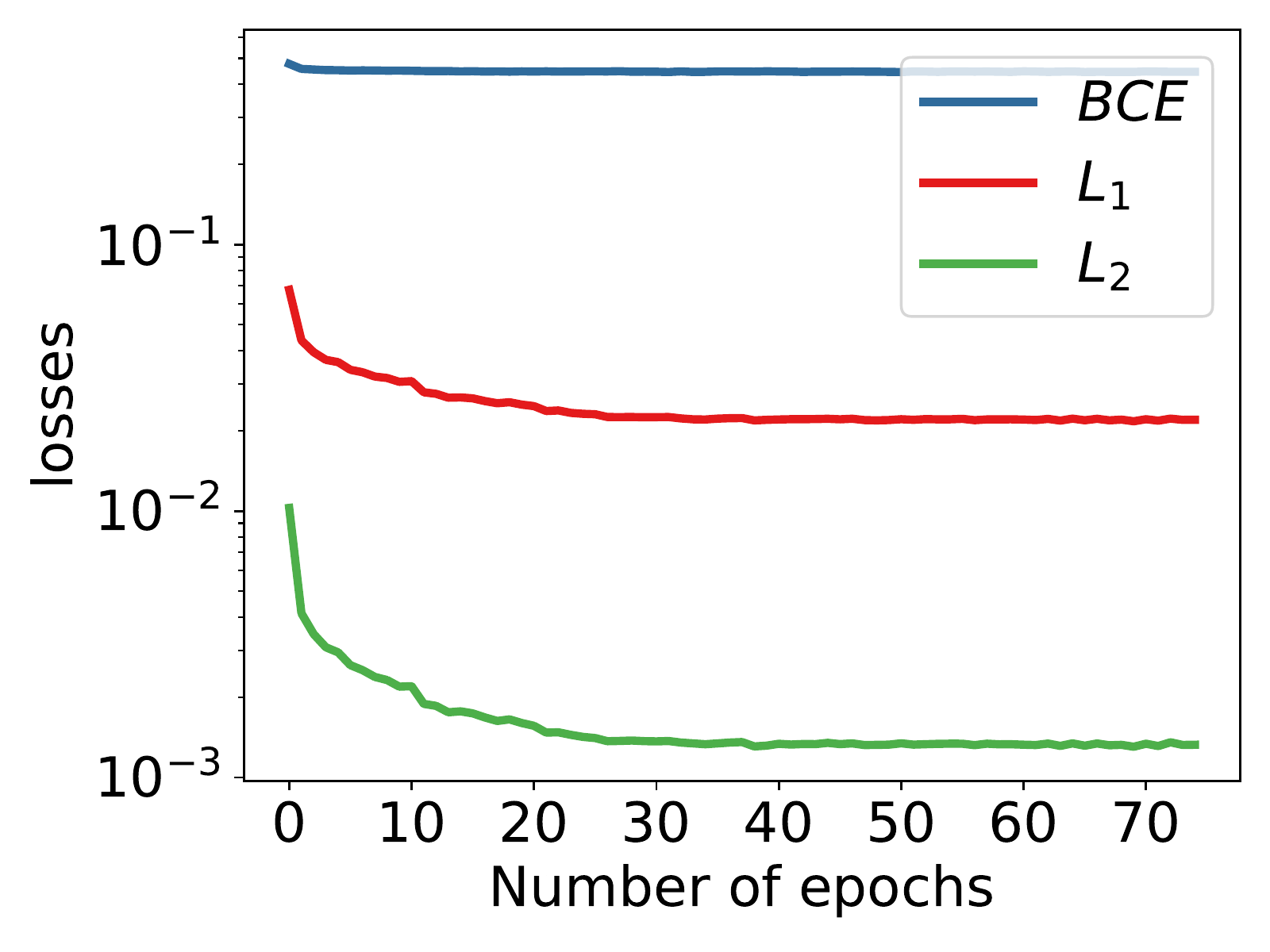}
    \includegraphics[width=0.32\linewidth,trim={0in 0.0in 0.0in 0.0in},clip]{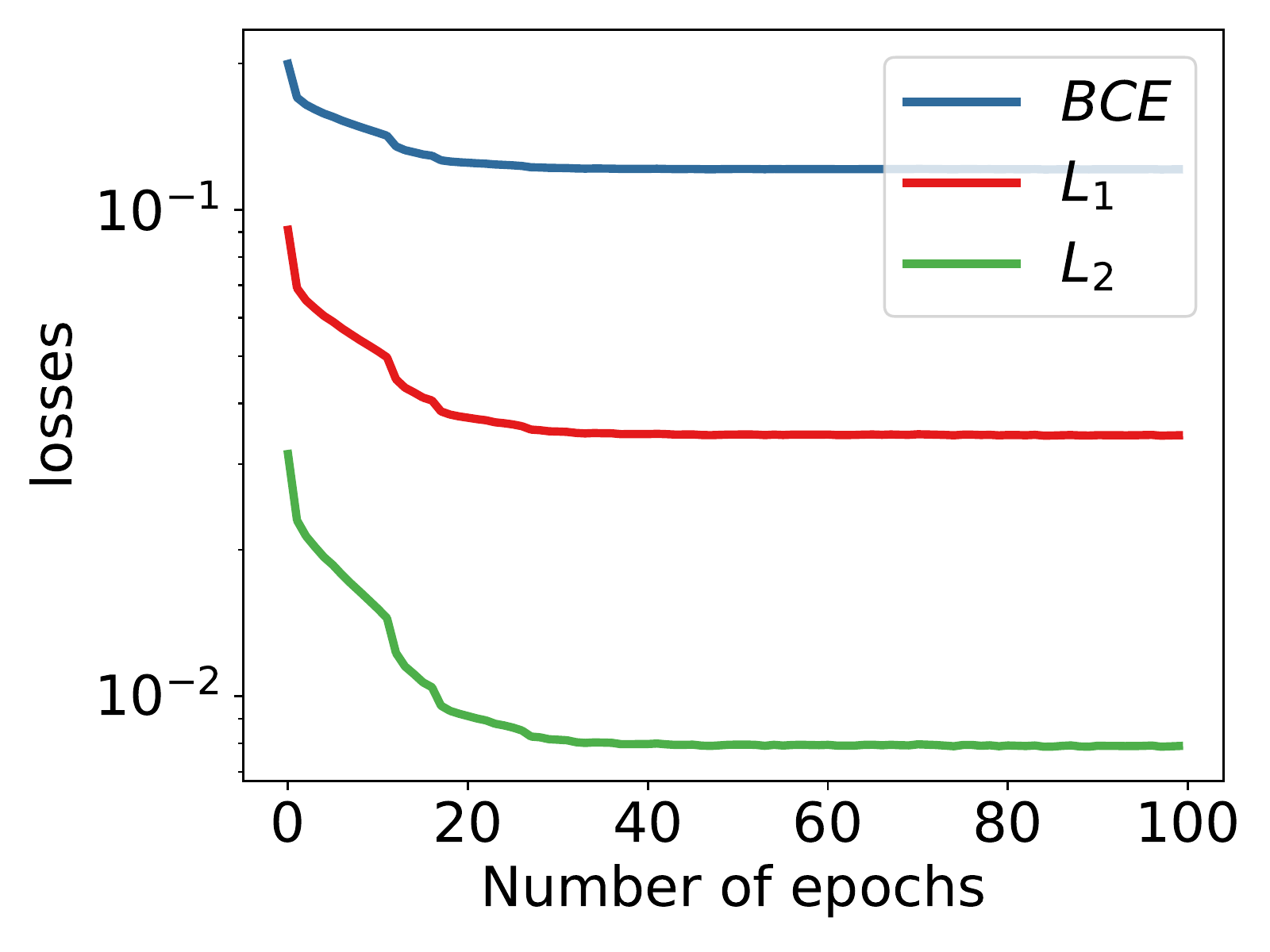}
    \caption{Performance plots of (a) Compliance Prediction Network (CPN) and (b) Density Prediction Network (DPN) and (c) Final Density Prediction Network (FDPN). Each plot shows different loss functions used in training with 2D data.}
    \label{fig:coupled_losses_2d} 
\end{figure*}

\begin{figure*}[t!]
    \centering
    \includegraphics[width=0.33\linewidth,trim={0in 0.0in 0.0in 0.0in},clip]{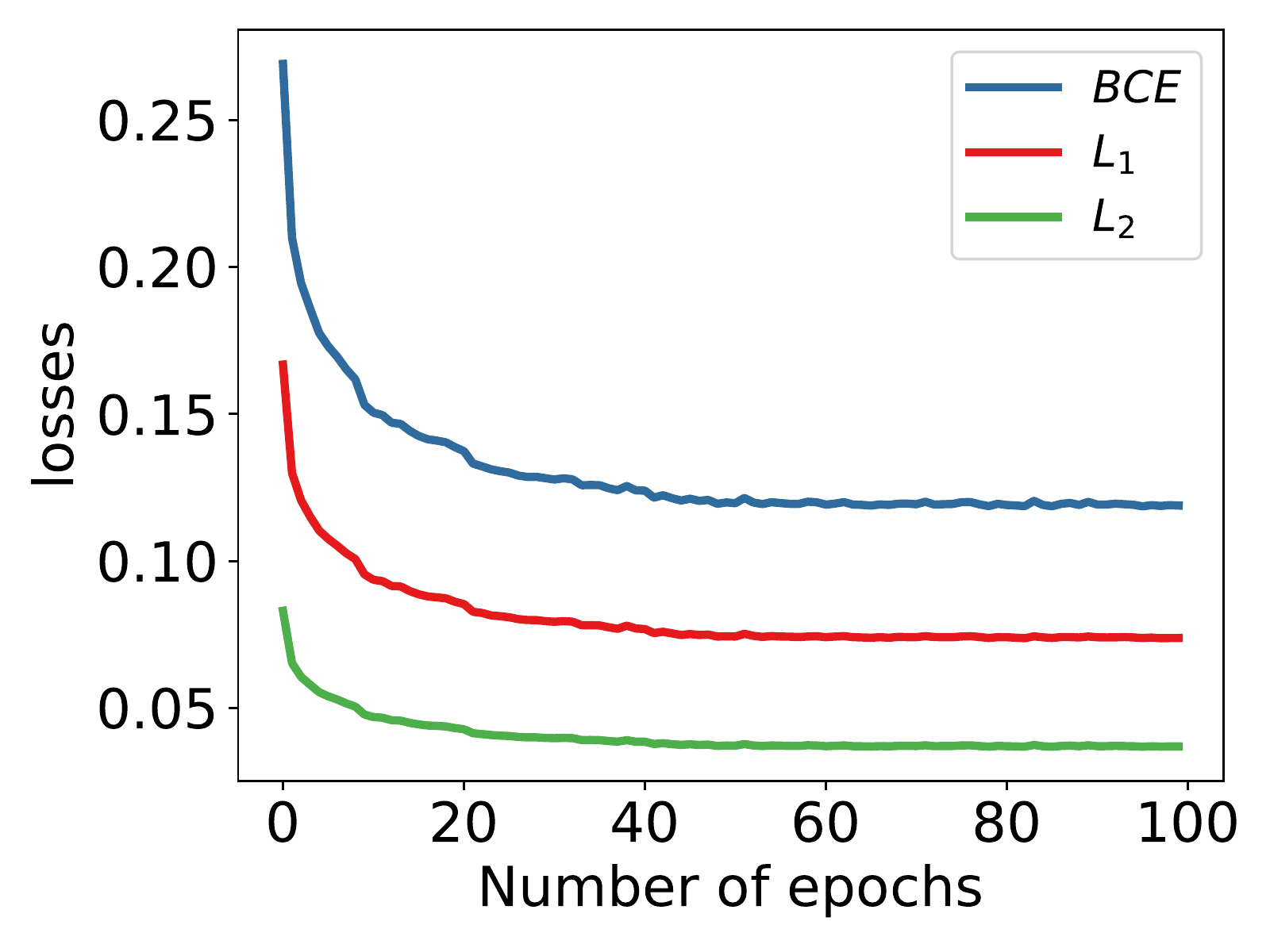}
    \caption{Performance plots of Direct Optimal Density Prediction. Plot shows different loss functions used in training with 3D data.}
    \label{fig:onestep_losses_3d} 
\end{figure*}

\begin{figure*}[t!]
    \centering
    \includegraphics[width=0.33\linewidth,trim={0in 0.0in 0.0in 0.0in},clip]{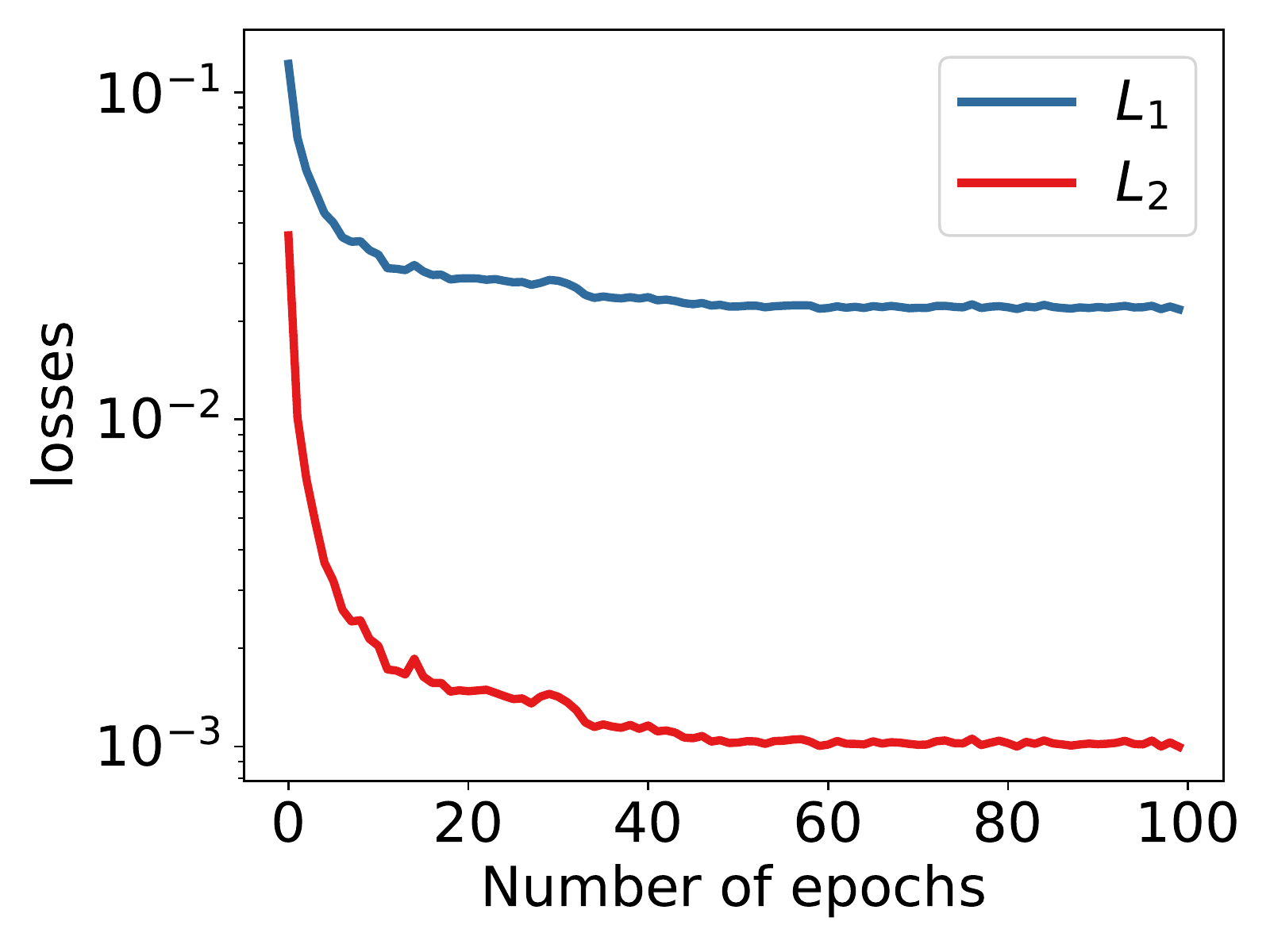}
    \includegraphics[width=0.33\linewidth,trim={0in 0.0in 0.0in 0.0in},clip]{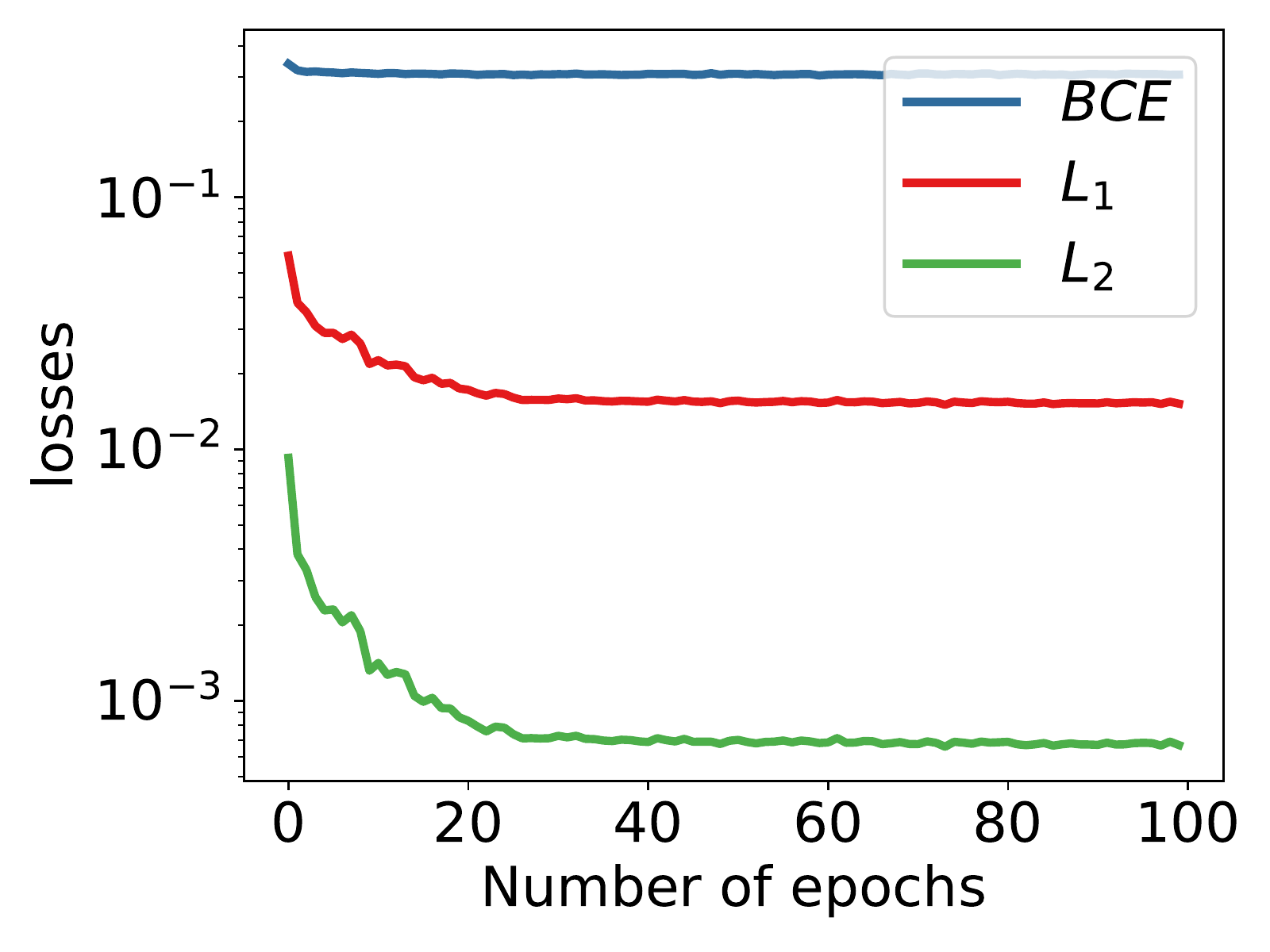}
    \includegraphics[width=0.33\linewidth,trim={0in 0.0in 0.0in 0.0in},clip]{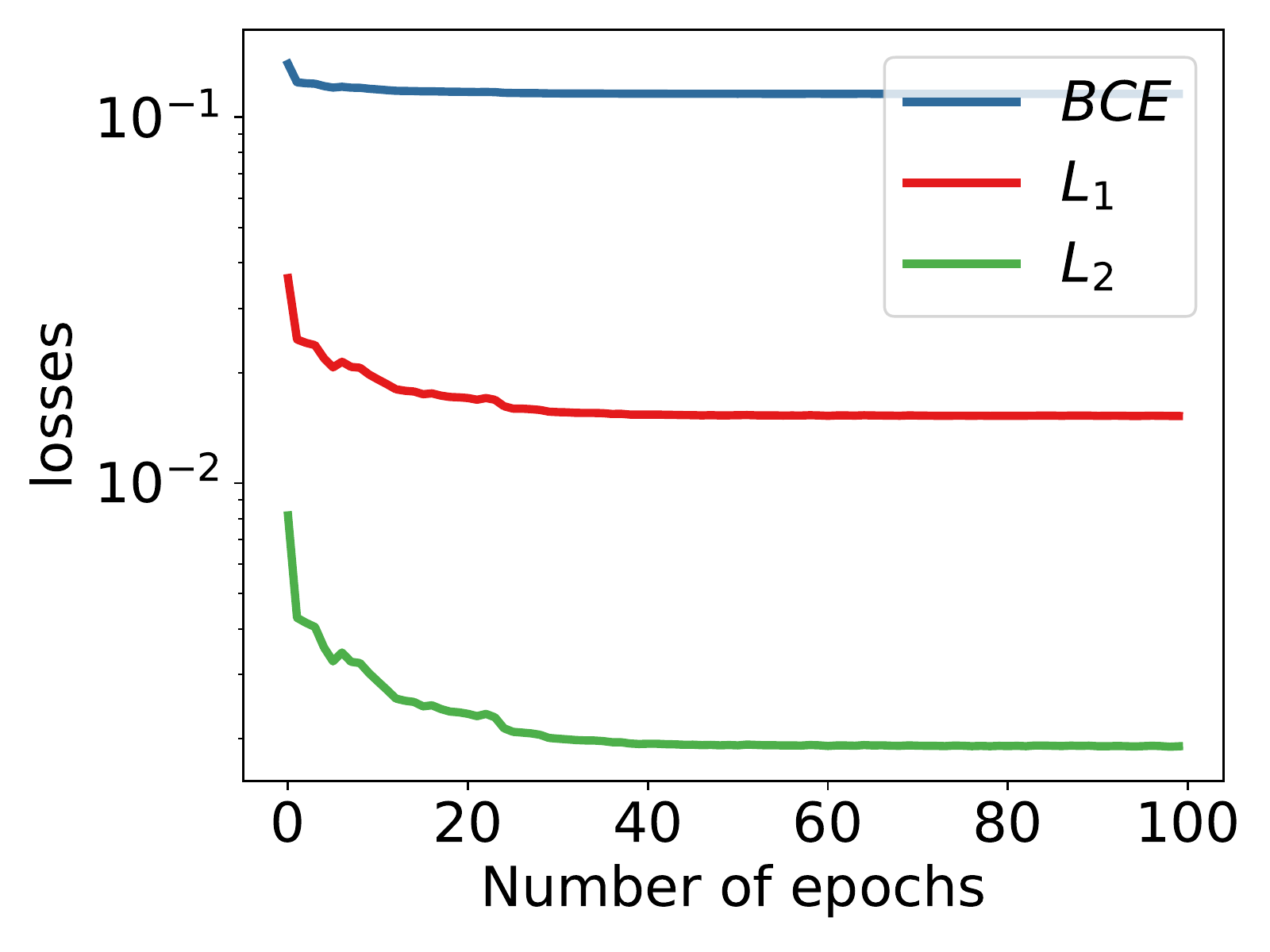}
    \caption{Performance plots of (a) Compliance Prediction Network (CPN) and (b) Density Prediction Network (DPN) and (c) Final Density Prediction Network (FDPN). Each plot shows different loss functions used in training with 3D data.}
    \label{fig:coupled_losses_3d} 
\end{figure*}

In this section, we summarize different performance plots of training the neural networks used in the three frameworks explored for both 2D and 3D data. \figref{fig:perf_plots_idpn} shows the L2 and L1 loss plots when DOD is used to predict the first, second, fifth, tenth, and final densities of the 2D dataset. In \figref{fig:DSlosses_2d}, the left plot shows the training losses(L2 loss, L1 loss, and L2 loss of VF) of IDPN(Phase1 of DS), and similarly, the right plot is of training losses(L2 loss and L2 loss of VF) of DPN(Phase2 of DS). \figref{fig:coupled_losses_2d} shows the training losses of all three networks of the CDCS method on 2D data. Similarly for 3D dataset, \figref{fig:onestep_losses_3d} has the plot of training losses of DOD, \figref{fig:coupled_losses_3d} has the loss plots for each of the three networks of CDCS. From these loss plots, we see the losses decrease as we progress in training.


\begin{figure*}[t!]
\centering
\includegraphics[width=0.33\linewidth,trim={0in 0.0in 0.0in 0.0in},clip]{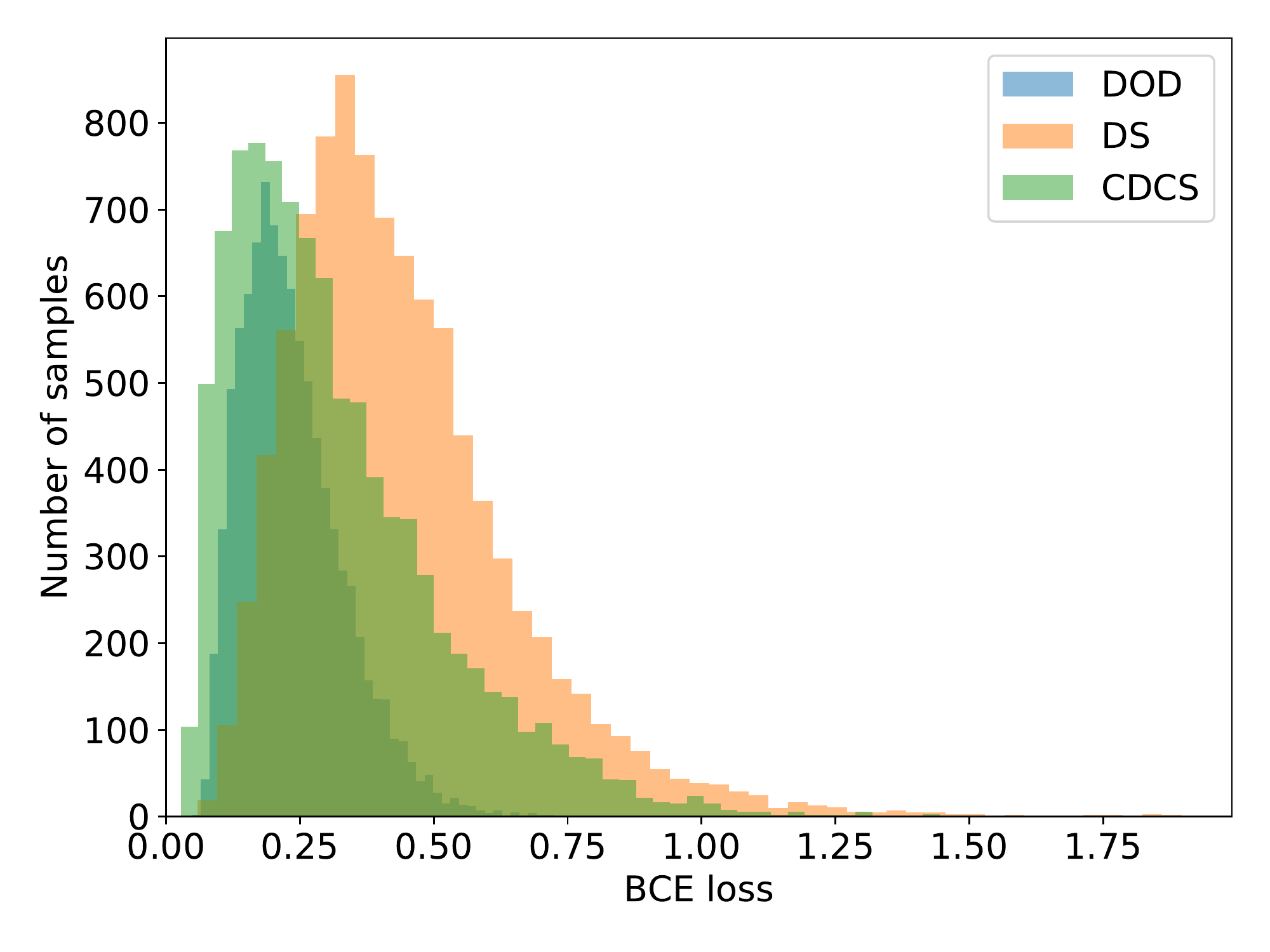}
\includegraphics[width=0.33\linewidth,trim={0in 0.0in 0.0in 0.0in},clip]{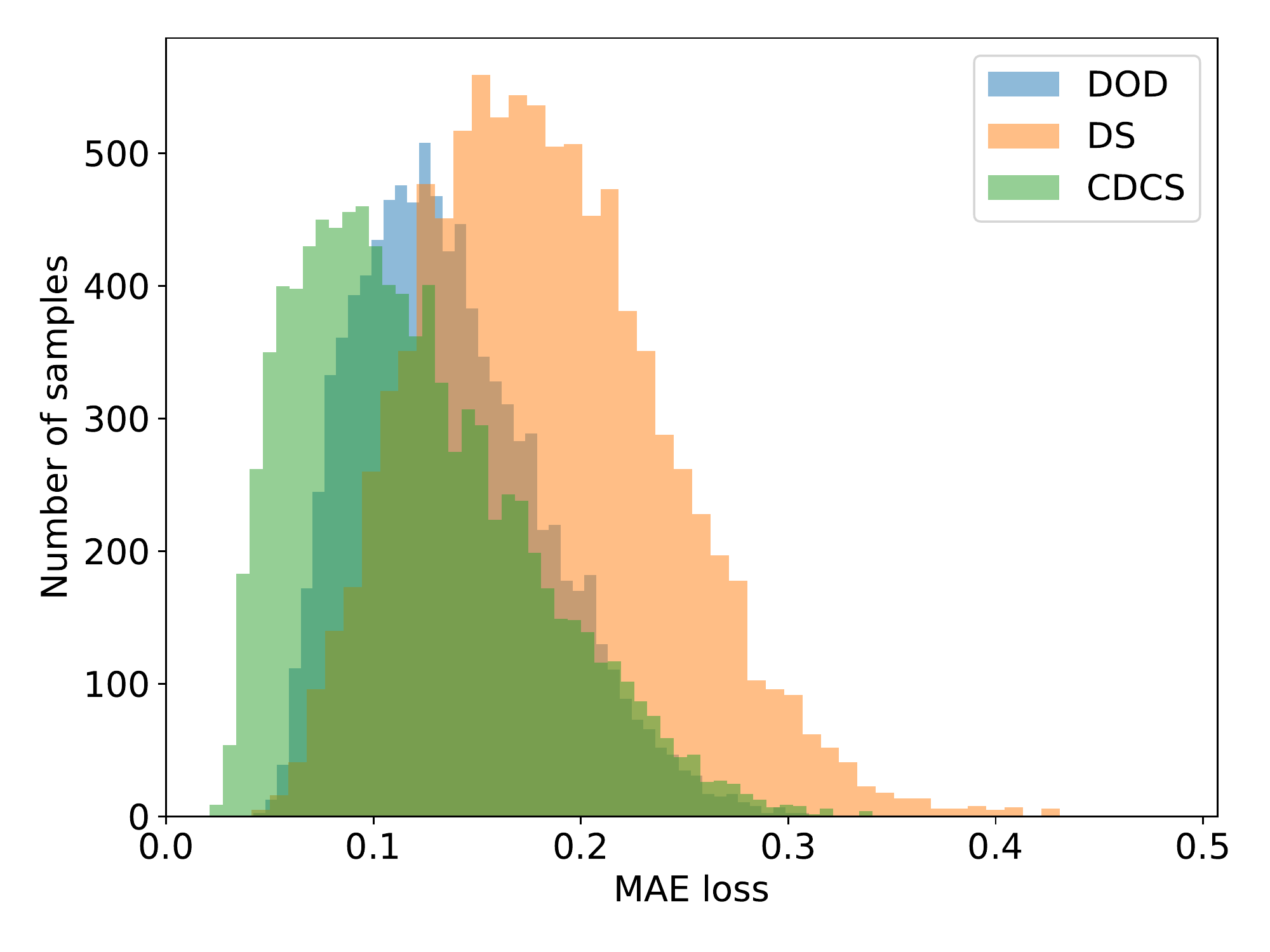}
\includegraphics[width=0.33\linewidth,trim={0in 0.0in 0.0in 0.0in},clip]{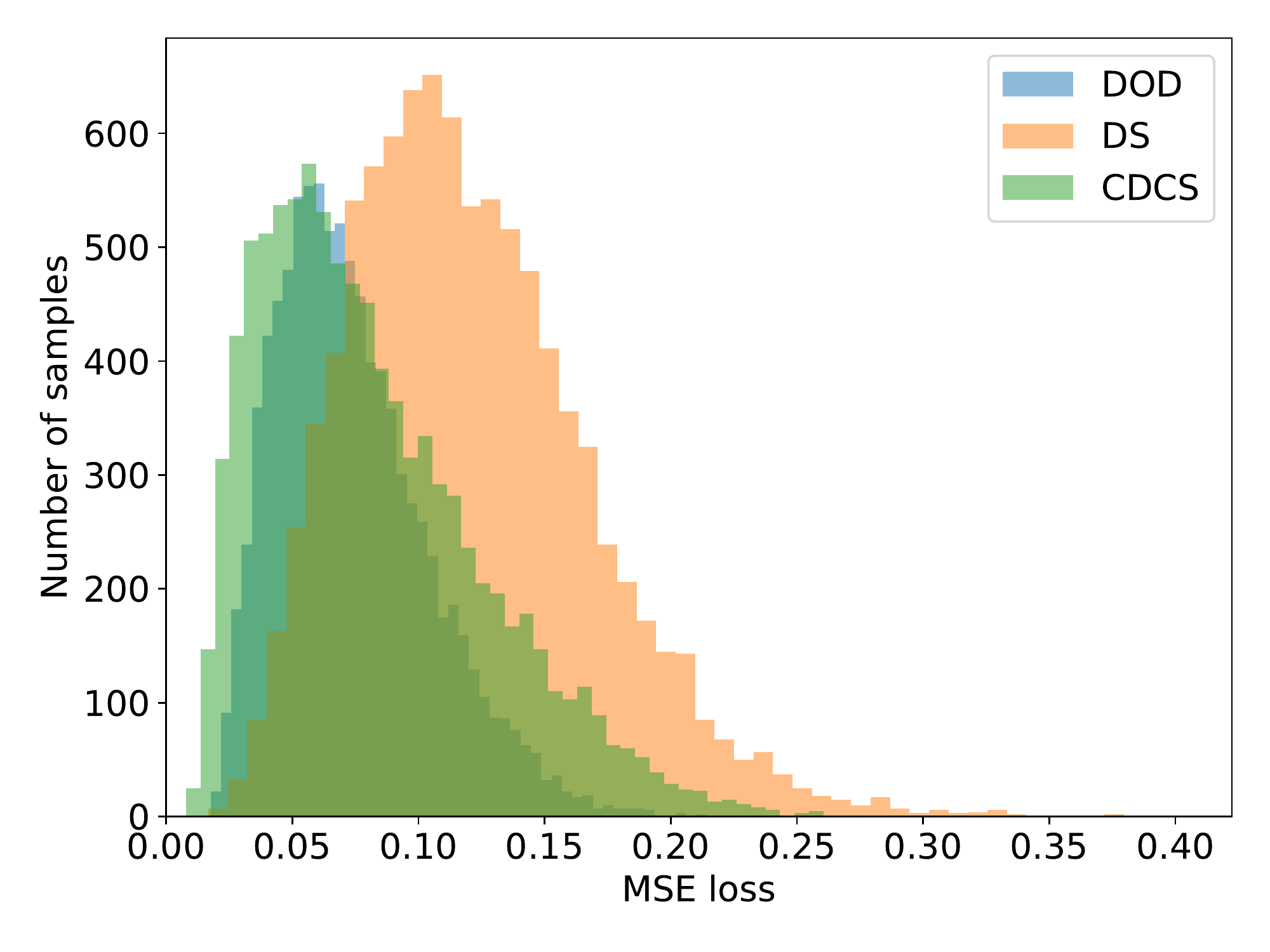}
\caption{Distribution of BCE, MAE, MSE losses on 2D test data.}
\label{fig:2d_loss_histograms} 
\end{figure*}

\begin{figure*}[t!]
\centering
\includegraphics[width=0.33\linewidth,trim={0in 0.0in 0.0in 0.0in},clip]{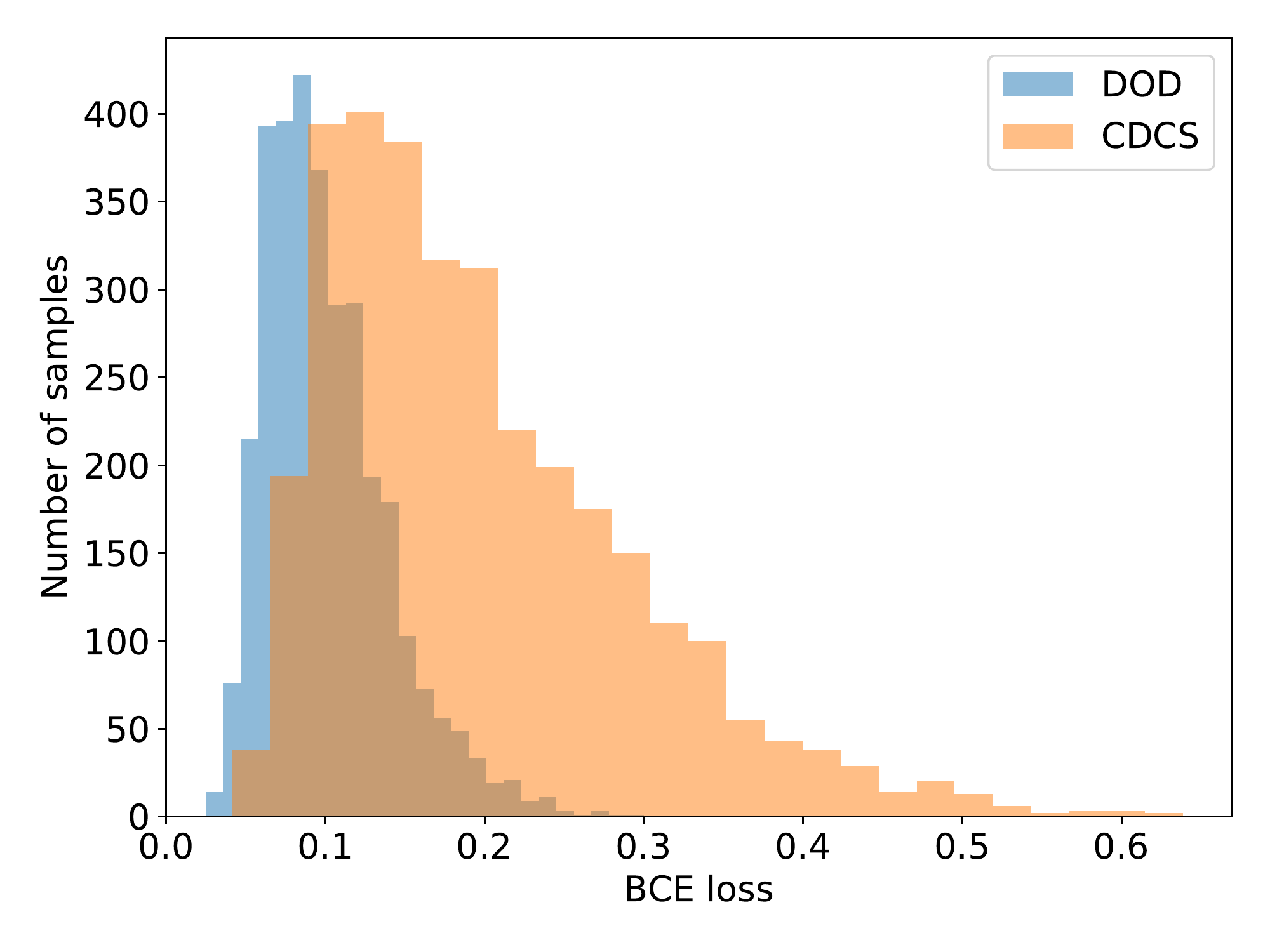}
\includegraphics[width=0.33\linewidth,trim={0in 0.0in 0.0in 0.0in},clip]{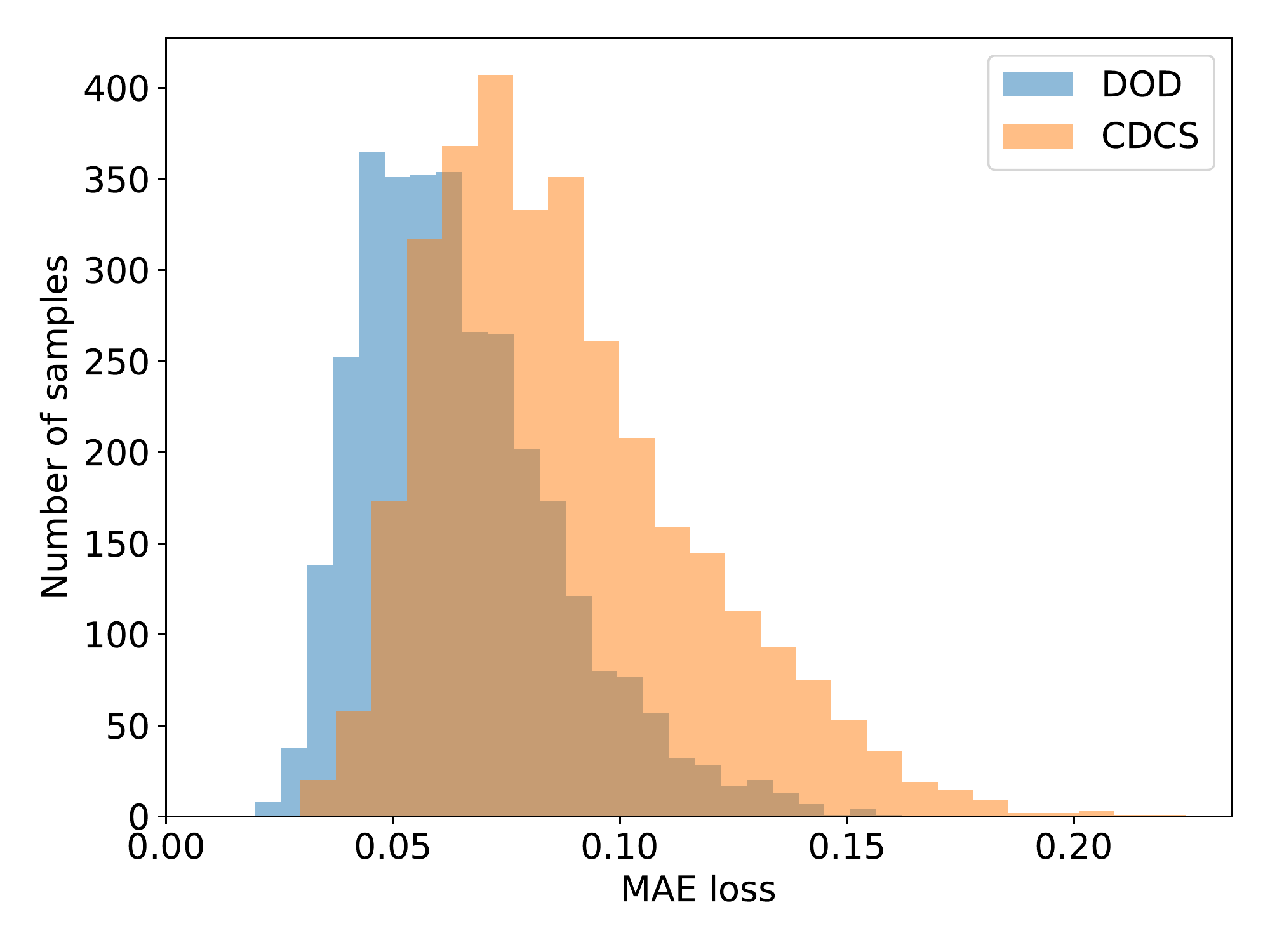}
\includegraphics[width=0.33\linewidth,trim={0in 0.0in 0.0in 0.0in},clip]{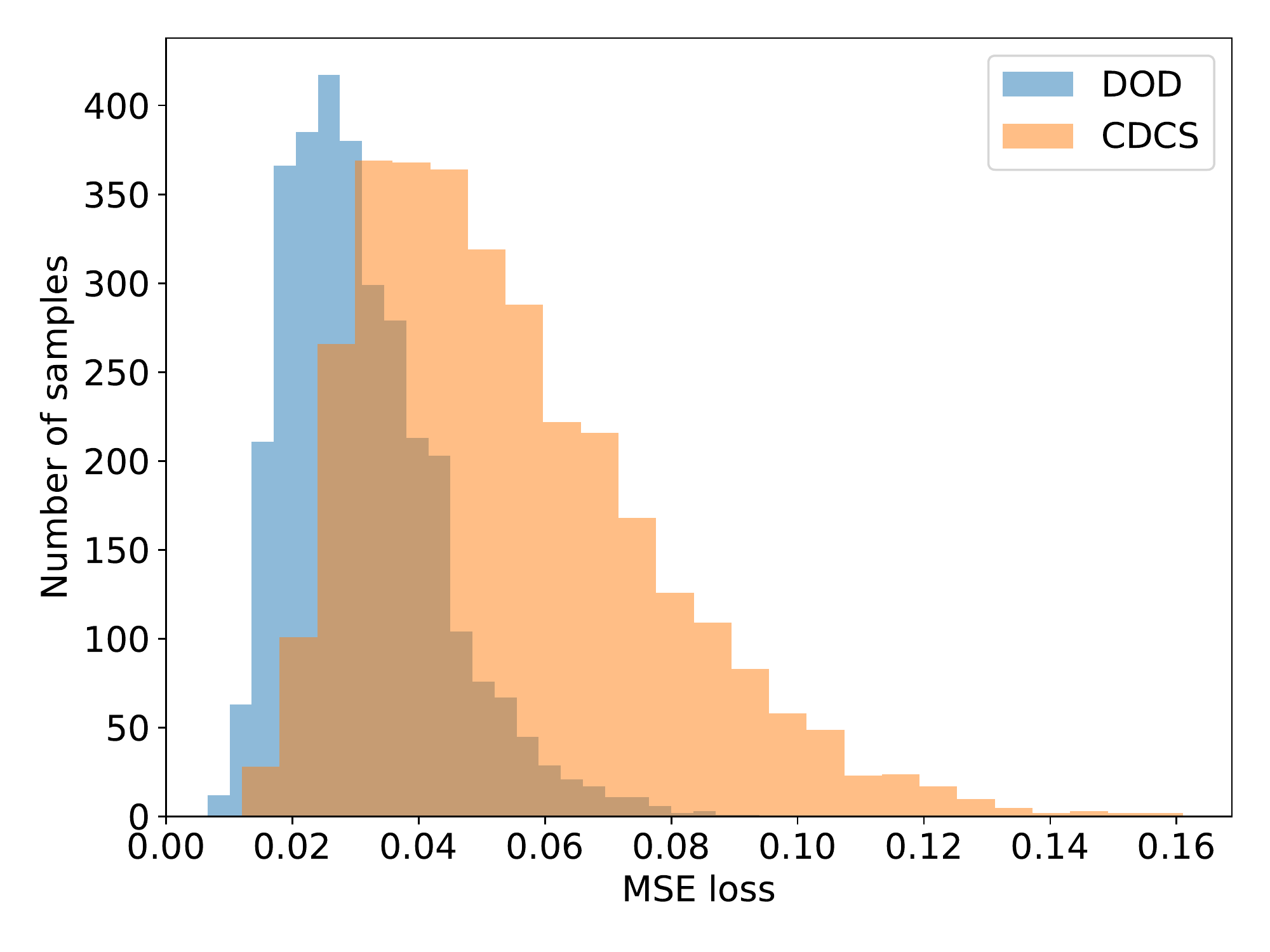}
\caption{Distribution of BCE, MAE, MSE losses on 3D test data.}
\label{fig:3d_loss_histograms} 
\end{figure*}

We also plot the histograms of different loss metrics values, between predicted and actual topology, like BCE, MAE and MSE loss for all the methods on 2D and 3D test dataset in \figref{fig:2d_loss_histograms} and \figref{fig:3d_loss_histograms}, respectively. Based on these histograms of metrics, in the 2D dataset, CDCS performs better than DOD and DS, while in the case of the 3D dataset, CDCS and DOD have comparable performance.

\section{Architectures}\label{sec:appendix_architecture}

In this section, we provide details about the different architectures used in the CDCS method. As mentioned earlier, we implemented Encoder-Decoder for CPN, U-SE-ResNet for DPN, and U-Net for the FDPN part.

\begin{figure*}[h!]
    \centering
    \includegraphics[width=0.9\linewidth,trim={1in 4.75in 1.25in 1.5in},clip]{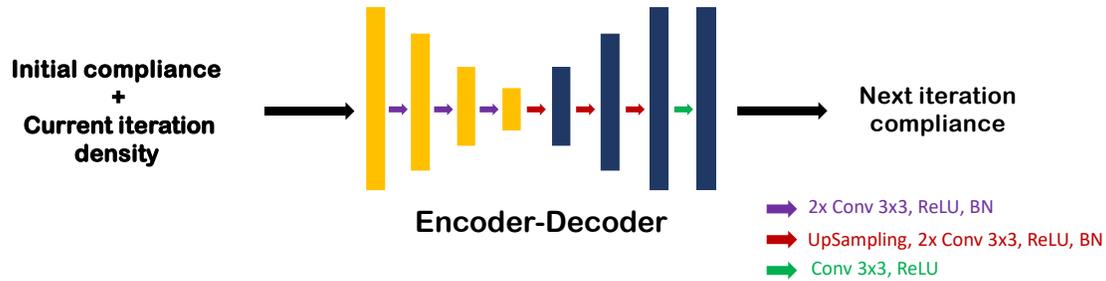}
    \caption{\textbf{Compliance Prediction Network (CPN) prediction:} This CNN Encoder-Decoder  model is used to predict the next iteration compliance.}
    \label{fig:cpn_architecture}
\end{figure*}

\begin{figure*}[h!]
    \centering
    \includegraphics[width=0.9\linewidth,trim={1in 4in 0.75in 1.5in},clip]{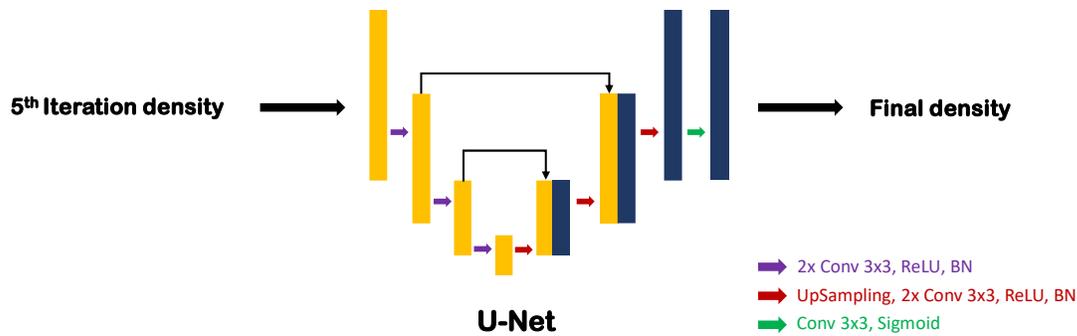}
    \caption{\textbf{Final Density Prediction Network (FDPN) prediction:} This U-Net architecture is used to predict the next iteration density.}
    \label{fig:fdpn_architecture}
\end{figure*}

\begin{figure*}[h!]
    \centering
    \includegraphics[width=0.9\linewidth,trim={1.1in 2.75in 0.5in 1.5in},clip]{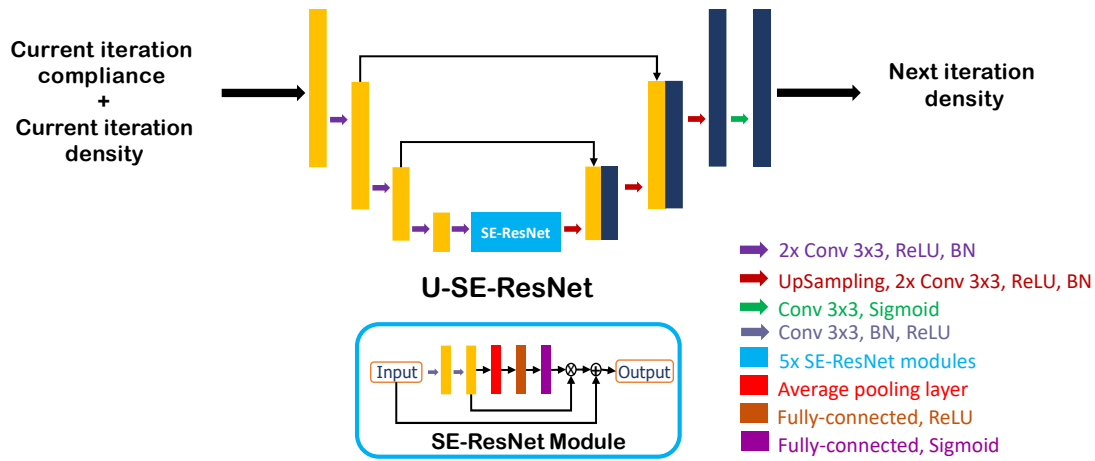}
    \caption{\textbf{Density Prediction Network (DPN) prediction:} This U-SE-ResNet\citep{nie2020topologygan} architecture is used to predict the next iteration density.}
    \label{fig:dpn_architecture}
    \vfill
\end{figure*}

The next architecture for FDPN is a U-Net~\citep{ronneberger2015u,cciccek20163d} based network as shown in \figref{fig:dpn_architecture}. This architecture is the modified version of the encoder-decoder architecture. As we can see in \figref{fig:fdpn_architecture}, the skip-connections are introduced from encoder part to decoder part at each resolution level. These connections help transfer the encoder's contextual information to the decoder for better localization~\citep{ronneberger2015u}. The encoder and decoder of U-Net are the same as in the encoder-decoder architecture discussed above.

The Encoder-Decoder architecture is a simple convolution neural network (CNN) consisting of two parts: the encoder and the decoder (\figref{fig:cpn_architecture}). The input is passed through the encoder and converted to a lower-dimensional latent space, further expanding to the higher dimension required by the decoder. The encoder is a collection of encoding blocks that consist of strided convolutional layers, followed by non-linearity (ReLU) transformation and batch normalization. Similarly, the decoder blocks of the decoder have an up-sampling layer, convolutional layers, non-linearity(ReLU), and batch normalization. Finally, we use the last convolution and non-linearity to get the output of the desired shape.

U-SE-ResNet~\citep{nie2020topologygan} is constructed using U-Net with addition of SE-ResNet blocks as shown in \figref{fig:dpn_architecture}. Each SE-ResNet block is a combination of ResNet and Squeeze-and-Excitation(SE) blocks~\citep{huSE}. These blocks are introduced in the U-Net architecture at the bottle-neck region between the encoder and decoder. SE block enhances the network's performance by recalibrating the channel-wise features by explicitly weighing the inter-dependencies between channels. SE block consists of a pooling layer followed by fully connected (FC) and ReLU transformation and again passing through the FC and sigmoid transformation. In the end, the output of the sigmoid layer is scaled by multiplying with the input of the SE block, with which we get the same shape as the input of the SE block. In SE-ResNet, ResNet is combined with SE block to improve the performance~\citep{nie2020topologygan} by adding the residual connection between the input to the output of the SE block. Also, in this architecture, we use the same encoder and decoder, as explained earlier.

\end{document}